\pdfoutput=1
\documentclass[11pt]{article}

\usepackage[final]{acl}

\usepackage{times}
\usepackage{latexsym}

\usepackage[T1]{fontenc}
\usepackage[utf8]{inputenc}

\usepackage{microtype}

\usepackage{inconsolata}

\usepackage{graphicx}

\usepackage{times}
\usepackage{url}

\usepackage{latexsym}

\usepackage{microtype}

\usepackage{times}
\usepackage{latexsym}
\usepackage{lingmacros}
\usepackage{amsmath}
\usepackage{graphicx}
\usepackage{array, tabularx, caption, boldline}
\usepackage{cellspace}
\usepackage{url}
\usepackage{makecell}
\usepackage{multirow}
\usepackage{xcolor}

\usepackage{subcaption}
\usepackage{booktabs}
\usepackage{comment}
\usepackage{arydshln}
\usepackage{amssymb}

\usepackage{gb4e}
\noautomath

\usepackage{pifont}% http://ctan.org/pkg/pifont

\definecolor{CBI1}{HTML}{636EFA}
\definecolor{CBI2}{HTML}{EF553B}
\definecolor{CBI3}{HTML}{00CC96}
\definecolor{CBI4}{HTML}{AB63FA}
\definecolor{CBI5}{HTML}{FFA15A}
\definecolor{CBI6}{HTML}{19D3F3}
\definecolor{CBI7}{HTML}{FF6692}

\newcommand{\secref}[1]{(\S\ref{#1})}

\usepackage{ulem}
\usepackage{amssymb}
\usepackage{mathtools}
\usepackage{mathabx}
\usepackage{wasysym}

\newdimen\supsymwidth
\newdimen\supsymheight
\newdimen\tgtsymwidth
\setbox0=\hbox{$_\bigboxvoid$}
\supsymwidth=\wd0
\supsymheight=\ht0
\setbox0=\hbox{$_\bigovoid$}
\tgtsymwidth=\wd0

\newcommand{\supsym}[1]{%
  \mathrlap{\bigboxvoid}
  \raisebox{0.5pt}{\hbox to \supsymwidth{\hfill{\tiny#1}\hfill}}
}

\newcommand{\tgtsym}[1]{
  \mathrlap{\bigovoid}
  \raisebox{0.5pt}{\hbox to \supsymwidth{\hfill{\tiny#1}\hfill}}
}

\newcommand{\superordinate}[2][\relax]{#2$_{\ifx#1\relax\bigovoid\else\supsym{#1}\fi}$}

\title{Representational Analysis of Binding in Language Models}

\author{Qin Dai$^1$, Benjamin Heinzerling$^{2,1}$, Kentaro Inui$^{3,1,2}$ \\ $^1$Tohoku University \qquad $^2$RIKEN AIP \qquad $^3$MBZUAI\\ \tt qin.dai.b8@tohoku.ac.jp, benjamin.heinzerling@riken.jp \\ \tt kentaro.inui@mbzuai.ac.ae}

\begin{document}
\maketitle
\begin{abstract}
Entity tracking is essential for complex reasoning. To perform in-context entity tracking, language models (LMs) must bind an entity to its attribute (e.g., bind a container to its content) to recall attribute for a given entity. For example, given a context mentioning ``The coffee is in Box Z, the stone is in Box M, the map is in Box H'', to infer ``Box Z contains the coffee'' later, LMs must bind ``Box Z'' to ``coffee''. 
To explain the binding behaviour of LMs, \citet{feng2023language} introduce a Binding ID mechanism and state that LMs use a abstract concept called Binding ID (BI) to internally mark entity-attribute pairs. However, they have not captured the Ordering ID (OI) from entity activations that directly determines the binding behaviour.
In this work, we provide a novel view of the BI mechanism by localizing OI and proving the causality between OI and binding behaviour. Specifically, 
%we apply principle component analysis as well as other dimension reduction methods such as independent component analysis to analyze the activations (or hidden state) of entity and attribute. 
by leveraging dimension reduction methods (e.g., PCA), we discover that there exists a low-rank subspace in the activations of LMs, that primarily encodes the order (i.e., OI) of entity and attribute. Moreover, we also discover the causal effect of OI on binding that when editing representations along the OI encoding direction, LMs tend to bind a given entity to other attributes accordingly.  For example, by patching activations along the OI encoding direction we can make the LM to infer ``Box Z contains the stone'' and ``Box Z contains the map''. The code and datasets used in this paper are available at \url{https://github.com/cl-tohoku/OI-Subspace}.
%\blfootnote{\hspace{-1.7em}$\ast$ Equal contribution}
\end{abstract}

\section{Introduction}

The ability of a model to track and maintain information associated with an entity in a context is essential for complex reasoning \cite{karttunen1976discourse,heim1983file,nieuwland2006peanuts,barzilay2008modeling,kamp2010discourse}. To recall attribute information for a given entity in a context, the model must bind entities to their attributes~\cite{feng2023language}. For example, given Sample \ref{ex:ent_track1} and \ref{ex:ent_track2}, a model must bind the entities (e.g., ``Box Z'', ``Box M'', ``Box H'', ``Alex'', ``John'' and ``Carl'') to their corresponding attributes (e.g., ``coffee'', ``stone'', ``map'', ``bean'', ``pie'' and ``fruit'') so as to recall (or answer) such as what is in ``Box Z'' or what is sold by ``Alex'' without confusion. Binding has also been studied as a fundamental problem in Psychology~\cite{treisman1996binding}. 

\begin{figure*}[t]
\centering
\includegraphics[width=12.0cm]{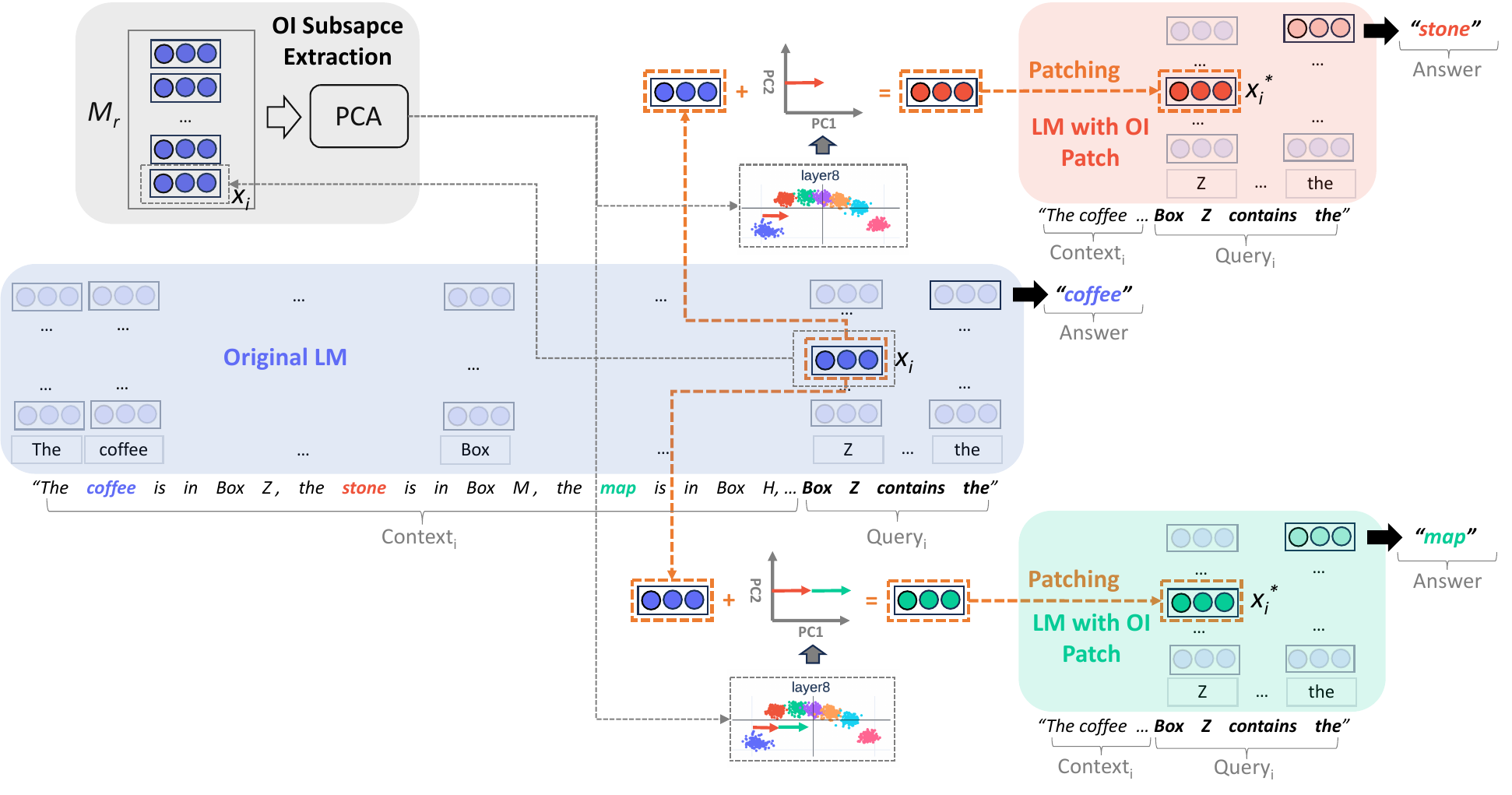}
\caption{Our main finding on Ordering ID (OI) subspace intervention. Patching entity (e.g., "Z") representations along OI direction (i.e., PC1) in activation space yields corresponding changes in model output.}
\label{fig:intro}
\end{figure*}

To uncover how Language Models (LMs) realize binding in term of internal representation, \citet{feng2023language} introduce a Binding ID mechanism and state that LMs apply a abstract concept called Binding ID (BI) to bind and mark Entity-Attribute (EA) pairs (e.g., ``Box Z'' and ``coffee'' in Sample~\ref{ex:ent_track1}, where BI is denoted as a numbered square). They also claim that the BI is represented as a vector to be added on the representation (or activation) of a EA pair so that the common vector is used as a key clue to search attribute for a given entity. However, they have not captured the Ordering ID (OI) information from the entity (or attribute) activations that causally affects binding behaviour and thus BI information as well. Here, OI is defined as the input order (or ordering index) of entities and attributes, no matter they are bound by a relation (e.g., ``is\_in'' in Sample~\ref{ex:ent_track1}) or not, such as the indexing number in Sample~\ref{ex:ent_track1} and Sample~\ref{ex:ent_track3}. We can observe that in a 1E-to-1A bound context, such as in Sample~\ref{ex:ent_track1}, BI and OI are interchangeable.

\begin{exe}
    \ex\label{ex:ent_track1}\textbf{Context}: The \superordinate[0]{coffee} is in Box \superordinate[0]{Z}, the \superordinate[1]{stone} is in Box \superordinate[1]{M}, the \superordinate[2]{map} is in Box \superordinate[2]{H}. \\ \textbf{Query}: Box \superordinate[0]{Z} contains the 
    \ex\label{ex:ent_track2}\textbf{Context}: The \superordinate[0]{bean} is sold by Person \superordinate[0]{Alex}, the \superordinate[1]{pie} is sold by Person \superordinate[1]{John}, the \superordinate[2]{fruit} is sold by Person \superordinate[2]{Carl}. \\ \textbf{Query}: Person \superordinate[0]{Alex} sells the
    \ex\label{ex:ent_track3}\textbf{Non-related Context}: The \superordinate[0]{coffee} and Box \superordinate[0]{Z} are scattered around, the \superordinate[1]{stone} is here and Box \superordinate[1]{M} is there, the \superordinate[2]{map} and Box \superordinate[2]{H} are in different place. \textbf{Query}: Box \superordinate[0]{Z} ...
\end{exe}

Since binding is the foundational skill that underlies entity tracking~\cite{feng2023language}, in this work, we take the entity tracking task~\cite{kim2023entity,prakash2024fine} as a benchmark to analyze the LM's binding behaviour. Based on the analysis of internal representation on this task, we localize the OI information from the activiations and provide a novel view of the BI mechanism. Specifically, we apply Principle Component Analysis (PCA) as well as other dimension reduction methods such as Independent Component Analysis (ICA)~\footnote{See Appendix~\secref{sec:pls} and Appendix~\secref{sec:ica} for details.} to analyze the activations of LMs, and which are empirically proven to be effective. We discover that LMs encode (or store) the OI information into a low-rank subspace (called OI subspace hereafter), and the discovered OI subspace can causally affect binding behaviour and thus BI information as well. That is, we find that by causally intervening along the OI encoding Principle Component (PC), LMs swap the binding and infer a new attribute for a given entity accordingly. For example, as shown in Figure~\ref{fig:intro}, by patching activations along the direction (i.e., PC1), we can make the LMs to infer ``Box Z contains the stone'' and ``Box Z contains the map'' instead of ``Box Z contains the coffee''. 
Therefore, our findings extend the previous BI based understanding of binding in LMs~\cite{feng2023language} by revealing the causality between OI and binding.

%Besids PCA, we also attempt artial least squares regression for capturing OI subspace. Since they achieve similar regression score, we adopt PCA for simplicity.

%Going beyond the prior work on BI mechanism~\cite{feng2023language}, we show that by causally intervening along the OI encoding Principle Component (PC), LMs swap the binding and infer a new attribute for a given entity accordingly. 
%That is, we find a consistent causal relationship between the OI subspace intervention and the inferred attributes by LMs. 
%For example, as shown in Figue~\ref{fig:intro}, by patching activations along a direction (i.e., PC1), we can make the LMs to infer "Box Z contains the stone" and "Box Z contains the map" instead of "Box Z contains the coffee". 

Overall, our findings suggest that LMs encode OI information into a subspace of LMs' activations that primarily encodes the order index of entities and attributes in a given context. What is more, the discovered OI subspace plays a crucial role in the in-context binding computation. In addition, we find that such OI subspace that determines binding is prevalent across multiple LM families such as Llama2~\cite{2307.09288} (and Llama3~\cite{llama3modelcard}), Qwen1.5~\cite{bai2023qwen} and Pythia~\cite{biderman2023pythia}, and the code fine-tuned LM Float-7B~\cite{prakash2024fine}.

\section{Finding OI Subspace}
\label{sec:subspace}
In this section we describe our Principle Component Analysis (PCA) based method to localize the OI subspace in activations of LMs. As shown in Figure~\ref{fig:intro}, we firstly extract entity activation from LMs. Given a LM (e.g., Llama2), and a collection of texts which describes a set of EA pairs related by a relation such as ``is\_in'' in Sample~\ref{ex:ent_track1}, we extract the activation of entity token (e.g., ``Z'') in query (denoted as $\mathbf{x}_i$) from certain layer~\footnote{See Appendix~\secref{sec:layer_ld} for the layer selection.} and construct a activation matrix $M_r \in R^{n \times d}$ for a relation $r$, where $n$ denotes the number of entities and $d$ denotes the dimension of the activation. The row $i$ of $M_r$ is the activation of an entity token (i.e., $\mathbf{x}_i$). 

PCA has been applied for identifying various subspace (or direction) such as the subspace encoding language bias~\cite{yang2021simple}, truth value of assertions~\cite{marks2023geometry} and sentiment~\cite{tigges2023linear}. Inspired by these studies, we choose PCA as our first attempt to localize OI subspace. In addition, we also apply other dimension reduction methods such as Independent Component Analysis (ICA) to capture OI subspace, and which are empirically proven to be effective. See Appendix~\secref{sec:pls} and Appendix~\secref{sec:ica} for details.

We hypothesize that in a activation subspace, entities with the same OI tend to cluster together (w.r.t the ones with different OIs), even though these entities have different semantic meaning, and the OIs are encoded as directions (or a PC) in the subspace. For convenience, we number OIs in left-to-right order, and the leftmost OI$=0$.

Since PCA is applied to identify the principle directions from a multidimensional space, we leverage PCA to capture OI subspace (or direction) from activations of LMs. Specifically, the PCA of a activation matrix is ${M_r} = U_r \Sigma_r V^T_r$, where the columns of $V_r \in R^{d \times d}$ are principle directions of $M_r$. We takes first $c$ columns of $V_r$ as the OI direction, denoted as $B_r \in R^{d \times c}$. 

\section{OI Subspace Visualization}

\begin{figure}[t]
\centering
\includegraphics[width=8.0cm]{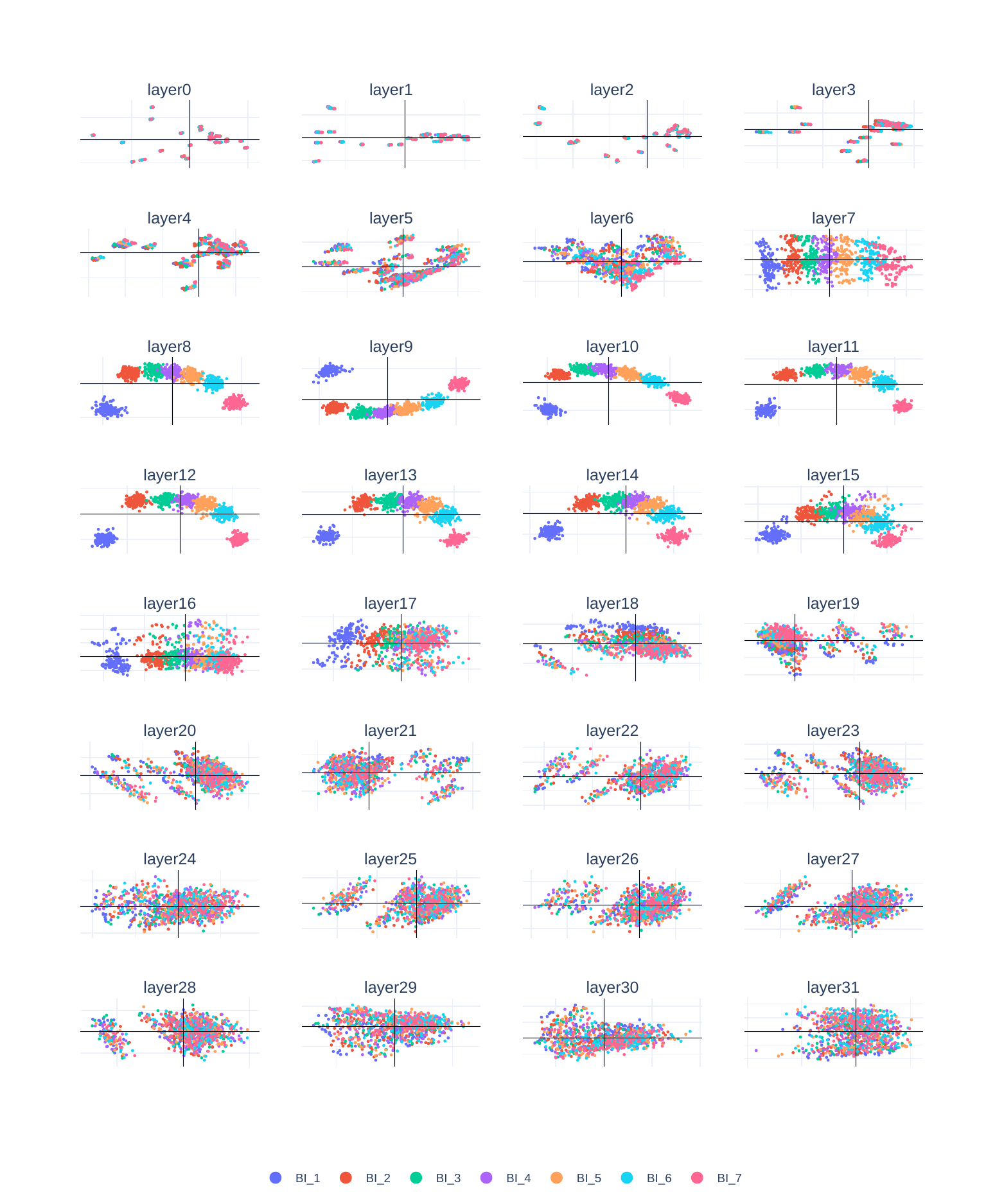}
\caption{Layer-wise OI subspace visualization on Llama2-7B, where ``BI'' primarily denotes OI.}
\label{fig:layer_wise_pca}
\end{figure}

We adopt a subset of the entity tracking dataset~\cite{kim2023entity,prakash2024fine}, which contains $n=1000$ samples, to create layer ($l$) wise activation matrix $M_r^l$. We then uses the $M_r^l$ to extract the layer-wise OI subspace projection matrix $B_r^l \in R^{d \times 2}$ to visualize the activations. Figure~\ref{fig:layer_wise_pca} shows the embedding visualization on Llama2-7B, where each point represents the activation of an entity projected via the $B_r^l$, and the colors represent OIs. From which, we can observe that middle layers, such as layer 8, have a clearly visible direction along which OI increases, while the others have tangled distribution. 

We also observe similar pattern of distribution on Llama3-8B, Float-7B~\secref{sec:2emb_visual} and other LM families such as Qwen1.5 and Pythia~\secref{sec:other_llm}. This indicates that LMs use the middle layers to encode OI information, and the finding is prevalent across multiple LM families. This finding is also consistent with the ``stages of inference hypothesis''~\cite{lad2024remarkable} stating that the function of early layers is to perform detokenization, middle layers do feature engineering, and late layers map the representations from the middle layers into the output embedding space for next-token prediction. According to the hypothesis, we would expect to find the ordering feature most prominently represented in middle layers, which is exactly what the visualization shows. We call this dimension that represents OI as OI Principle Component (OI-PC). In the following section, we apply causal intervention on the OI-PC to analyze how OI-PC affect the model output.

\section{Causal Interventions on OI Subspace}
\label{sec:intervene}

\begin{table*}[t]
\centering
\scalebox{0.7}{
\begin{tabular}{lccccccc}
\hline
 & &\multicolumn{6}{c}{\makecell[c]{\textbf{Answer for \# Step}}} \\\cmidrule(lr){3-8}
\textbf{Context} & \textbf{Query} & 1 & 2 & 3 & 4 & 5 & 6 \\\cmidrule(lr){1-8}
 \makecell[l]{The \textcolor{CBI1}{coffee} is in Box Z, the \textcolor{CBI2}{stone} is in Box M,\\ the \textcolor{CBI3}{map} is in Box H, the \textcolor{CBI4}{coat} is in Box L,\\ the \textcolor{CBI5}{string} is in Box T, the \textcolor{CBI6}{watch} is in Box E,\\ the \textcolor{CBI7}{meat} is in Box F.} & 
 \makecell[l]{Box Z\\ contains\\ the}& \textcolor{CBI2}{stone} & \textcolor{CBI3}{map} & \textcolor{CBI3}{map} & \textcolor{CBI5}{string} & \textcolor{CBI6}{watch} & \textcolor{CBI7}{meat} \\\cmidrule(lr){1-2}\cmidrule(lr){2-8}
\makecell[l]{The \textcolor{CBI1}{letter} is in Box Q, the \textcolor{CBI2}{boot} is in Box C,\\ the \textcolor{CBI3}{fan} is in Box N, the \textcolor{CBI4}{crown} is in Box R,\\ the \textcolor{CBI5}{guitar} is in Box E, the \textcolor{CBI6}{bag} is in Box D,\\ the \textcolor{CBI7}{watch} is in Box K.} & 
 \makecell[l]{Box Q\\ contains\\ the}& \textcolor{CBI2}{boot} & \textcolor{CBI3}{fan} & \textcolor{CBI4}{crown} & \textcolor{CBI5}{guitar} & \textcolor{CBI7}{watch} & \textcolor{CBI7}{watch} \\\cmidrule(lr){1-2}\cmidrule(lr){2-8}
\makecell[l]{The \textcolor{CBI1}{cross} is in Box Z, the \textcolor{CBI2}{ice} is in Box D,\\ the \textcolor{CBI3}{ring} is in Box F, the \textcolor{CBI4}{plane} is in Box Q,\\ the \textcolor{CBI5}{clock} is in Box X, the \textcolor{CBI6}{paper} is in Box I,\\ the \textcolor{CBI7}{engine} is in Box K.} & 
 \makecell[l]{Box Z\\ contains\\ the}& \textcolor{CBI2}{ice} & \textcolor{CBI3}{ring} & \textcolor{CBI3}{ring} & \textcolor{CBI5}{clock} & \textcolor{CBI6}{paper} & \textcolor{CBI7}{engine}
\\\hline
\end{tabular}
}
\caption{Attributes inferred by Llama2-7B as a result of directed activation patching along OI-PC in the OI subspace on the dataset of ``r: is\_in'', where color denotes the BI.}
\label{tab:ent_track}
\end{table*}

By projecting the activation matrix $M_r$ into the OI subspace, we have found a correlative evidence for the existence of the direction (i.e., OI-PC) that encodes OI information. However, it is possible that the OI information is encoded in the OI subspace but has no effect on LMs' binding behaviour.

In order to test if OIs are not only encoded in the OI subspace, but that these representations can be steered so as to swap the binding and change LM's output, in this section, we perform interventions to analyze the causality. That is, we want to find out if making interventions along OI-PC leads to a change in LM's binding computation. 

\subsection{Activation Patching}

Activation Patching (AP)~\cite{vig2020investigating,geiger2020neural,geiger2021causal,wang2022interpretability,stolfo2023mechanistic,heinzerling2024monotonic,2405.14860,hanna2024does} has been recently proposed to causally intervene computational graph of a LM so as to interpret the function of a target computational node (or edge). The process of AP usually involves preparing corrupted input, using its activations to replace the corresponding ones obtained from original input and analyzing the effect on model output. Different with the common AP setup, we realize AP by directly editing activations along a particular direction (i.e., along OI-PC), similar to the activation editing method of \cite{matsumoto2023tracing,heinzerling2024monotonic,2405.14860}. 

\subsection{Setting}
\textbf{Dataset} To explore the internal representation that enables binding, we adopt the entity tracking dataset~\cite{kim2023entity,prakash2024fine}. The dataset consists of English sentence describing a set of objects (here called attributes) located in a set of boxes with difference labels (here called entities), and the task is to infer what is contained by a given box. For instance, when a LM is presented with ``The coffee is in Box Z, the stone is in Box M, the map is in Box H, ... Box Z contains the'', the LM should infer the next token as ``coffee''. Each sample involves $7$ EA pairs. To evaluate the binding in various context, we also apply the templates shown in Table~\ref{tab:datat_temp} to generate other $5$ datasets with different relation, where $a_i$ and $e_i$ denotes the attribute and entity, and they are sampled from a fixed pool of $224$ one-token objects (e.g., ``dog'', ``corn'' and ``cookie'') and $523$ of one-token names (e.g., ``Alex'', ``Juli'' and ``Dan'') respectively. We sample $n=1000$ context from each dataset to run the following analysis.

\begin{table}[t]
\centering
\scalebox{0.9}{
\begin{tabular}{ll}
\hline
  & \textbf{Template} \\\cmidrule(l){1-2}
 $1$ & \makecell[l]{The $a_0$ is sold by person $e_0$, ..., the $a_i$ is ..., \\ $a_7$ is sold by person $e_7$. Person $e_i$ is selling the} \\\cmidrule(l){1-2}
 $2$ & \makecell[l]{The $a_0$ is applied by person $e_0$, ..., the $a_i$ is ..., \\ $a_7$ is applied by person $e_7$. Person $e_i$ applies the} \\\cmidrule(l){1-2}
 $3$ & \makecell[l]{The $a_0$ is moved by person $e_0$, ..., the $a_i$ is ..., \\ $a_7$ is moved by person $e_7$. Person $e_i$ moved the} \\\cmidrule(l){1-2}
 $4$ & \makecell[l]{The $a_0$ is brought by person $e_0$, ..., the $a_i$ is ..., \\ $a_7$ is moved by person $e_7$. Person $e_i$ brings the} \\\cmidrule(l){1-2}
 $5$ & \makecell[l]{The $a_0$ is pushed by person $e_0$, ..., the $a_i$ is ..., \\ $a_7$ is pushed by person $e_7$. Person $e_i$ pushes the} \\\cmidrule(l){1-2}
 \hline
\end{tabular}
}
\caption{Templates of Dataset.}
\label{tab:datat_temp}
\end{table}

\textbf{Metrics} We apply two evaluation metrics: logit difference and logit flip. The logit difference metric is introduced in \citet{wang2022interpretability}, which calculates difference in logits of a target token between original and intervened setting. The "logit flip" accuracy metric is introduced by \citet{geiger2022inducing}, which represents the proportion of candidate tokens in model output after a causal intervention.  

\begin{figure*}[htb]
    \centering % <-- added
\begin{subfigure}{0.3\textwidth}
  \includegraphics[width=\linewidth]{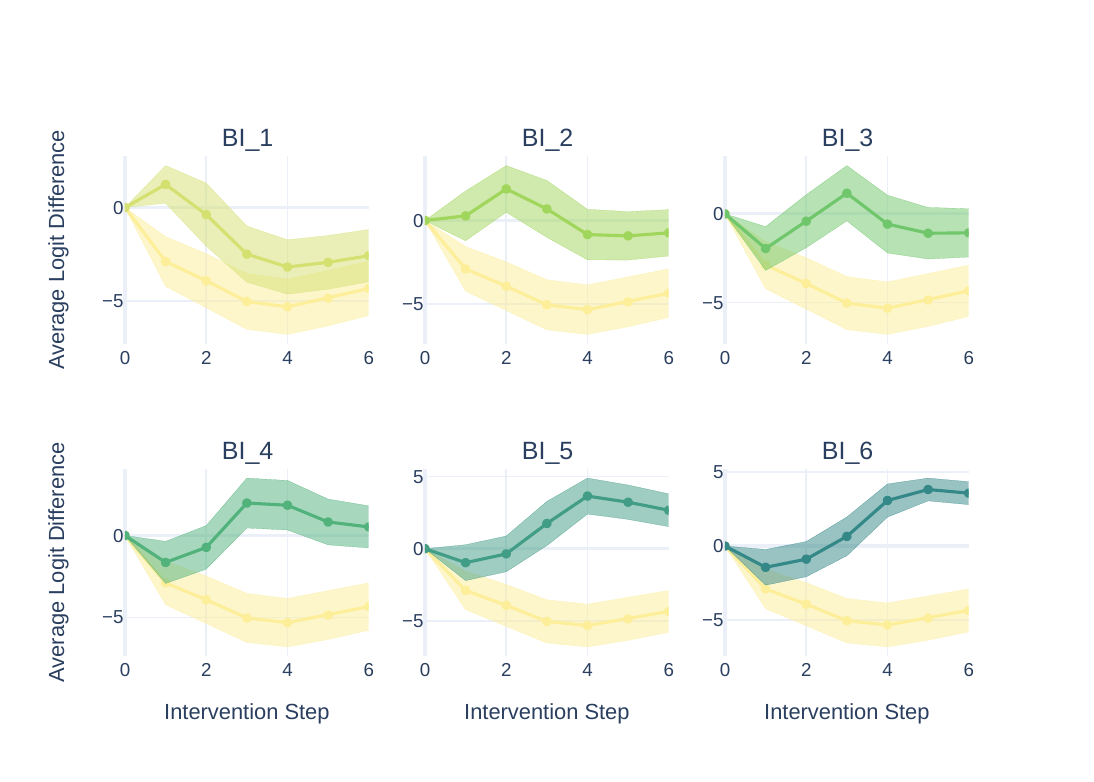}
  \caption{r: is\_in}
  \label{fig:l1}
\end{subfigure}\hfil % <-- added
\begin{subfigure}{0.3\textwidth}
  \includegraphics[width=\linewidth]{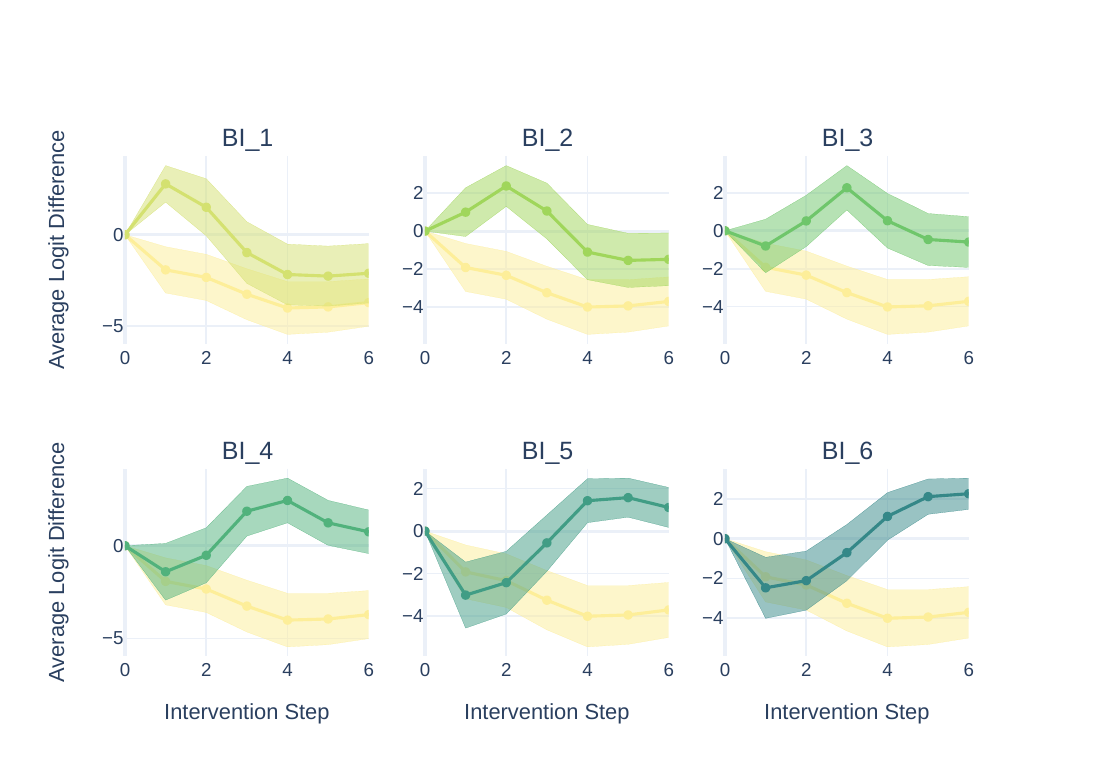}
  \caption{r: sell}
  \label{fig:l2}
\end{subfigure}\hfil % <-- added
\begin{subfigure}{0.3\textwidth}
  \includegraphics[width=\linewidth]{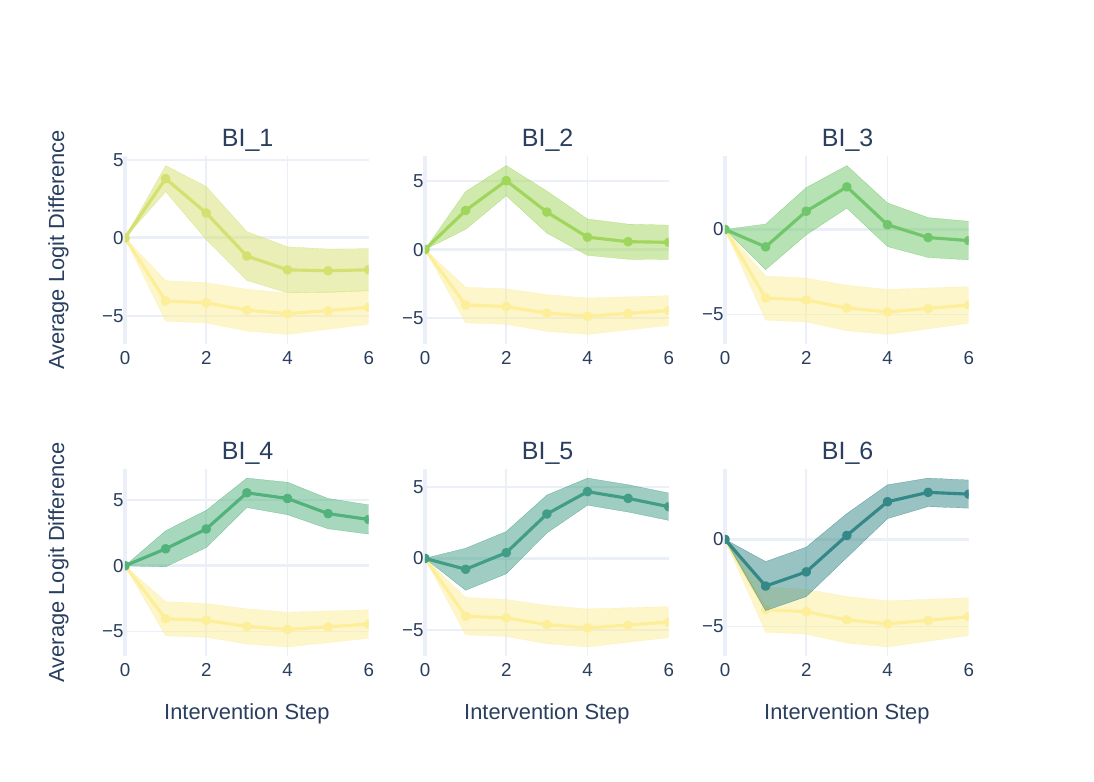}
  \caption{r: apply}
  \label{fig:l3}
\end{subfigure}

\medskip
\begin{subfigure}{0.3\textwidth}
  \includegraphics[width=\linewidth]{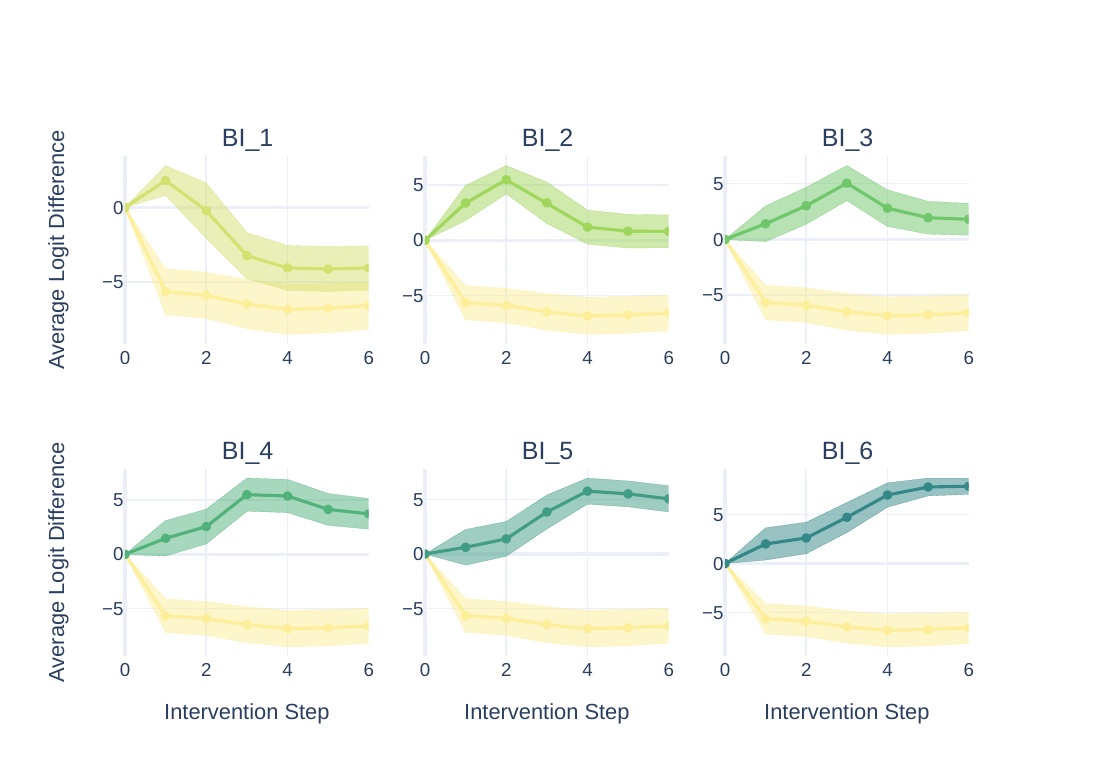}
  \caption{r: move}
  \label{fig:l4}
\end{subfigure}\hfil % <-- added
\begin{subfigure}{0.3\textwidth}
  \includegraphics[width=\linewidth]{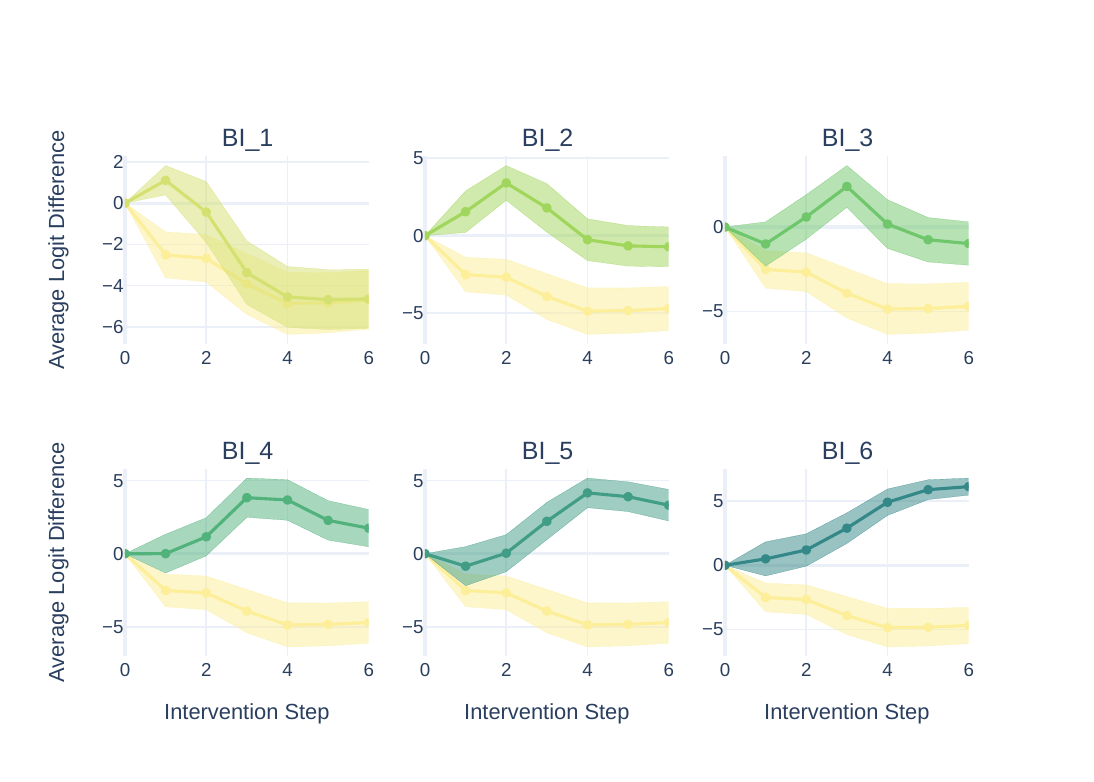}
  \caption{r:bring}
  \label{fig:l5}
\end{subfigure}\hfil % <-- added
\begin{subfigure}{0.3\textwidth}
  \includegraphics[width=\linewidth]{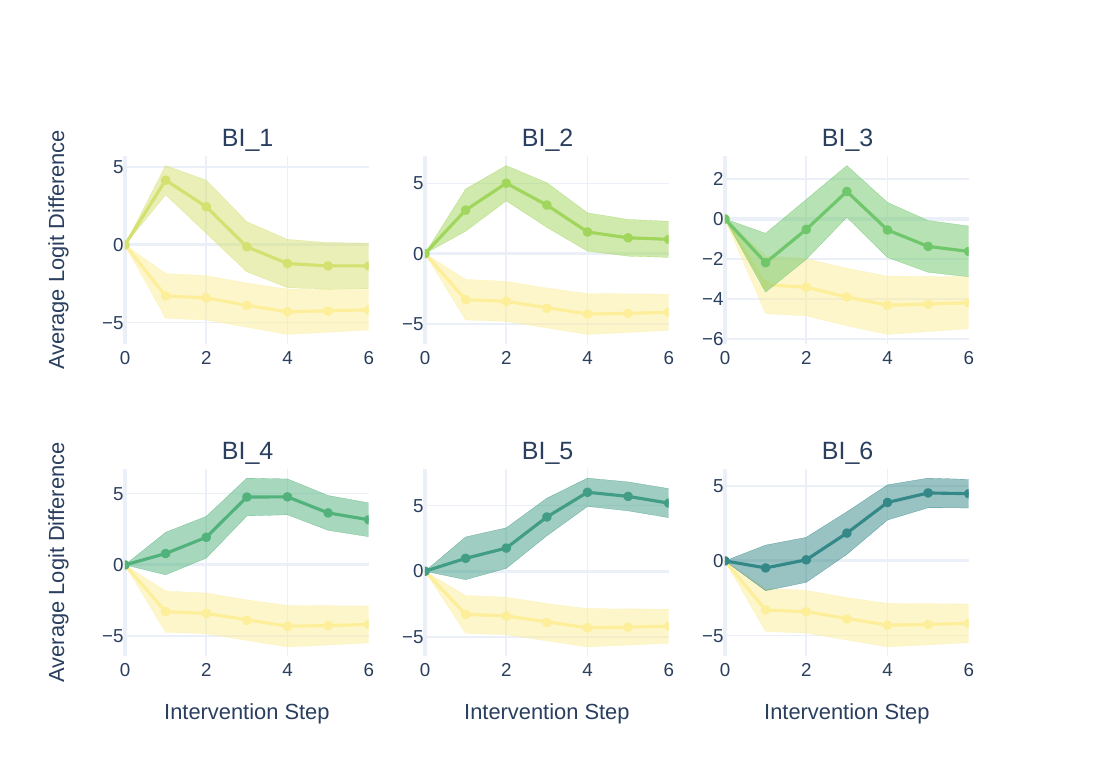}
  \caption{r: push}
  \label{fig:l6}
\end{subfigure}
\caption{Logit Difference (LD) for OI-PC based intervention across datasets on Llama2-7B, where x axis denotes the number of intervention steps on $e_0$, y axis does the LD, BI\_i represents each target attribute and the light yellow bottom line indicates the LD of original attribute (i.e., $a_0$). Here, $l = 8$, $v = 2.5$,  and $\alpha = 3.0$.}
\label{fig:ld}
\end{figure*}

\subsection{Results: Direct Editing OI Subspace}
\label{sec:ap_subspace}
We hypothesize that LMs encode OI information into a low-rank subspace that causally affect BI information and thus binding behaviour as well. Therefore, we wonder if a LM changes the binding behavior, when adding a particular value $v$ (called step hereafter) along the OI-PC mentioned in Section~\secref{sec:subspace}~\footnote{where PC2 is assigned a fixed value, which is a hyperparameter.}. For example, if we add one unit of $v$ on the OI-PC of $e_0$, the LM will reset its BI as $1$ and bind attribute $a_1$ to the entity based on the similarity of BI so that infers the $a_1$ as the attribute of $e_0$ instead of the original $a_0$. Similarly, adding two units of $v$ will make the LM infer $a_2$ for $e_0$, and so on. We intervene via the Equation~\ref{eq:pc_patch}, where $\mathbf{x}_{0,l}$ is the original activation of $e_0$ (i.e., the leftmost entity) in layer $l$, $\mathbf{x}_{0,l}^*$ is the intervened activation, $B_r$ is the OI subspace projection matrix mentioned in Section \secref{sec:subspace}, $\alpha$ is a hyper-parameter to scale the effect of intervention and $\beta$ ($0 \leq \beta \leq 6$) denotes the number of steps.
\begin{align}\label{eq:pc_patch}
\mathbf{x}_{0,l}^* = \mathbf{x}_{0,l} + \alpha B_r^T(B_r\mathbf{x}_{0,l} + \beta v)
\end{align}

Table~\ref{tab:ent_track} lists several examples under the OI subspace intervention on the entity tracking dataset~\cite{kim2023entity,prakash2024fine}. We also list the examples from other datasets in Appendix~\secref{sec:case_study}. We can see that when adding $1$ step along OI-PC, the model selects ``stone'' for entity ``Z'' instead of its original attribute ``coffee''. Similarly, when the step is doubled, the model will select attribute ``map'' for the entity, and so on. 
This indicates that changing the value along OI-PC can induce the swap of attribute. In addition, we notice that some attributes are repeated or skipped after the intervention. The reason is that, as shown in Figure~\ref{fig:layer_wise_pca}, OI is represented as a continuous range (e.g., $[a_0, b_0], [a_1, b_1], [a_2, b_2], ...$) on OI-PC, implying the points in the same range share OI, and if a sample (e.g., its OI-PC is $s_i$) is still in the range (e.g., $a_0<(s_i+v)<b_0$) or skip its neighbouring range (e.g., $a_2<(s_i+v)<b_2$) after the intervention (e.g., $s_i+v$), its OI will be the same or added by $2$, and thus the binding information will change accordingly.

Besides the qualitative analysis, we also conduct quantitative analysis for the causality between the OI subspace based AP and the binding behaviour of LMs. We plot mean-aggregated effect of the OI-PC based AP across multiple datasets in Figure~\ref{fig:ld}. Figure~\ref{fig:ld} indicates how the Logit Difference (LD) of each attribute changes as the step increases. We can observe that as the number of steps increases, LD of the original attribute decreases. In contrast, LD of other attributes gradually increase until a certain point and then gradually decrease. Given a candidate attribute, its LD peak roughly corresponds to the number of steps that is equal to its BI. For instance, when adding $3$ steps, the points of BI\_3 (i.e., attributes of BI$=3$) achieve the highest LD score. This indicates that by adjusting the value along the OI-PC, we can adjust BI information and thus increase the logit score of the corresponding attribute.  

\begin{figure*}[htb]
    \centering % <-- added
\begin{subfigure}{0.30\textwidth}
  \includegraphics[width=\linewidth]{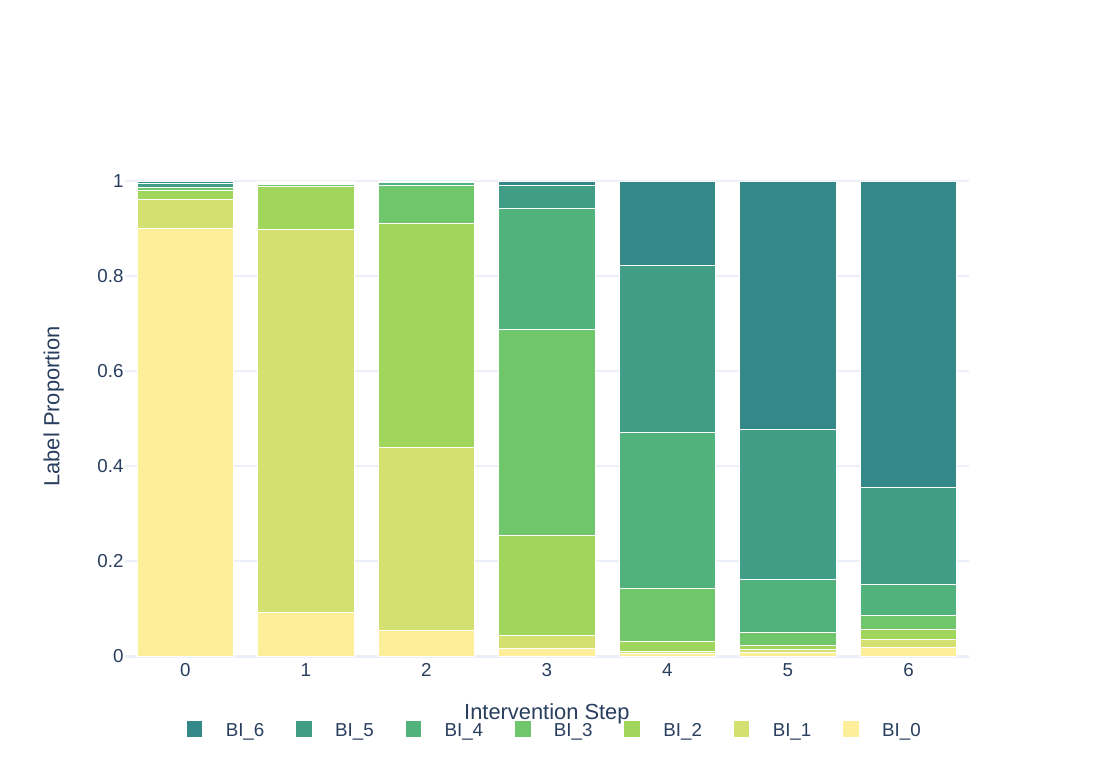}
  \caption{r: is\_in}
  \label{fig:f1}
\end{subfigure}\hfil % <-- added
\begin{subfigure}{0.30\textwidth}
  \includegraphics[width=\linewidth]{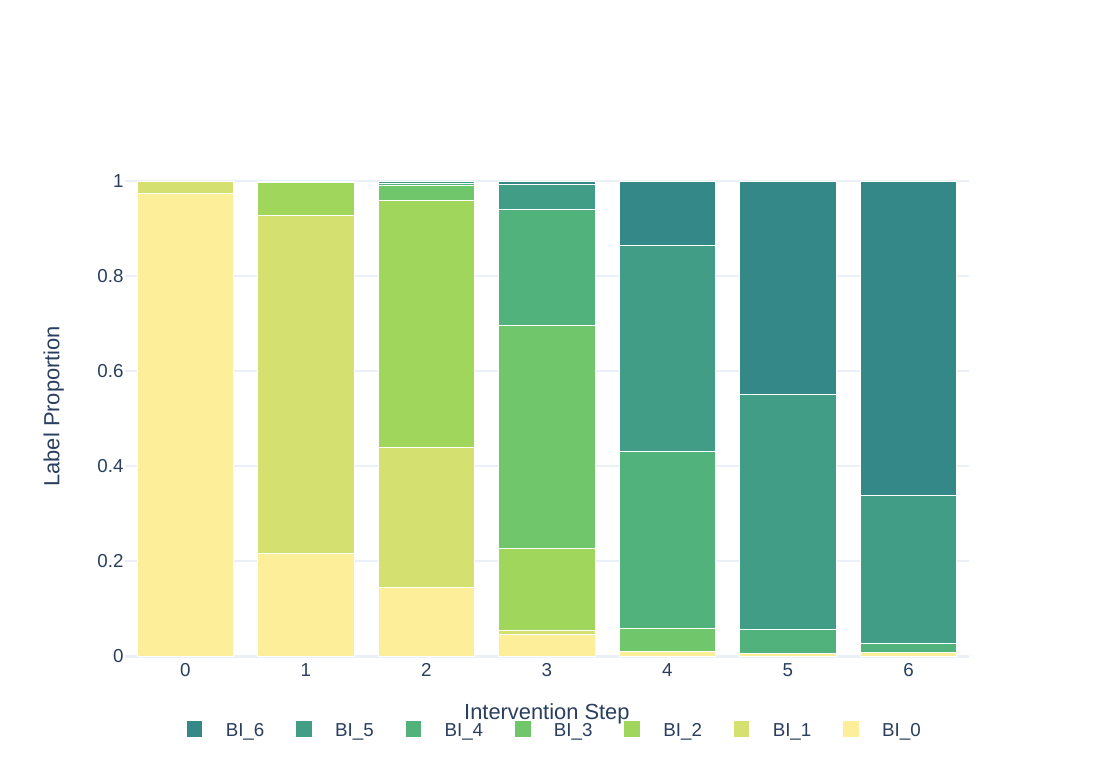}
  \caption{r: sell}
  \label{fig:f2}
\end{subfigure}\hfil % <-- added
\begin{subfigure}{0.30\textwidth}
  \includegraphics[width=\linewidth]{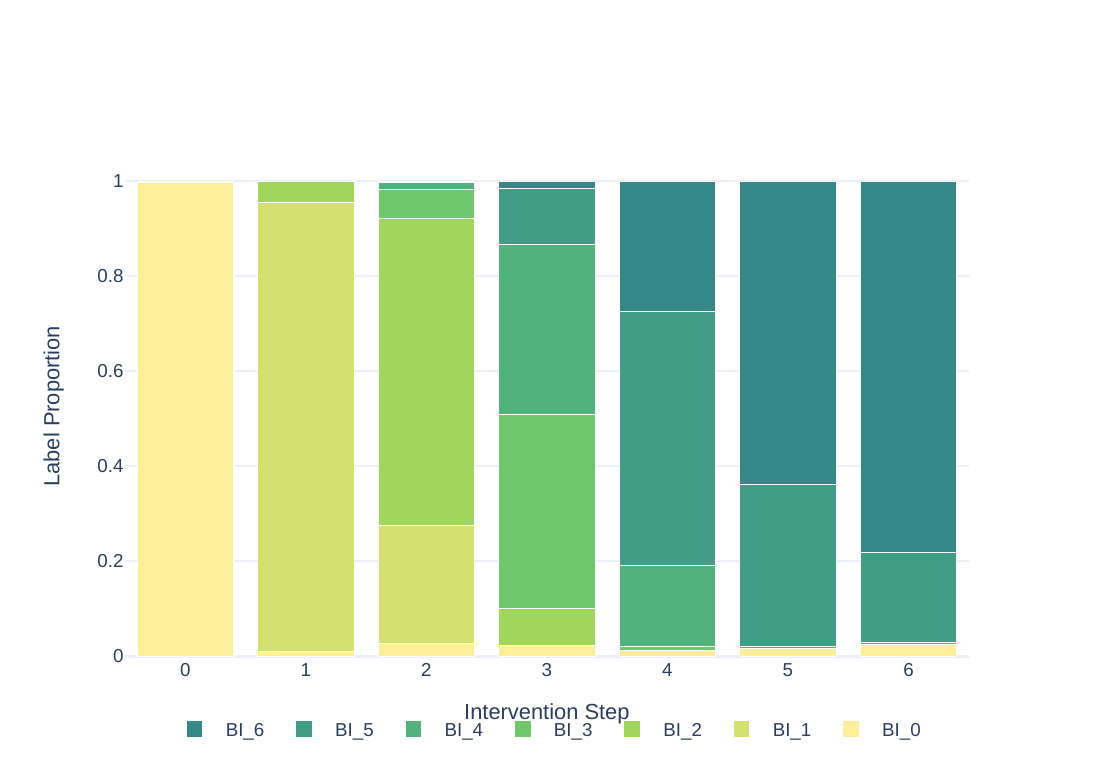}
  \caption{r: apply}
  \label{fig:f3}
\end{subfigure}

\medskip
\begin{subfigure}{0.30\textwidth}
  \includegraphics[width=\linewidth]{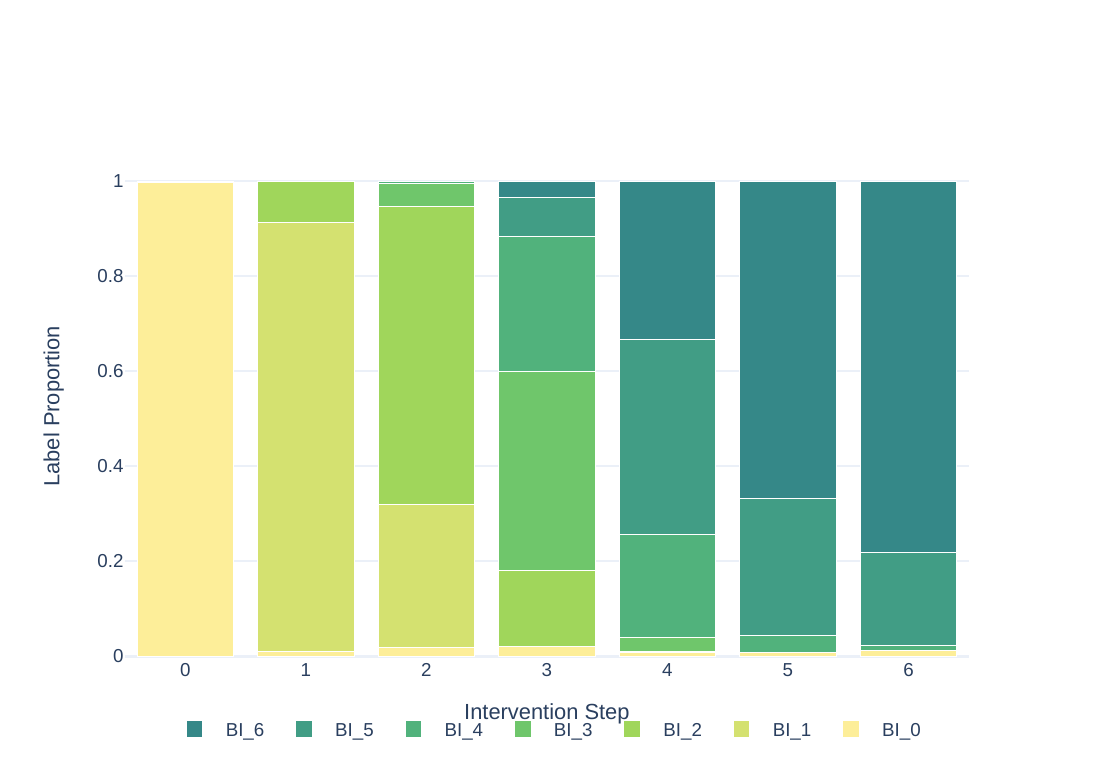}
  \caption{r: move}
  \label{fig:f4}
\end{subfigure}\hfil % <-- added
\begin{subfigure}{0.30\textwidth}
  \includegraphics[width=\linewidth]{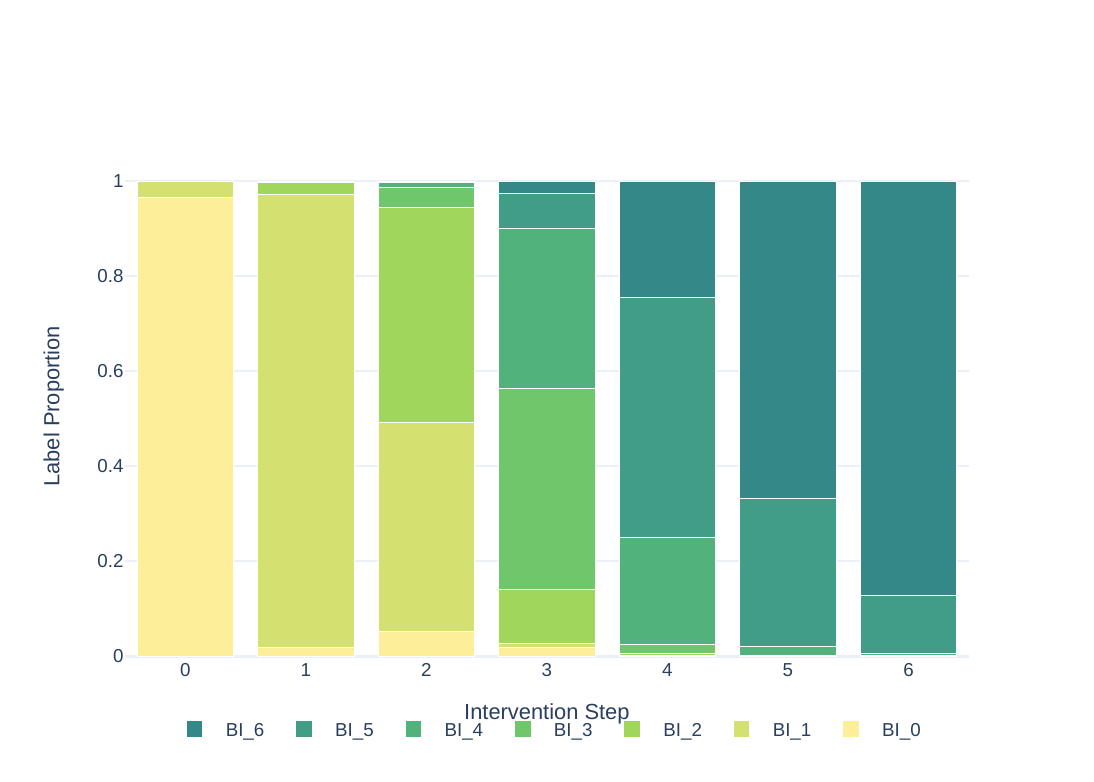}
  \caption{r: bring}
  \label{fig:f5}
\end{subfigure}\hfil % <-- added
\begin{subfigure}{0.30\textwidth}
  \includegraphics[width=\linewidth]{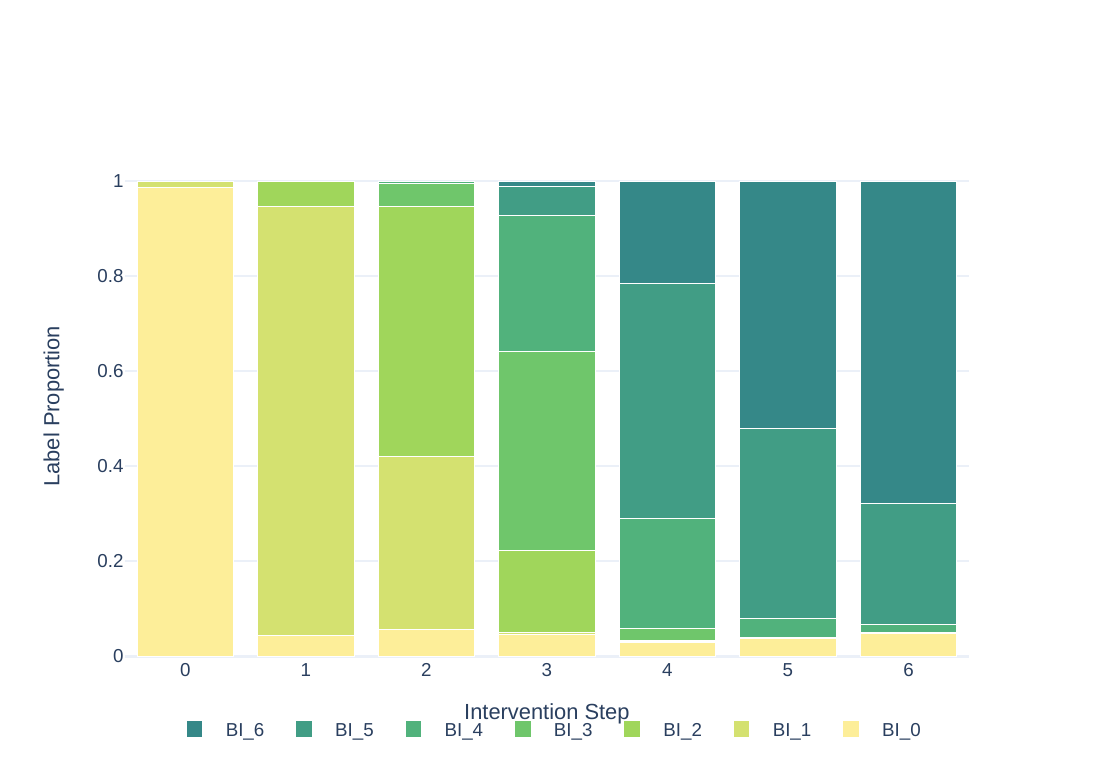}
  \caption{r: push}
  \label{fig:f6}
\end{subfigure}
\caption{Logit flip for OI-PC based intervention across datasets on Llama2-7B, where x axis denotes the number of intervention steps on $e_0$, y axis does the proportion of each inferred attribute in model output.}
\label{fig:lf}
\end{figure*}

Similarly, Figure~\ref{fig:lf} illustrates the relation between the number of steps and the logit flip, which gauges the percentage of the predicted attributes under an intervention. Figure~\ref{fig:lf} shows that as the step increases, the proportion bar becomes darker, it means that the model promotes the proportion of the corresponding attribute in its inference. For instance, when adding $3$ step on the subspace, the $a_3$ (i.e., BI\_3) becomes the major of the answers. This proves that the OI-PC based interventions can causally affect BI information as well as the computation of Binding in a LM. See Appendix~\secref{sec:llama3} for the results on Llama3-8B, Appendix~\secref{sec:other_llm} for the results on Qwen1.5-7B and Pythia-6.9B, and Appendix~\secref{sec:ap_float} for the results on Float-7B.

\subsection{Results: Activation Steering on OI Subspace}
\begin{figure}
\centering
\begin{subfigure}{.25\textwidth}
  \centering
  \includegraphics[width=\columnwidth]{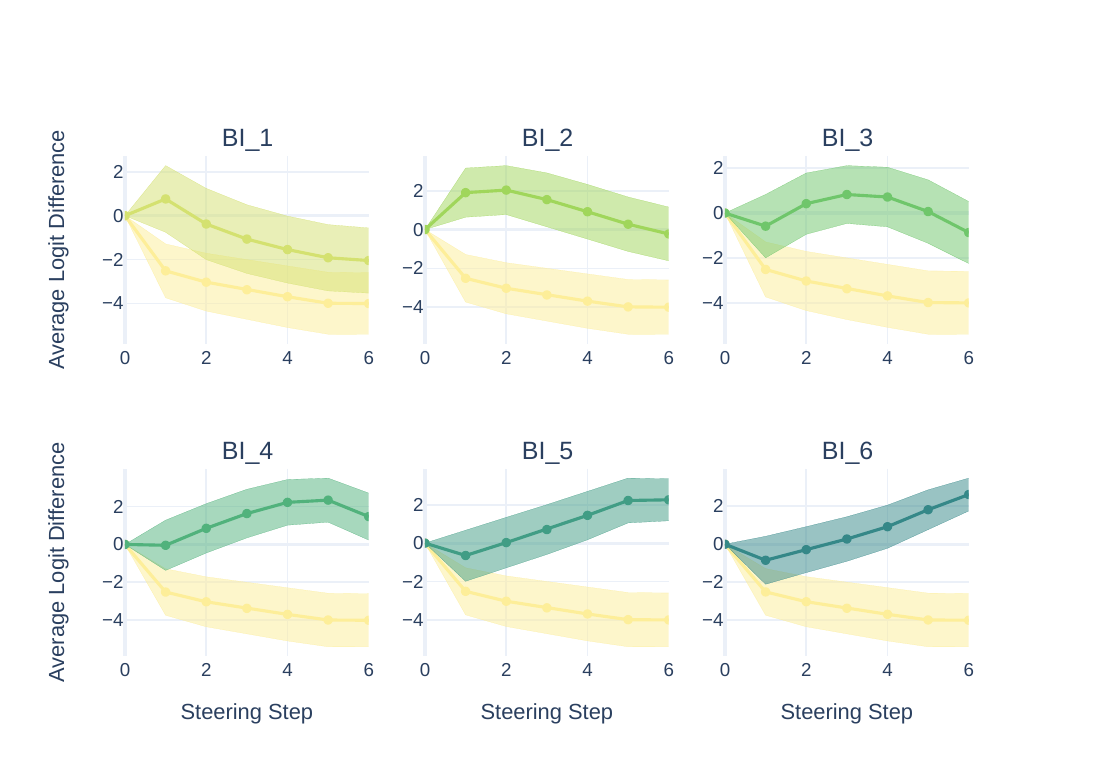}
\end{subfigure}%
\begin{subfigure}{.25\textwidth}
  \centering
  \includegraphics[width=\columnwidth]{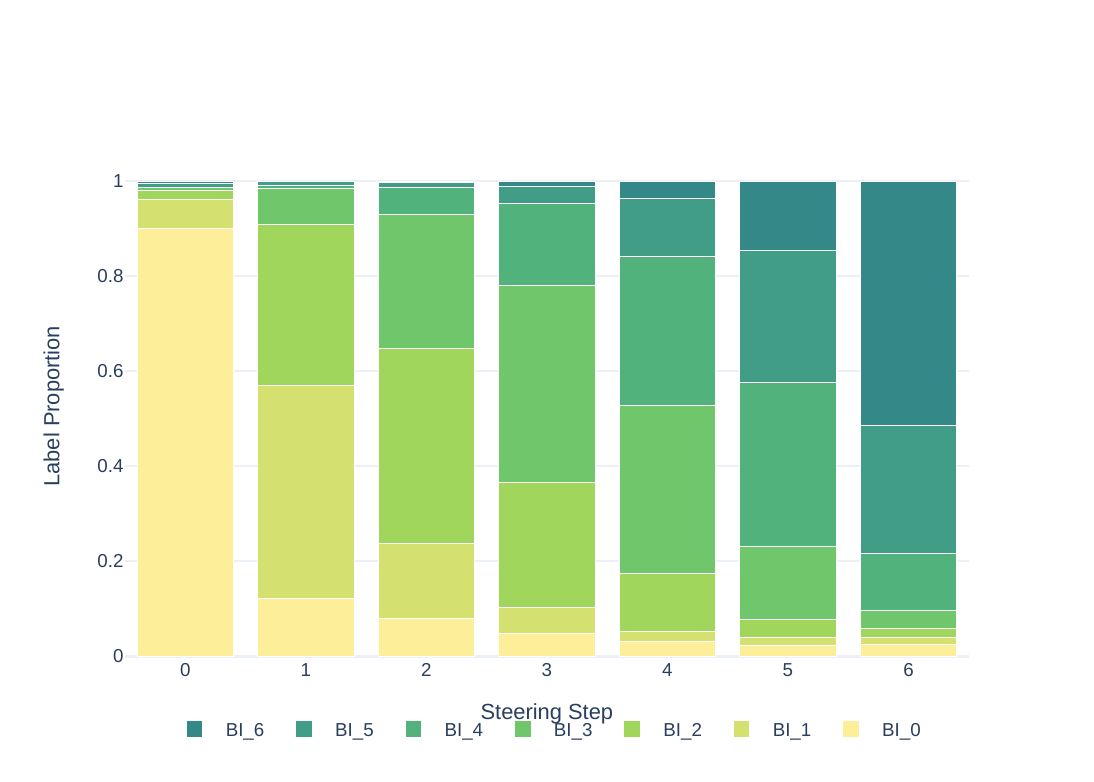}
\end{subfigure}
\caption{Logit Difference and Logit Flip for activation steering on the entity tracking dataset (i.e., r: is\_in), where x axis represents the intervention of $\mathbf{s}_{0 \rightarrow bi}$.}
\label{fig:st_ld_lf}
\end{figure}

Inspired by the research on Activation Steering (AS)~\cite{turner2023activation}, we apply an AS method to verify the importance of the OI subspace on LM's binding behaviour. Specifically, we use the following Equation~\ref{eq:as_patch} to extract a subspace steering vector $\mathbf{s}_{0 \rightarrow bi}$, which is proposed to swap BI from $0$ to $bi$, where $n$ is the number of target entities, $\mathbf{x}_{bi,l}^i$ represents the activation of entity $e_i$ from layer $l$, and its BI is $bi$. We intervene via Equation~\ref{eq:as_patch1} and assume that by adding $\mathbf{s}_{0 \rightarrow bi}$ on the original activation $\mathbf{x}_{0,l}$, we can increase the LD and the proportion of the attribute $a_{bi}$. Figure~\ref{fig:st_ld_lf} shows the results on the entity tracking dataset~\cite{kim2023entity,prakash2024fine}. (Appendix~\secref{sec:as_all} shows the results on other datasets) These results indicate that AS can achieve the similar tendency as the direct value intervention mentioned in Section~\secref{sec:ap_subspace}. For instance, adding $\mathbf{s}_{0 \rightarrow 3}$, which is supposed to swap BI from $0$ to $3$, can increase the LD of $a_3$ and its proportion in the predicted answers. The consistent tendency with the results of the direct subspace editing, shown Figure~\ref{fig:ld} and Figure~\ref{fig:lf}, further illustrates that the discovered OI subspace can causally affect BI information and binding behaviour. Therefore, OI subspace plays an important role when LMs perform in-context binding computation.

\begin{align}\label{eq:as_patch}
\mathbf{s}_{0 \rightarrow bi} = \frac{1}{n}\sum_{i=1}^{n}(B_r\mathbf{x}_{bi,l}^i - B_r\mathbf{x}_{0,l}^i)
\end{align}
\begin{align}\label{eq:as_patch1}
\mathbf{x}_{0,l}^* = \mathbf{x}_{0,l} + \alpha B_r^T\mathbf{s}_{0 \rightarrow bi}
\end{align}

\subsection{OI Subspace and Position}

\begin{table}[t]
\centering
\scalebox{0.9}{
\begin{tabular}{ll}
\hline
  & \textbf{Input (original)} \\\cmidrule(l){1-2}
 $C_1$ & \makecell[l]{$a_0^{p0} \, r \, e_0^{p1}, \, a_1^{p2} \, r \, e_1^{p3}, \, a_2^{p4} \, r \, e_2^{p5}. \quad e_1^{p3} \ r^{-1} \ ?$} \\\cmidrule(l){1-2}
 $C_2$ & \makecell[l]{$a_0^{p0} \, r \, e_0^{p1}, \, a_1^{p2} \, r \, e_1^{p3}, \, a_2^{p4} \, r \, e_2^{p5}. \quad e_2^{p5} \ r^{-1} \ ?$} \\\cmidrule(l){1-2}
  & \textbf{Input (with pseudo)} \\\cmidrule(l){1-2}
 $C'_1$ & \makecell[l]{$a_{*0}^{p0} \, r \, e_{*0}^{p1}, \, a_{*0}^{p2} \, r \, e_{*0}^{p3}, \, a_1^{p4} \, r \, e_1^{p5}. \quad e_1^{p5} \ r^{-1} \ ?$} \\\cmidrule(l){1-2}
 $C'_2$ & \makecell[l]{$a_{*0}^{p0} \, r \, e_{*0}^{p1}, \, a_1^{p2} \, r \, e_1^{p3}, \, a_2^{p4} \, r \, e_2^{p5}. \quad e_2^{p5} \ r^{-1} \ ?$} \\\cmidrule(l){1-2}
 \hline
\end{tabular}
}
\caption{Simplified expression of original inputs and the one modified with pseudo relation, which is proposed to equalize PI for PCA analysis, where $a_{0}^{p1} \, r \, e_{0}^{p2}$ represents a relation such as ``the apple is in Box C'', and $e_0^{p2}$ denotes an entity with OI of $0$ and PI of $p2$, $e_2^{p5} \ r^{-1} \ ?$ denotes the query on entity $e_1$, such as ``Box C contains the''.}
\label{tab:ctx_pseu}
\end{table}
\begin{table}[t]
\centering
\scalebox{0.9}{
\begin{tabular}{l}
\hline
\textbf{Input (original)} \\\cmidrule(l){1-1}
\makecell[l]{``The apple is in Box E, the bell is in Box F, ...''} \\\cmidrule(l){1-1}
\textbf{Input (with filler words)} \\\cmidrule(l){1-1}
\makecell[l]{``\textit{I will find out that} the apple is in Box E, the bell is\\ in Box F, ...``} \\\cmidrule(l){1-1}
\hline
\end{tabular}
}
\caption{An example of the dataset with filler words ``\textit{I will find out that}''.}
\label{tab:ctx_fil}
\end{table}

In this section, we discuss the relationship between the OI subspace and Positional Information (PI), which is namely the $postion\_ids$ of input tokens. As mentioned in Section \secref{sec:ap_subspace}, the discovered subspace can causally determine BI information. Therefore, direct intervention on the subspace can swap the answer of a LM. However, one counter hypothesis is that the subspace is not used for storing OI information but the PI of attributes, and thus the direct intervention merely changes the PI of answer token so that the LM applies the new PI to generate corresponding token. In other words, the OI space (or OI-PC) might have high correlation with PI but not OI.
%Regarding the relationship between BI and PI, \citet{feng2023language} found that even when PI of attributes is swapped, the model still makes correct predictions, thus confirming the independence between BI and PI. Based on this finding, we go one step further and illustrate the independence between the OI subspace and PI. 

To prove the independence between OI subspace and PI, we create three datasets, the first one is by extending the original dataset with pseudo relation, as shown in Table~\ref{tab:ctx_pseu}, the second one is by prefixing the original dataset with filler words(e.g., ``It can be seen that'') and third one is by inserting a sequence of interjection(e.g., ``ah, ah, ...''), as listed in Table~\ref{tab:ctx_ij}.

\textbf{New Dataset with Pseudo Relation.} In Table~\ref{tab:ctx_pseu}, $a_{*0}^{p0} \, r \, e_{*0}^{p1}$ refers to a pseudo relation, which is a fixed expression, such as ``the PC is in Box Z''. The pseudo relation is applied to adjust the PI while keeping the OI. For instance, in Table~\ref{tab:ctx_pseu}, adding one or two $a_{*0}^{p0} \, r \, e_{*0}^{p1}$ before $a_{1} \, r \, e_{1}$ (i.e., $C'_2$ and $C'_1$) does not affect the OI of $e_1$ but its PI, because $e_1$ is still the second unique entity from the left (i.e., its ordering index 0I$=1$), but its PI is $p3$ and $p5$ respectively. Using the pseudo relation, we create the data in a manner that the target entity for activation analysis (e.g., $e_1^{p5}$ and $e_2^{p5}$) have the same PI but different OI, such as $C'_1$ and $C'_2$ in Table~\ref{tab:ctx_pseu}.

\begin{figure}[t]
\centering
\includegraphics[width=8.0cm]{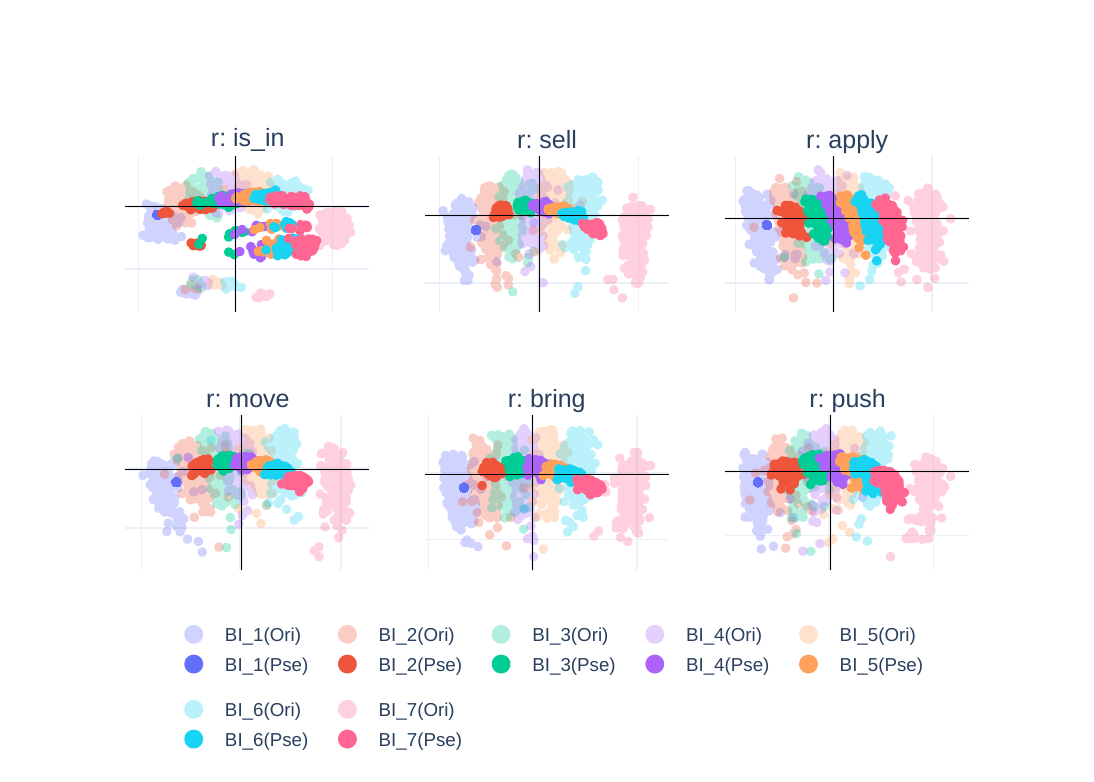}
\caption{Embedding visualization for activation with equalized PI, where ``Ori'' denotes the distribution of original dataset, while ``Pse'' denotes the distribution of the new dataset with pseudo relation.}
\label{fig:bi_pos_pca}
\end{figure}

We apply the method mentioned in Section~\secref{sec:subspace} on the set of activations $\{ \overrightarrow{e_1^{p5}}, \overrightarrow{e_2^{p5}}, ... \}$, where $\overrightarrow{e_1^{p5}}$ denotes the activation of $e_1$ in query (i.e., $e_1^{p5} \ r^{-1} \ ?$), so as to capture the OI difference and exclude the PI difference, because they share the same PI (i.e., $P5$) but different OI (i.e., $1$, $2$, ...). Then we compare its OI subspace with the original one (e.g., $\{ \overrightarrow{e_1^{p2}}, \overrightarrow{e_2^{p5}}, ... \}$) to analyze how the distribution of OI subspace changes after removing the PI variance. Figure~\ref{fig:bi_pos_pca} visualizes the OI subspace distribution, where the light colored points denote the original distribution, and the dark ones are from the new one with equalized PI. We can observe that after removing the PI difference, the distribution is still similar to the original one that there is a clearly visible direction along which OI increases. This illustrates that our PCA based method can capture OI information, that is, along the direction of OI-PC, and it does not causally depend on PI.

\begin{figure}[!t]
\centering
\includegraphics[width=4.5cm]{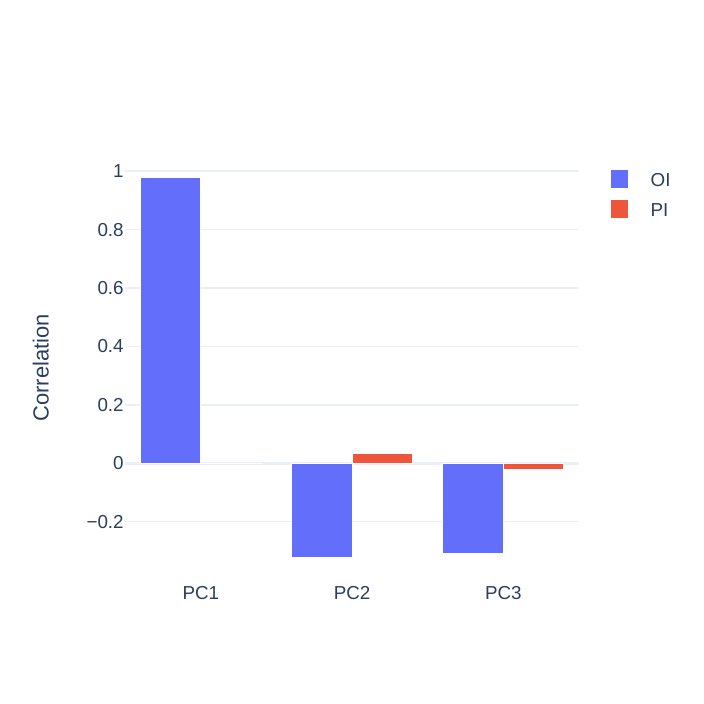}
\caption{Spearman's rank correlation between OI-PC and PI (or OI), where ``PC$i$'' denotes the $i$-th PC of the OI subspace and ``PI'' is the length of FW.}
\label{fig:cor_pi_bi}
\end{figure}

\textbf{New Dataset with Filler Words.} The dataset is created by adding Filler Words (FW) with various length, such as ``OK'', ``I see that'' and ``There is no particular reason'', in front of the entity tracking dataset~\cite{kim2023entity,prakash2024fine}, as shown in Table~\ref{tab:ctx_fil}. 
Since the length (i.e., the number of tokens) of FW directly changes the PI of its following entities and attributes without affecting their OIs, we take the length as the measure of intervention on PI and apply Spearman's rank correlation $\rho$ to calculate the correlation between the length (denoted as PI) and the OI-PC value. Figure~\ref{fig:cor_pi_bi} shows $\rho$ between PI and OI-PC as well as between OI and OI-PC. We can observe that OI-PC has high $\rho$ with OI but almost zero $\rho$ with PI, indicating that the discovered OI-PC is highly correlated with OI information but independent on PI. Therefore, the OI-PC does not simply encode absolute token position. See Appendix~\secref{sec:ij_all} for the analysis on the \textbf{New Dataset with Interjections}.

\subsection{Consistency of OI Subspace}

The BI mechanism~\cite{feng2023language} mentions that a LM represents a related EA pair through the consistency of their BI information. This naturally raises a research question that if there is the consistency between OI-PC of entities (e.g., OI-PC of ``Box Z'' in Sample~\ref{ex:ent_track1}) and of their corresponding attributes~\footnote{The OI-PC of attribute is extracted from the query of attribute such as ``(context) The coffee is in''.} (e.g., OI-PC of``coffee'').

\begin{figure}[t]
\centering
\begin{subfigure}{.23\textwidth}
  \centering
  \includegraphics[width=\columnwidth]{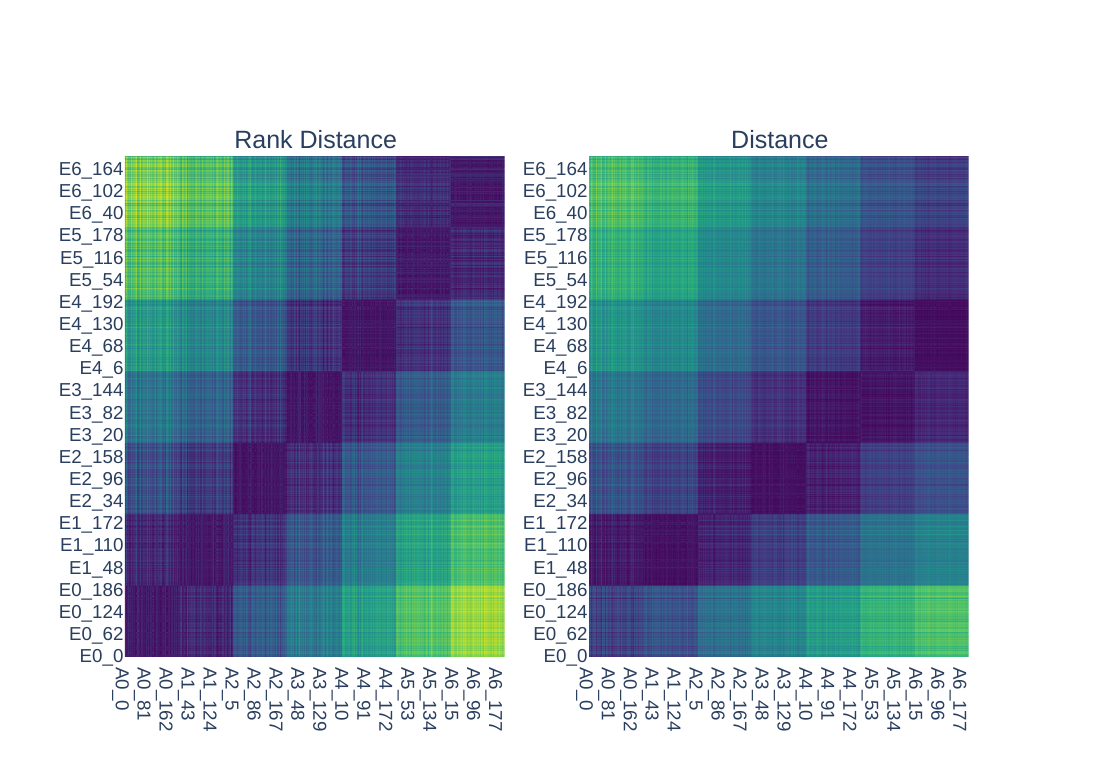}
  \caption{7A-7E}
\end{subfigure}
\begin{subfigure}{.23\textwidth}
  \centering
  \includegraphics[width=\columnwidth]{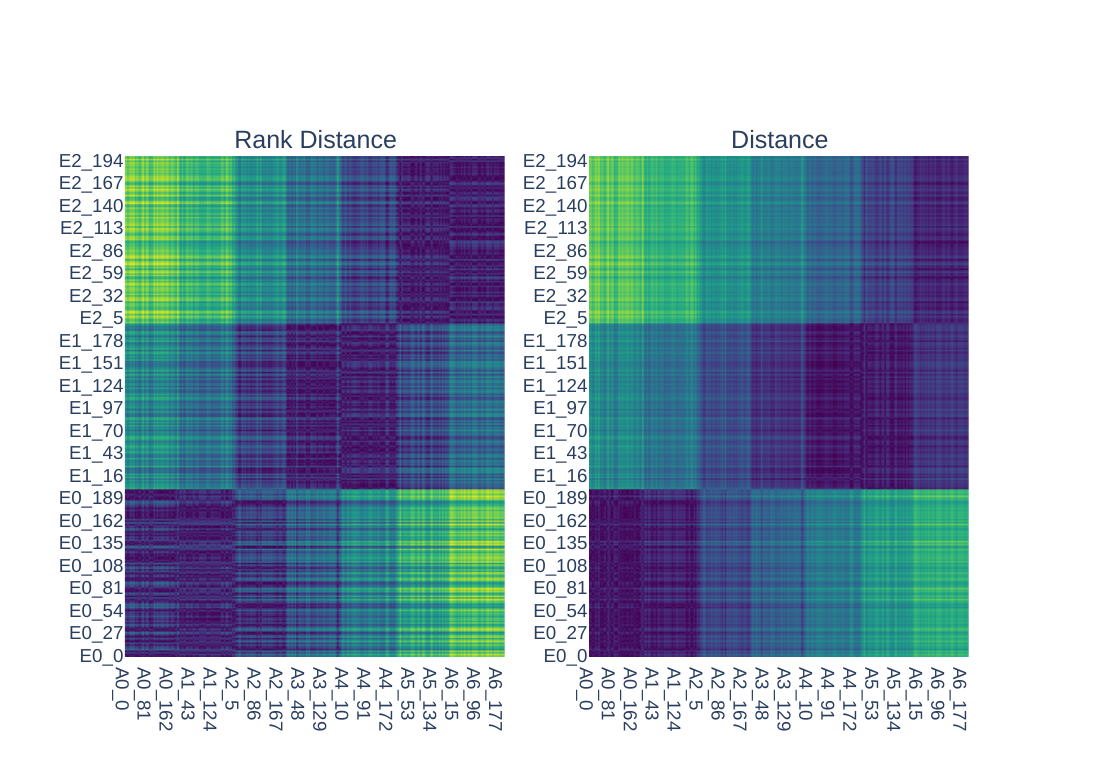}
  \caption{7A-3E}
\end{subfigure}
\begin{subfigure}{.23\textwidth}
  \centering
  \includegraphics[width=\columnwidth]{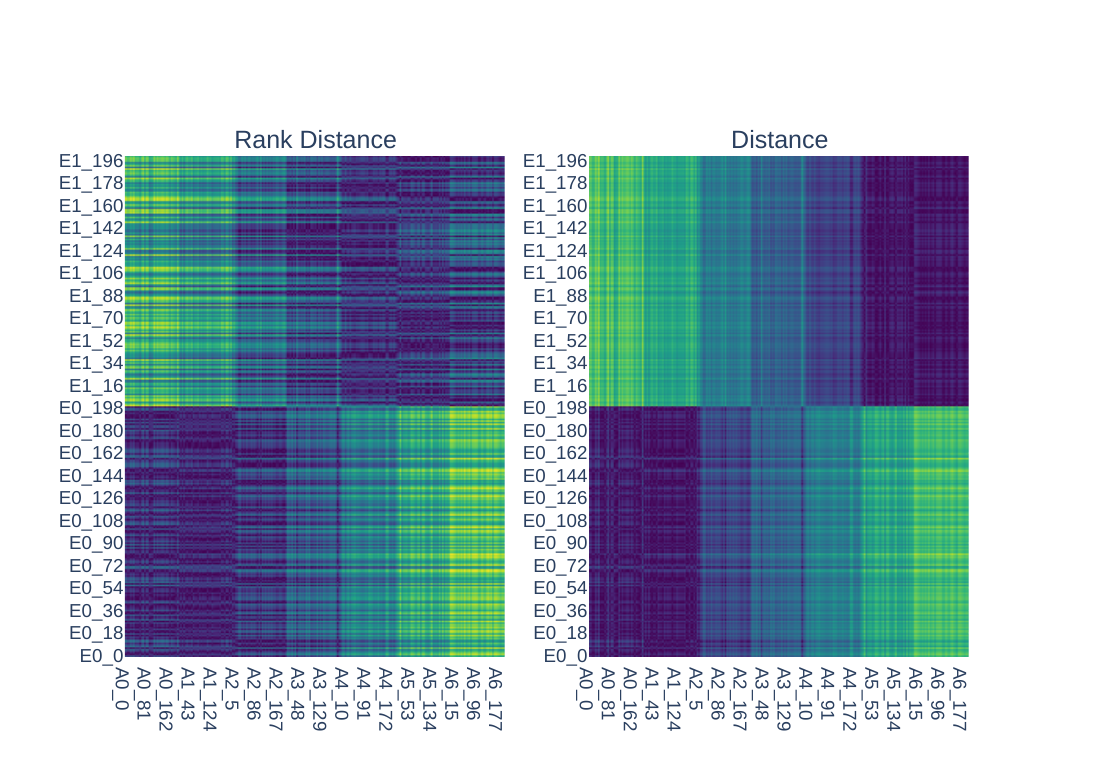}
  \caption{7A-2E}
\end{subfigure}%
\begin{subfigure}{.23\textwidth}
  \centering
  \includegraphics[width=\columnwidth]{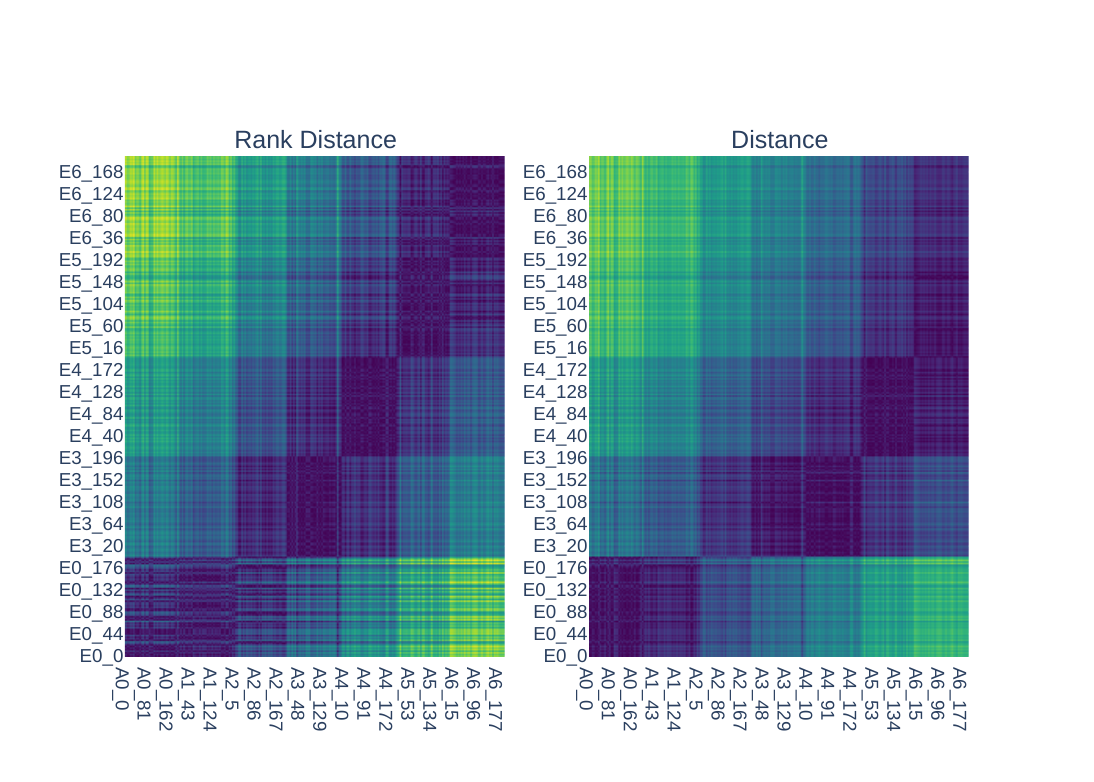}
  \caption{7A-5E}
\end{subfigure}
\caption{\label{fig:pca_ae}: OI-PC based distance heat map, where ``Rank Distance'' denotes the distance based on the rank of OI-PC value, ``A$i$\_'' (or ``E$i$\_'') is a sample of Attribute (or Entity) with OI$=i$, each cell represents the distance between a corresponding EA pair, and the darker the color, the smaller the distance is.}
\label{fig:ae_dis}
\end{figure}

\begin{figure}[t]
\centering
\includegraphics[width=6.0cm]{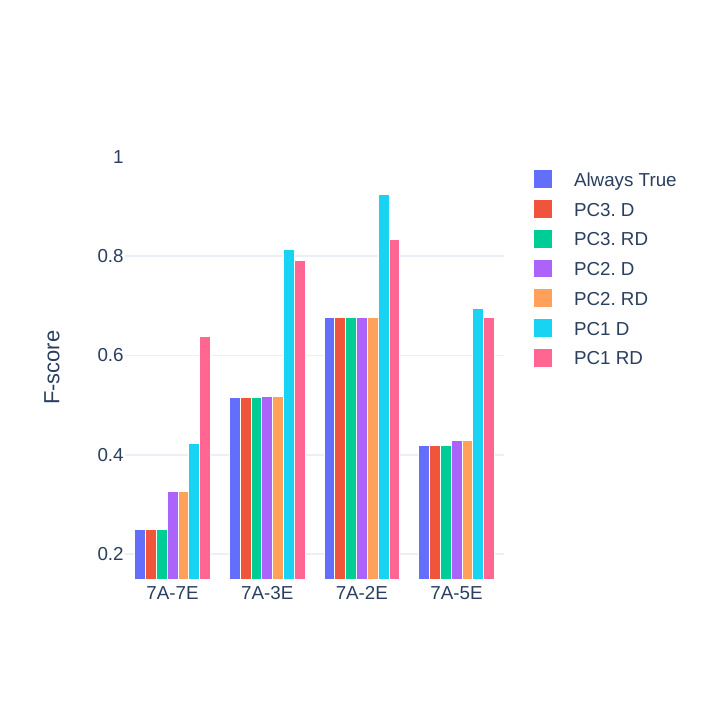}
\caption{OI-PC distance based classification results, where ``Always True'' denotes the baseline that predicts all candidates as True, ``PC$i$ D'' denotes the Hamming distance on $i$-th PC of the OI subspace, and ``PC$i$ RD'' denotes the distance calculated by the rank of PC$i$'s value among samples.}
\label{fig:f1_ae}
\end{figure}

To analyze the consistency of OI-PC between a related EA pair, we prepare four alternative datasets with various binding pattern as shown in Table~\ref{tab:ctx_mul}. We use the distance on OI-PC to measure consistency and show partial result of mutual OI-PC based distance in Figure~\ref{fig:ae_dis}. We can observe that potentially related EA pairs (e.g., ``E0\_0'' and ``A0\_0'') tend to have lower OI-PC based distance than arbitrary pairs in all binding patterns, indicating that to some extent, OI-PC based distance could be seen as an important feature to represent binding. To further illustrate its significance, we attempt to classify related EA pairs only relying on their OI-PC based distances. Specifically, we search an optimal threshold value from a development set, and which is applied to classify potentially related EA pairs in a testing set. The results are shown in Figure~\ref{fig:f1_ae}. We can observe that the performance of the OI-PC (i.e., ``PC1'') based distance is significantly better than other PCs across all binding patterns, indicating that comparing to other PCs, the OI-PC could be used to compute binding information.

\subsection{OI Subspace and Relatedness}

\begin{table}[t]
\centering
\scalebox{0.8}{
\begin{tabular}{l}
\hline
\textbf{Input (original)} \\\cmidrule(l){1-1}
\makecell[l]{``The \textit{apple} is in \textit{Box E}, the \textit{bell} is in \textit{Box F}, ...''} \\\cmidrule(l){1-1}
\textbf{Input (Non-related)} \\\cmidrule(l){1-1}
\makecell[l]{``I see \textit{apple}, somewhere else there is \textit{Box E}, \\ the \textit{bell} and \textit{Box F} are scattered around, ...''} \\\cmidrule(l){1-1}
\hline
\end{tabular}
}
\caption{An example of the dataset with non-related expression.}
\label{tab:ctx_norel}
\end{table}
\begin{table}[t]
\centering
\scalebox{0.8}{
\begin{tabular}{l}
\hline
\textbf{Input (7A-7E)} \\\cmidrule(l){1-1}
\makecell[l]{``A$_0$ is in E$_0$, A$_1$ is in E$_1$, A$_2$ is in E$_2$, A$_3$ is in E$_3$,\\ A$_4$ is in E$_4$, A$_5$ is in E$_5$, A$_6$ is in E$_6$.''} \\\cmidrule(l){1-1}
\textbf{Input (7A-3E)} \\\cmidrule(l){1-1}
\makecell[l]{``A$_0$ is in E$_0$, A$_1$ is in E$_0$, A$_2$ is in E$_0$, A$_3$ is in E$_1$,\\ A$_4$ is in E$_1$, A$_5$ is in E$_2$, A$_6$ is in E$_2$.''} \\\cmidrule(l){1-1}
\textbf{Input (7A-2E)} \\\cmidrule(l){1-1}
\makecell[l]{``A$_0$ is in E$_0$, A$_1$ is in E$_0$, A$_2$ is in E$_0$, A$_3$ is in E$_1$,\\ A$_4$ is in E$_1$, A$_5$ is in E$_1$, A$_6$ is in E$_1$.''} \\\cmidrule(l){1-1}
\textbf{Input (7A-5E)} \\\cmidrule(l){1-1}
\makecell[l]{``A$_0$ is in E$_0$, A$_1$ is in E$_0$, A$_2$ is in E$_0$, A$_3$ is in E$_1$,\\ A$_4$ is in E$_2$, A$_5$ is in E$_3$, A$_6$ is in E$_4$.''} \\\cmidrule(l){1-1}
\hline
\end{tabular}
}
\caption{Simplified expression of the datasets with various binding pattern, where ``7A-3E'' represents the pattern containing 7 attributes and 3 entities.}
\label{tab:ctx_mul}
\end{table}

Binding of a EA pair means that the EA pair is bound by a binding relation such as property and location~\cite{treisman1996binding}. In turn, there is no binding when the EA pair is unrelated in its context. This raises a research question that if the correlation between OI-PC of attributes (e.g., OI-PC of ``apple'') and of their corresponding entities (e.g., OI-PC of ``Box E'') represents the relatedness namely the existence of a relation.

\begin{figure}[t]
\centering
\includegraphics[width=4.0cm]{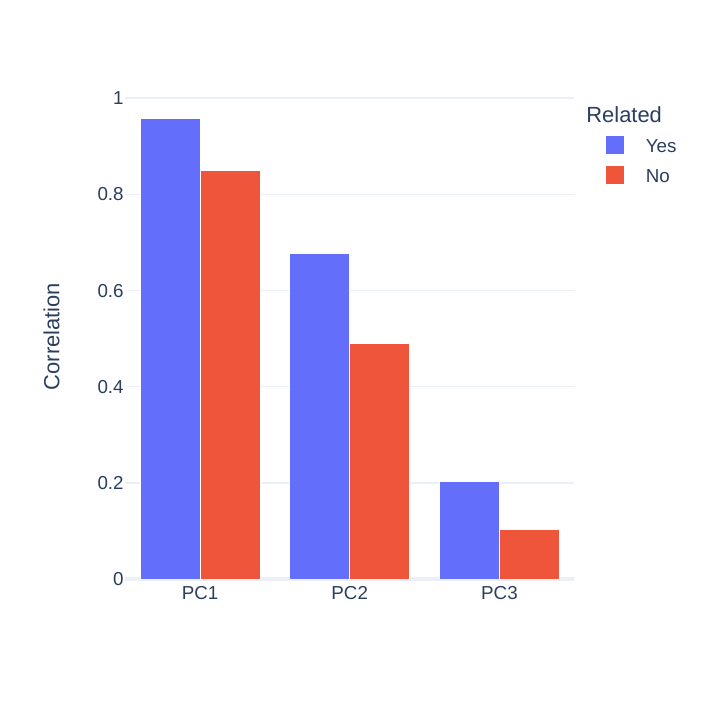}
\caption{OI-PC based correlation between attributes and their corresponding entities, where ``PC$i$'' denotes the $i$-th PC of the OI subspace, ``Yes'' and ``No'' represent the related (i.e., original) and non-related dataset respectively.}
\label{fig:cor_a_e}
\end{figure}

In order to uncover the relationship between OI-PC and the relatedness, we create an alternative dataset by converting relational expression of the entity tracking dataset~\cite{kim2023entity,prakash2024fine} into non-related one. Specifically, we prepare a set of non-related expression templates and randomly select one to replace the original expression of relation as shown in Table~\ref{tab:ctx_norel}. We can observe that the template could make a target EA pair (e.g., ``Box E’’ and ``apple’’) semantically unrelated but retain their OI (e.g., the OI of ``apple’’ and ``bell’’ are still $0$ and $1$ respectively). We select Spearman's rank correlation $\rho$ as the correlation metric and compare the $\rho$ of the non-related dataset with the related one in the Figure~\ref{fig:cor_a_e}. 

We can observe that $\rho$ of non-related dataset is slightly lower than the related (i.e., original) one, indicating that the OI-PC might contain limited relational information so that removing it can marginally decrease the $\rho$. However, there is still strong correlation between the non-related (or non-bound) entity attribute pair, indicating that the OI-PC primarily encodes the OI information but not binding information specifically the information of binding relation.
%The computational circuit of converting OI information into BI is still unclear, and we will leave the issue as our future work.

%

\section{Related Work}

\textbf{Linear Representation} Recent research found that sequence models trained only on next token prediction linearly represent various semantic concepts including Othello board positions~\cite{li2022emergent,nanda2023emergent}, the truth value of assertions~\cite{marks2023geometry}, sentiment~\cite{tigges2023linear}, and numeric values such as elevation, population, birth year, and death year~\cite{gurnee2023language,heinzerling2024monotonic}. Continuing this line of research, in this work, we discover that multiple LMs such as Llama2 can also linearly encode OI along a OI increasing direction in the activations. 

See Appendix~\secref{sec:relw} for additional related work.

\section{Conclusion and Future Work}
In this work, we study the in-context binding, a fundamental skill underlying many complex reasoning and natural language understanding tasks. We provide a novel view of the Binding ID mechanism introduced by \citet{feng2023language} that there exists a low-rank subspace in the hidden state (or activation) of LMs that primarily encodes the ordering information and which is used as the prototype of BIs to causally determine binding.
%To identify the subspace, we present a simple but effective unsupervised method, which is based on PCA. 
%What is more, we also discover that when editing representations along OI-PC in the subspace, LMs tend to bind a given entity to other attributes accordingly. 
Our future work includes: 1. the analysis of OI subspace in a more realistic setting; 2. the study of interaction between in-context binding and factual knowledge learned from pretraining; 3. OI subspace based mechanistic analysis. 

%Although this work takes a step towards a better understanding of the binding problem in LMs, it remains many issues unexplored, and our future work may study around the unanswered questions. Some of these are listed below:
%\begin{itemize}
%\item As we know, binding is not just limited in the ideal setting that contains a single type of relation (e.g., ``is in''). However, this work has not answered how LMs realize binding in a setting with multiple types of predicate, and how an attribute entity pair is linked to its predicate internally.
%\item In reality, complex reasoning requires the interaction between factual associations learned from pretraining and binding information from context. This work has not answered how the knowledge stored in model weights affect the in-context binding.

%\end{itemize}

\section*{Acknowledgement}
This work was supported by JST CREST Grant Number JPMJCR20D2 and JSPS KAKENHI Grant Number 21K17814. We are grateful to the anonymous reviewers for their constructive comments.

\section*{Limitations}
The limitations of our research include the following points: 
1. We only analyze OI subspace on the attribute prediction task, but not on the entity inference task (i.e., given an attribute to infer its entity); 
2. Although PCA based analysis is empirically proven to be effective, we lack the theoretical analysis on why PCA could capture OI subspace;
3. We lack the analysis on how predicate (or relation) affect the OI subspace, and how the results of OI subspace based intervention differ with the type of predicate; 
4. Although we use a publicly available entity tracking dataset, it is still a synthesized dataset. Therefore, for uncovering how LMs bind and track entity in reality, it is necessary to analyze the binding via a real world dataset;
5. We only analyze the activations in the query part instead of in the context part, thus this work can not explain how LMs encode OI in the context and how it is used for OI encoding in query part and how it contributes the binding computation; 
6. We infer the change of BI information via the result of model output but the interaction between OI and BI information is not directly observed. This research thus lacks the mechanistic interpretation on the interaction between them;
7. We only analyze binding from the perspective of representation and localize OI subspace. However, we have not answered what is the mechanism that generates the subspace and what is the circuit that utilizes the subspace for binding. 

\section*{Ethical Statement}
The existing dataset~\cite{kim2023entity,prakash2024fine} and LMs (i.e., Llama2-7B, Float-7B, Llama3-8B, Qwen1.5-7B and Pythia-6.9B) are applied according to their intended research purpose. The synthetic datasets we adopted in this work are automatically created by strictly following the rule (or pattern) of the existing dataset, where the entities and attributes are sampled from a pool of wide variety of one-token names and concepts. Therefore, there is no ethical concern on human annotation bias and semantic biases. 
%The datasets and code will be publicly available to ensure the reproducibility of our experiments.

\bibliography{acl_latex}

\begin{thebibliography}{41}
\providecommand{\natexlab}[1]{#1}

\bibitem[{AI@Meta(2024)}]{llama3modelcard}
AI@Meta. 2024.
\newblock \href {https://github.com/meta-llama/llama3/blob/main/MODEL_CARD.md} {Llama 3 model card}.

\bibitem[{Bai et~al.(2023)Bai, Bai, Chu, Cui, Dang, Deng, Fan, Ge, Han, Huang et~al.}]{bai2023qwen}
Jinze Bai, Shuai Bai, Yunfei Chu, Zeyu Cui, Kai Dang, Xiaodong Deng, Yang Fan, Wenbin Ge, Yu~Han, Fei Huang, et~al. 2023.
\newblock Qwen technical report.
\newblock \emph{arXiv preprint arXiv:2309.16609}.

\bibitem[{Barzilay and Lapata(2008)}]{barzilay2008modeling}
Regina Barzilay and Mirella Lapata. 2008.
\newblock Modeling local coherence: An entity-based approach.
\newblock \emph{Computational Linguistics}, 34(1):1--34.

\bibitem[{Biderman et~al.(2023)Biderman, Schoelkopf, Anthony, Bradley, O’Brien, Hallahan, Khan, Purohit, Prashanth, Raff et~al.}]{biderman2023pythia}
Stella Biderman, Hailey Schoelkopf, Quentin~Gregory Anthony, Herbie Bradley, Kyle O’Brien, Eric Hallahan, Mohammad~Aflah Khan, Shivanshu Purohit, USVSN~Sai Prashanth, Edward Raff, et~al. 2023.
\newblock Pythia: A suite for analyzing large language models across training and scaling.
\newblock In \emph{International Conference on Machine Learning}, pages 2397--2430. PMLR.

\bibitem[{Dai et~al.(2021)Dai, Dong, Hao, Sui, Chang, and Wei}]{dai2021knowledge}
Damai Dai, Li~Dong, Yaru Hao, Zhifang Sui, Baobao Chang, and Furu Wei. 2021.
\newblock Knowledge neurons in pretrained transformers.
\newblock \emph{arXiv preprint arXiv:2104.08696}.

\bibitem[{Elhage et~al.(2021)Elhage, Nanda, Olsson, Henighan, Joseph, Mann, Askell, Bai, Chen, Conerly et~al.}]{elhage2021mathematical}
Nelson Elhage, Neel Nanda, Catherine Olsson, Tom Henighan, Nicholas Joseph, Ben Mann, Amanda Askell, Yuntao Bai, Anna Chen, Tom Conerly, et~al. 2021.
\newblock A mathematical framework for transformer circuits.
\newblock \emph{Transformer Circuits Thread}, 1:1.

\bibitem[{Engels et~al.(2024)Engels, Liao, Michaud, Gurnee, and Tegmark}]{2405.14860}
Joshua Engels, Isaac Liao, Eric~J. Michaud, Wes Gurnee, and Max Tegmark. 2024.
\newblock \href {https://arxiv.org/abs/arXiv:2405.14860} {Not all language model features are linear}.

\bibitem[{Feng and Steinhardt(2023)}]{feng2023language}
Jiahai Feng and Jacob Steinhardt. 2023.
\newblock How do language models bind entities in context?
\newblock \emph{arXiv preprint arXiv:2310.17191}.

\bibitem[{Geiger et~al.(2021)Geiger, Lu, Icard, and Potts}]{geiger2021causal}
Atticus Geiger, Hanson Lu, Thomas Icard, and Christopher Potts. 2021.
\newblock Causal abstractions of neural networks.
\newblock \emph{Advances in Neural Information Processing Systems}, 34:9574--9586.

\bibitem[{Geiger et~al.(2020)Geiger, Richardson, and Potts}]{geiger2020neural}
Atticus Geiger, Kyle Richardson, and Christopher Potts. 2020.
\newblock Neural natural language inference models partially embed theories of lexical entailment and negation.
\newblock \emph{arXiv preprint arXiv:2004.14623}.

\bibitem[{Geiger et~al.(2022)Geiger, Wu, Lu, Rozner, Kreiss, Icard, Goodman, and Potts}]{geiger2022inducing}
Atticus Geiger, Zhengxuan Wu, Hanson Lu, Josh Rozner, Elisa Kreiss, Thomas Icard, Noah Goodman, and Christopher Potts. 2022.
\newblock Inducing causal structure for interpretable neural networks.
\newblock In \emph{International Conference on Machine Learning}, pages 7324--7338. PMLR.

\bibitem[{Geva et~al.(2023)Geva, Bastings, Filippova, and Globerson}]{geva2023dissecting}
Mor Geva, Jasmijn Bastings, Katja Filippova, and Amir Globerson. 2023.
\newblock Dissecting recall of factual associations in auto-regressive language models.
\newblock \emph{arXiv preprint arXiv:2304.14767}.

\bibitem[{Geva et~al.(2020)Geva, Schuster, Berant, and Levy}]{geva2020transformer}
Mor Geva, Roei Schuster, Jonathan Berant, and Omer Levy. 2020.
\newblock Transformer feed-forward layers are key-value memories.
\newblock \emph{arXiv preprint arXiv:2012.14913}.

\bibitem[{Gurnee and Tegmark(2023)}]{gurnee2023language}
Wes Gurnee and Max Tegmark. 2023.
\newblock Language models represent space and time.
\newblock \emph{arXiv preprint arXiv:2310.02207}.

\bibitem[{Hanna et~al.(2024)Hanna, Liu, and Variengien}]{hanna2024does}
Michael Hanna, Ollie Liu, and Alexandre Variengien. 2024.
\newblock How does gpt-2 compute greater-than?: Interpreting mathematical abilities in a pre-trained language model.
\newblock \emph{Advances in Neural Information Processing Systems}, 36.

\bibitem[{Heim(1983)}]{heim1983file}
Irene Heim. 1983.
\newblock File change semantics and the familiarity theory of definiteness.
\newblock \emph{Semantics Critical Concepts in Linguistics}, pages 108--135.

\bibitem[{Heinzerling and Inui(2024)}]{heinzerling2024monotonic}
Benjamin Heinzerling and Kentaro Inui. 2024.
\newblock Monotonic representation of numeric properties in language models.
\newblock \emph{arXiv preprint arXiv:2403.10381}.

\bibitem[{Hernandez et~al.(2023)Hernandez, Sharma, Haklay, Meng, Wattenberg, Andreas, Belinkov, and Bau}]{hernandez2023linearity}
Evan Hernandez, Arnab~Sen Sharma, Tal Haklay, Kevin Meng, Martin Wattenberg, Jacob Andreas, Yonatan Belinkov, and David Bau. 2023.
\newblock Linearity of relation decoding in transformer language models.
\newblock \emph{arXiv preprint arXiv:2308.09124}.

\bibitem[{Kamp et~al.(2010)Kamp, Van~Genabith, and Reyle}]{kamp2010discourse}
Hans Kamp, Josef Van~Genabith, and Uwe Reyle. 2010.
\newblock Discourse representation theory.
\newblock In \emph{Handbook of Philosophical Logic: Volume 15}, pages 125--394. Springer.

\bibitem[{Karttunen(1976)}]{karttunen1976discourse}
Lauri Karttunen. 1976.
\newblock Discourse referents.
\newblock In \emph{Notes from the linguistic underground}, pages 363--385. Brill.

\bibitem[{Kim and Schuster(2023)}]{kim2023entity}
Najoung Kim and Sebastian Schuster. 2023.
\newblock Entity tracking in language models.
\newblock \emph{arXiv preprint arXiv:2305.02363}.

\bibitem[{Kim et~al.(2024)Kim, Schuster, and Toshniwal}]{kim2024code}
Najoung Kim, Sebastian Schuster, and Shubham Toshniwal. 2024.
\newblock Code pretraining improves entity tracking abilities of language models.
\newblock \emph{arXiv preprint arXiv:2405.21068}.

\bibitem[{Kram{\'a}r et~al.(2024)Kram{\'a}r, Lieberum, Shah, and Nanda}]{kramar2024atp}
J{\'a}nos Kram{\'a}r, Tom Lieberum, Rohin Shah, and Neel Nanda. 2024.
\newblock Atp*: An efficient and scalable method for localizing llm behaviour to components.
\newblock \emph{arXiv preprint arXiv:2403.00745}.

\bibitem[{Lad et~al.(2024)Lad, Gurnee, and Tegmark}]{lad2024remarkable}
Vedang Lad, Wes Gurnee, and Max Tegmark. 2024.
\newblock The remarkable robustness of llms: Stages of inference?
\newblock \emph{arXiv preprint arXiv:2406.19384}.

\bibitem[{Li et~al.(2022)Li, Hopkins, Bau, Vi{\'e}gas, Pfister, and Wattenberg}]{li2022emergent}
Kenneth Li, Aspen~K Hopkins, David Bau, Fernanda Vi{\'e}gas, Hanspeter Pfister, and Martin Wattenberg. 2022.
\newblock Emergent world representations: Exploring a sequence model trained on a synthetic task.
\newblock \emph{arXiv preprint arXiv:2210.13382}.

\bibitem[{Marks and Tegmark(2023)}]{marks2023geometry}
Samuel Marks and Max Tegmark. 2023.
\newblock The geometry of truth: Emergent linear structure in large language model representations of true/false datasets.
\newblock \emph{arXiv preprint arXiv:2310.06824}.

\bibitem[{Matsumoto et~al.(2023)Matsumoto, Heinzerling, Yoshikawa, and Inui}]{matsumoto2023tracing}
Yuta Matsumoto, Benjamin Heinzerling, Masashi Yoshikawa, and Kentaro Inui. 2023.
\newblock Tracing and manipulating intermediate values in neural math problem solvers.
\newblock \emph{arXiv preprint arXiv:2301.06758}.

\bibitem[{Meng et~al.(2022)Meng, Bau, Andonian, and Belinkov}]{meng2022locating}
Kevin Meng, David Bau, Alex Andonian, and Yonatan Belinkov. 2022.
\newblock Locating and editing factual associations in gpt.
\newblock \emph{Advances in Neural Information Processing Systems}, 35:17359--17372.

\bibitem[{Nanda et~al.(2023)Nanda, Lee, and Wattenberg}]{nanda2023emergent}
Neel Nanda, Andrew Lee, and Martin Wattenberg. 2023.
\newblock Emergent linear representations in world models of self-supervised sequence models.
\newblock \emph{arXiv preprint arXiv:2309.00941}.

\bibitem[{Nieuwland and Van~Berkum(2006)}]{nieuwland2006peanuts}
Mante~S Nieuwland and Jos~JA Van~Berkum. 2006.
\newblock When peanuts fall in love: N400 evidence for the power of discourse.
\newblock \emph{Journal of cognitive neuroscience}, 18(7):1098--1111.

\bibitem[{Prakash et~al.(2024)Prakash, Shaham, Haklay, Belinkov, and Bau}]{prakash2024fine}
Nikhil Prakash, Tamar~Rott Shaham, Tal Haklay, Yonatan Belinkov, and David Bau. 2024.
\newblock Fine-tuning enhances existing mechanisms: A case study on entity tracking.
\newblock \emph{arXiv preprint arXiv:2402.14811}.

\bibitem[{Stolfo et~al.(2023)Stolfo, Belinkov, and Sachan}]{stolfo2023mechanistic}
Alessandro Stolfo, Yonatan Belinkov, and Mrinmaya Sachan. 2023.
\newblock A mechanistic interpretation of arithmetic reasoning in language models using causal mediation analysis.
\newblock \emph{arXiv preprint arXiv:2305.15054}.

\bibitem[{Tigges et~al.(2023)Tigges, Hollinsworth, Geiger, and Nanda}]{tigges2023linear}
Curt Tigges, Oskar~John Hollinsworth, Atticus Geiger, and Neel Nanda. 2023.
\newblock Linear representations of sentiment in large language models.
\newblock \emph{arXiv preprint arXiv:2310.15154}.

\bibitem[{Touvron et~al.(2023)Touvron, Martin, Stone, Albert, Almahairi, Babaei, Bashlykov, Batra, Bhargava, Bhosale, Bikel, Blecher, Ferrer, Chen, Cucurull, Esiobu, Fernandes, Fu, Fu, Fuller, Gao, Goswami, Goyal, Hartshorn, Hosseini, Hou, Inan, Kardas, Kerkez, Khabsa, Kloumann, Korenev, Koura, Lachaux, Lavril, Lee, Liskovich, Lu, Mao, Martinet, Mihaylov, Mishra, Molybog, Nie, Poulton, Reizenstein, Rungta, Saladi, Schelten, Silva, Smith, Subramanian, Tan, Tang, Taylor, Williams, Kuan, Xu, Yan, Zarov, Zhang, Fan, Kambadur, Narang, Rodriguez, Stojnic, Edunov, and Scialom}]{2307.09288}
Hugo Touvron, Louis Martin, Kevin Stone, Peter Albert, Amjad Almahairi, Yasmine Babaei, Nikolay Bashlykov, Soumya Batra, Prajjwal Bhargava, Shruti Bhosale, Dan Bikel, Lukas Blecher, Cristian~Canton Ferrer, Moya Chen, Guillem Cucurull, David Esiobu, Jude Fernandes, Jeremy Fu, Wenyin Fu, Brian Fuller, Cynthia Gao, Vedanuj Goswami, Naman Goyal, Anthony Hartshorn, Saghar Hosseini, Rui Hou, Hakan Inan, Marcin Kardas, Viktor Kerkez, Madian Khabsa, Isabel Kloumann, Artem Korenev, Punit~Singh Koura, Marie-Anne Lachaux, Thibaut Lavril, Jenya Lee, Diana Liskovich, Yinghai Lu, Yuning Mao, Xavier Martinet, Todor Mihaylov, Pushkar Mishra, Igor Molybog, Yixin Nie, Andrew Poulton, Jeremy Reizenstein, Rashi Rungta, Kalyan Saladi, Alan Schelten, Ruan Silva, Eric~Michael Smith, Ranjan Subramanian, Xiaoqing~Ellen Tan, Binh Tang, Ross Taylor, Adina Williams, Jian~Xiang Kuan, Puxin Xu, Zheng Yan, Iliyan Zarov, Yuchen Zhang, Angela Fan, Melanie Kambadur, Sharan Narang, Aurelien Rodriguez, Robert Stojnic, Sergey Edunov, and Thomas
  Scialom. 2023.
\newblock \href {https://arxiv.org/abs/arXiv:2307.09288} {Llama 2: Open foundation and fine-tuned chat models}.

\bibitem[{Treisman(1996)}]{treisman1996binding}
Anne Treisman. 1996.
\newblock The binding problem.
\newblock \emph{Current opinion in neurobiology}, 6(2):171--178.

\bibitem[{Turner et~al.(2023)Turner, Thiergart, Udell, Leech, Mini, and MacDiarmid}]{turner2023activation}
Alex Turner, Lisa Thiergart, David Udell, Gavin Leech, Ulisse Mini, and Monte MacDiarmid. 2023.
\newblock Activation addition: Steering language models without optimization.
\newblock \emph{arXiv preprint arXiv:2308.10248}.

\bibitem[{Vig et~al.(2020)Vig, Gehrmann, Belinkov, Qian, Nevo, Singer, and Shieber}]{vig2020investigating}
Jesse Vig, Sebastian Gehrmann, Yonatan Belinkov, Sharon Qian, Daniel Nevo, Yaron Singer, and Stuart Shieber. 2020.
\newblock Investigating gender bias in language models using causal mediation analysis.
\newblock \emph{Advances in neural information processing systems}, 33:12388--12401.

\bibitem[{Wang et~al.(2022)Wang, Variengien, Conmy, Shlegeris, and Steinhardt}]{wang2022interpretability}
Kevin Wang, Alexandre Variengien, Arthur Conmy, Buck Shlegeris, and Jacob Steinhardt. 2022.
\newblock Interpretability in the wild: a circuit for indirect object identification in gpt-2 small.
\newblock \emph{arXiv preprint arXiv:2211.00593}.

\bibitem[{Wold et~al.(2001)Wold, Sj{\"o}str{\"o}m, and Eriksson}]{wold2001pls}
Svante Wold, Michael Sj{\"o}str{\"o}m, and Lennart Eriksson. 2001.
\newblock Pls-regression: a basic tool of chemometrics chemometr.
\newblock \emph{Intell. Lab}, 58(2):109--130.

\bibitem[{Wu et~al.(2024)Wu, Geiger, Icard, Potts, and Goodman}]{wu2024interpretability}
Zhengxuan Wu, Atticus Geiger, Thomas Icard, Christopher Potts, and Noah Goodman. 2024.
\newblock Interpretability at scale: Identifying causal mechanisms in alpaca.
\newblock \emph{Advances in Neural Information Processing Systems}, 36.

\bibitem[{Yang et~al.(2021)Yang, Yang, Cer, and Darve}]{yang2021simple}
Ziyi Yang, Yinfei Yang, Daniel Cer, and Eric Darve. 2021.
\newblock A simple and effective method to eliminate the self language bias in multilingual representations.
\newblock \emph{arXiv preprint arXiv:2109.04727}.

\end{thebibliography}

\clearpage
\onecolumn
\appendix
\section{Appendix}

\subsection{Other Related Work}
\label{sec:relw}
\textbf{Knowledge Localization} Many works aim to localize and edit factual relations (e.g., ``capital of'') that LMs learn from pretraining and are stored into model weights~\cite{geva2020transformer,dai2021knowledge,meng2022locating,geva2023dissecting,hernandez2023linearity}. Different from this line of research, this work studies in-context representations of relations and analyzes how they are represented in model activations.

\textbf{Mechanistic Interpretability} Notable progress has been made in uncovering circuits performing various tasks within LMs~\cite{elhage2021mathematical,wang2022interpretability,wu2024interpretability}. Recently, \citet{prakash2024fine} identify the circuit for entity tracking task. \citet{feng2023language} introduce a Binding ID Mechanism for explaining the binding problem, state that LMs use the abstract concept BI to internally mark entity-attribute pairs. However, they have not captured the OI information from entity activations that directly determines the binding behaviour.
%Therefore, they have not answered how LMs store the BI information into entity activations, how to localize the BI information and whether the localized BI information causally affect the model binding behaviour. 
%In this work, we provide a novel view of the Binding ID mechanism and discover that there exists a BI encoding subspace in entity activations, and the direct intervention on the subspace will alter model output.

\subsection{Partial least squares regression and PCA}
\label{sec:pls}

\begin{figure*}[h]
\centering
\includegraphics[width=10.0cm]{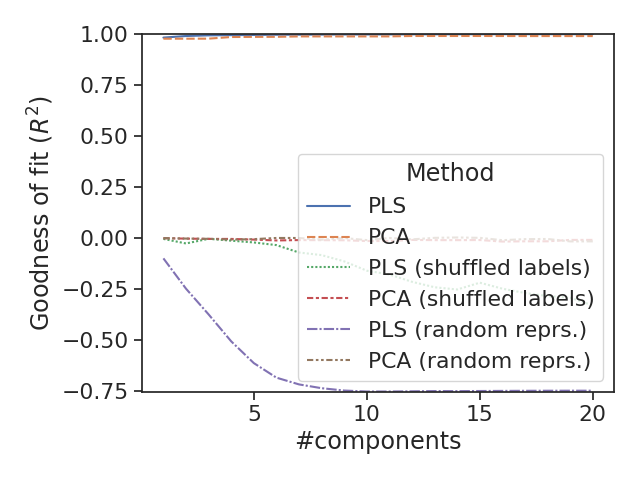}
\caption{Regression curves for PLS and PCA.}
\label{fig:pls}
\end{figure*}

Besides PCA, a commonly used unsupervised Dimension Reduction (DR) method, we also attempt Partial Least Squares regression (PLS)~\cite{wold2001pls}, a supervised DR method. PLS extracts a set of ordered latent variables that maximizes the co-variability between the features (e.g., activations) and the scores to be predicted (e.g., OI). We perform PCA and PLS on a development set and compare their regression cures in Figure~\ref{fig:pls}. We can observe that both the first PCA component and the first PLS direction contain almost all information about OI of target entity, because their regression score is close to one. The consistency indicates that PCA is an effective method to capture OI subspace.

\clearpage

\subsection{ICA for OI Subspace}
\label{sec:ica}
\begin{figure*}[h]
\centering
\includegraphics[width=10.0cm]{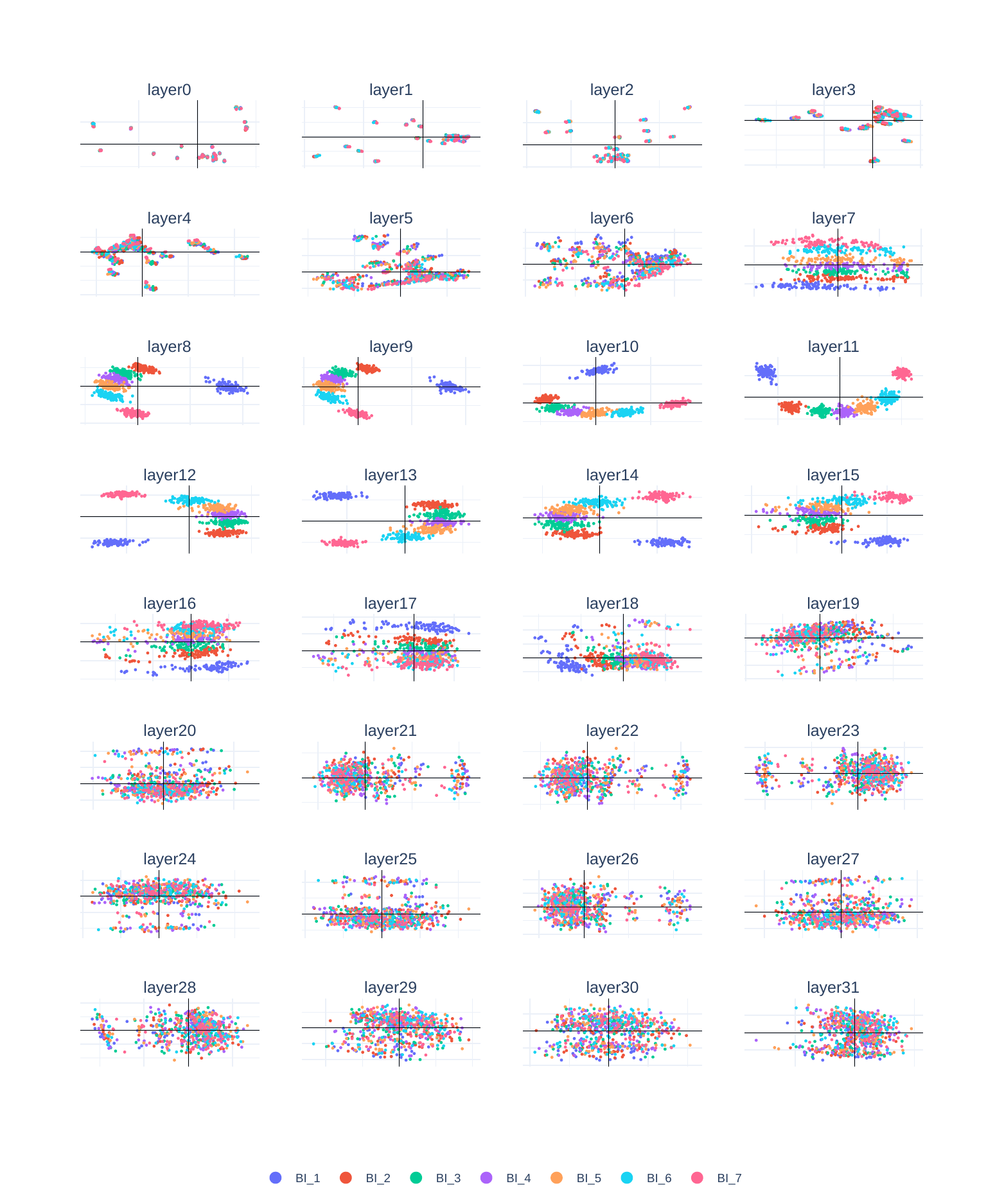}
\caption{Subspace visualization from ICA on Llama2-7B, where ``BI'' primarily denotes OI.}
\label{fig:ica_space}
\end{figure*}

\begin{figure*}[!h]
\centering
\begin{subfigure}{.4\textwidth}
  \centering
  \includegraphics[width=\columnwidth]{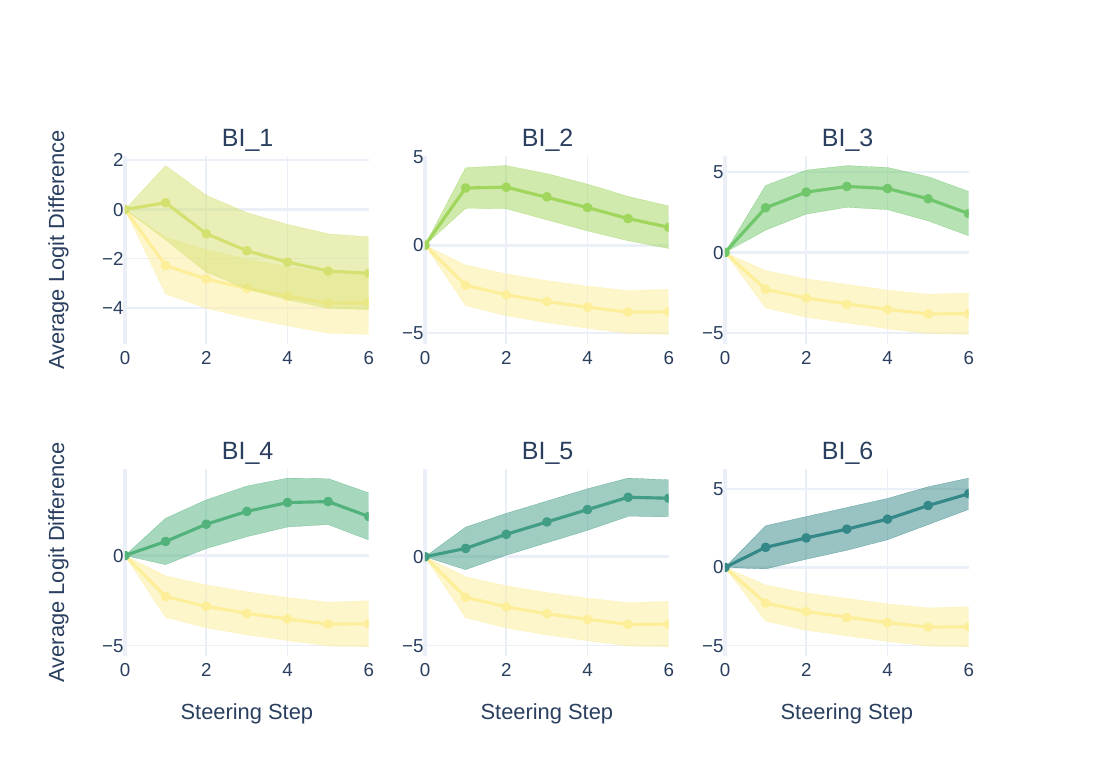}
\end{subfigure}%
\begin{subfigure}{.4\textwidth}
  \centering
  \includegraphics[width=\columnwidth]{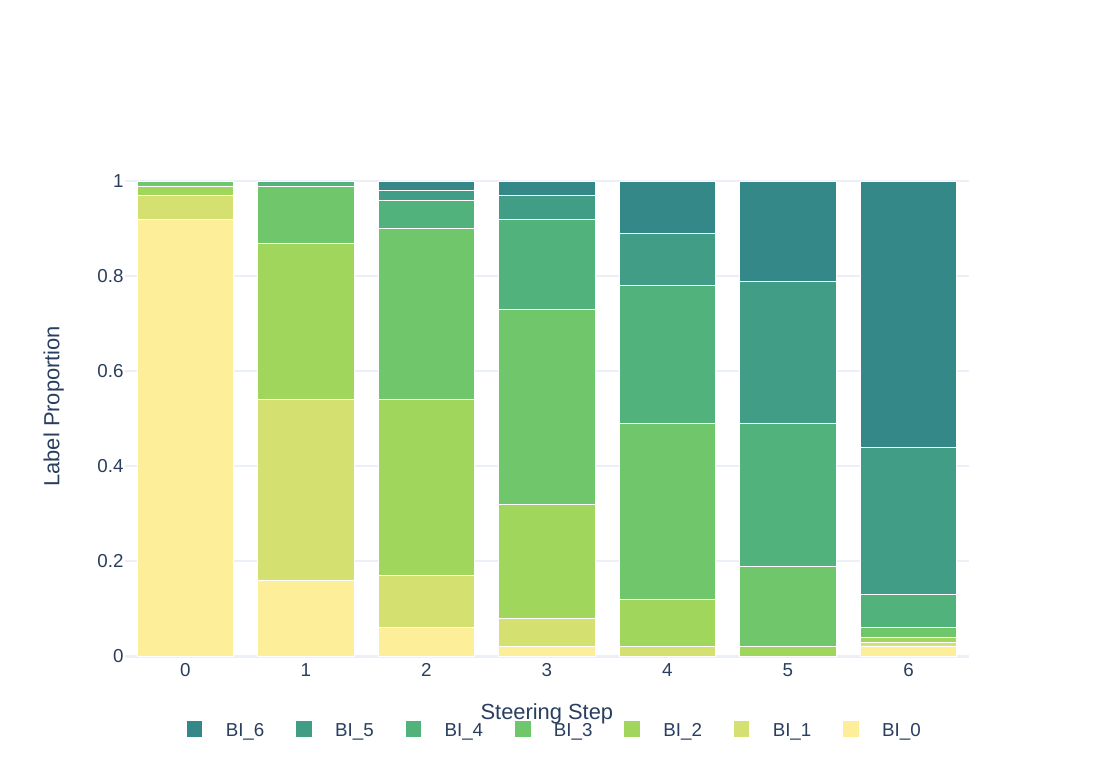}
\end{subfigure}
\caption{Logit Difference and Logit Flip for ICA based activation steering on the entity tracking dataset (i.e., r: is\_in) on Llama2-7B.}
\label{fig:ica_results}
\end{figure*}

Independent Component Analysis (ICA) is a statistical method to reveal hidden subcomponent from multivariate signal. Similar to PCA, ICA is utilized as a dimension reduction method. We apply ICA to analyze the activations of entities and visualize the subspace in Figure~\ref{fig:ica_space}. We can observe that the distribution is generally similar to the one of PCA. In addition, we also conduct causal intervention (i.e., activation steering) based on the subspace and present the results in Figure~\ref{fig:ica_results}. We can observe that the results are also similar to those from PCA based intervention. This indicates that besides PCA, ICA can also be used as a computational method to capture OI subspace.


\clearpage

\subsection{Layer-wise Intervention}
\label{sec:layer_ld}
\begin{figure*}[h]
\centering
\begin{subfigure}{.5\textwidth}
  \centering
  \includegraphics[width=\columnwidth]{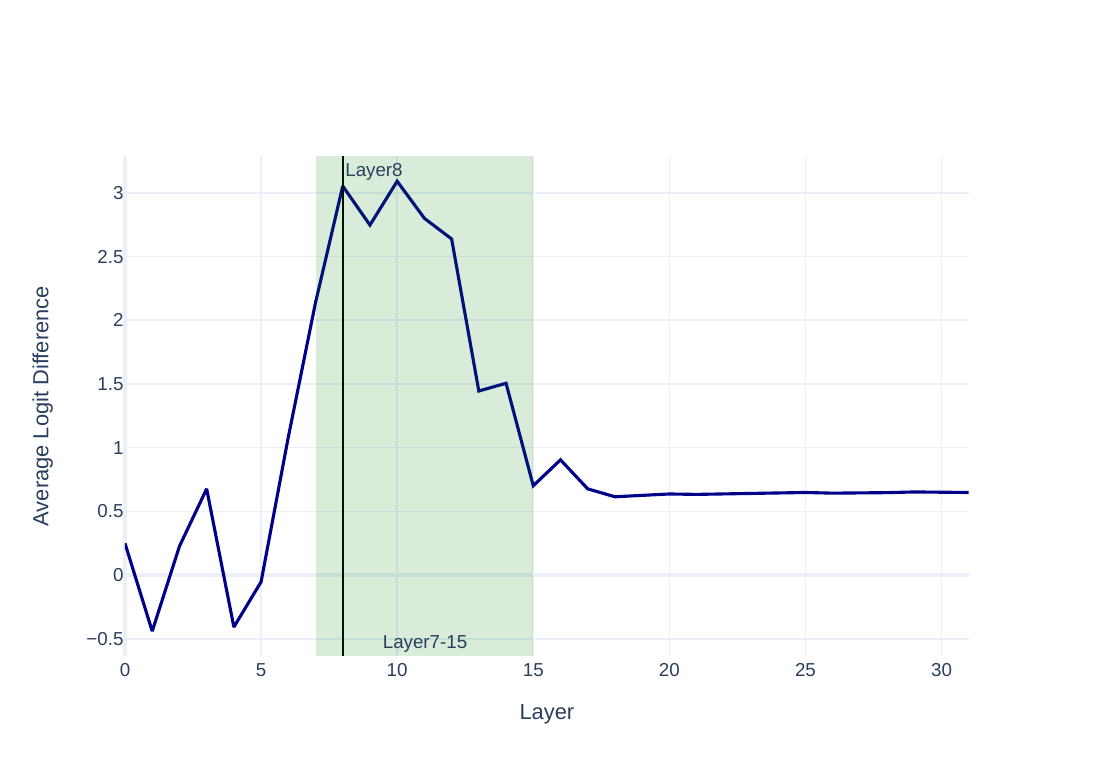}
  \caption{}
  \label{fig:lwald}
\end{subfigure}%
\begin{subfigure}{.5\textwidth}
  \centering
  \includegraphics[width=\columnwidth]{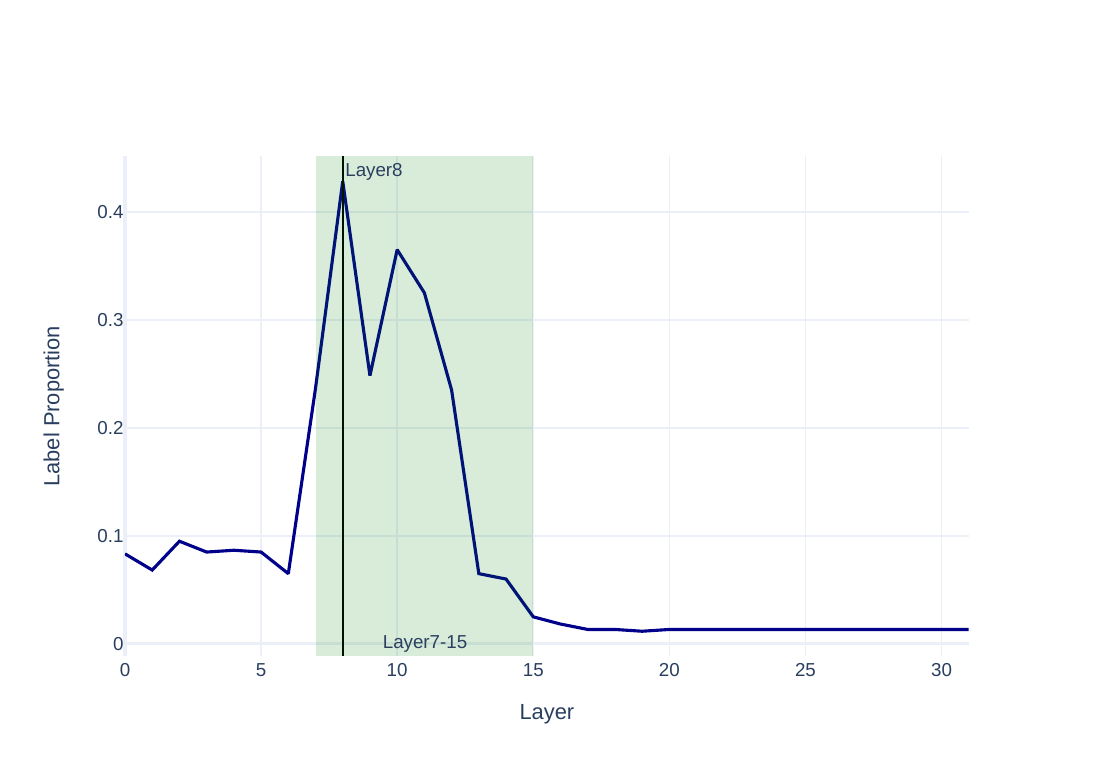}
  \caption{}
  \label{fig:lwafp}
\end{subfigure}
\caption{Average Logit Difference (LD) and logit flip for layer-wise OI-PC based intervention on Llama2-7B, where x axis denotes the layer, the colored zone indicates the layers that are sensitive to the intervention, and the vertical line represents the most effective layer (i.e., Layer 8), Y axis in Figure~\ref{fig:lwald} and Figure~\ref{fig:lwafp} denotes the average LD and the proportion of inferred attributes (excluding the original one) respectively.}
\label{fig:layer_wise_inter_llama2}
\end{figure*}

To localize the layer that contributes to binding behaviour, we perform layer-wise OI-PC based intervention mentioned in Section~\secref{sec:ap_subspace} on our development set. In Figure~\ref{fig:layer_wise_inter_llama2}, we can observe that OI subspace from middle layers (i.e., from layer7 to layer15, especially layer8) significantly affect the computation of binding, and interestingly, these layers also overlap with the ones that clearly encode OI information, as shown in Figure~\ref{fig:layer_wise_pca}. Based on the analysis, we select the layer to perform activation patching.

\clearpage
\subsection{Layer-wise Embedding Visualization on Llama3-8B and Float-7B}
\label{sec:2emb_visual}
\begin{figure*}[!h]
\centering
\begin{subfigure}{.5\textwidth}
  \centering
  \includegraphics[width=\columnwidth]{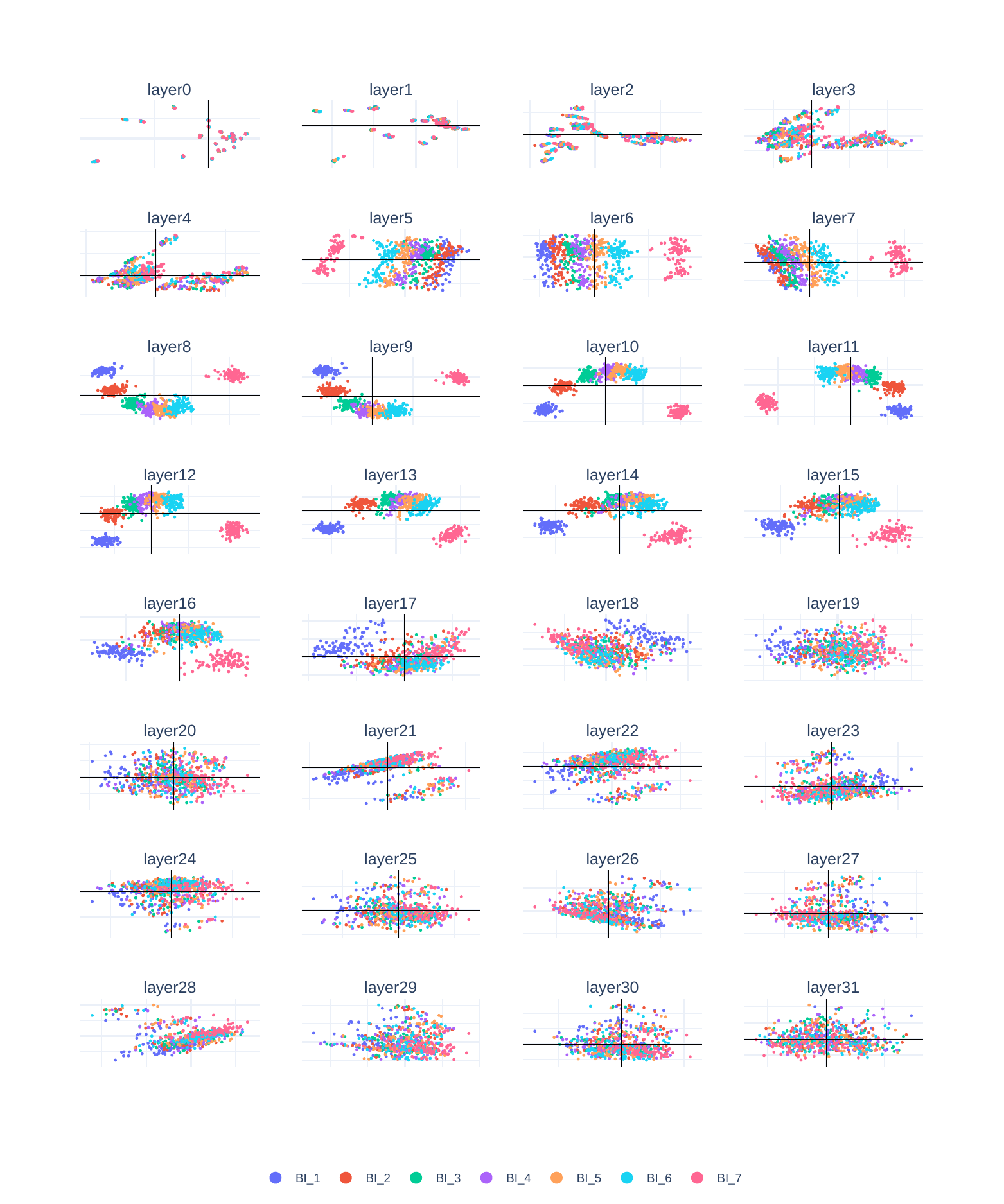}
  \caption{Llama3-8B}
\end{subfigure}%
\begin{subfigure}{.5\textwidth}
  \centering
  \includegraphics[width=\columnwidth]{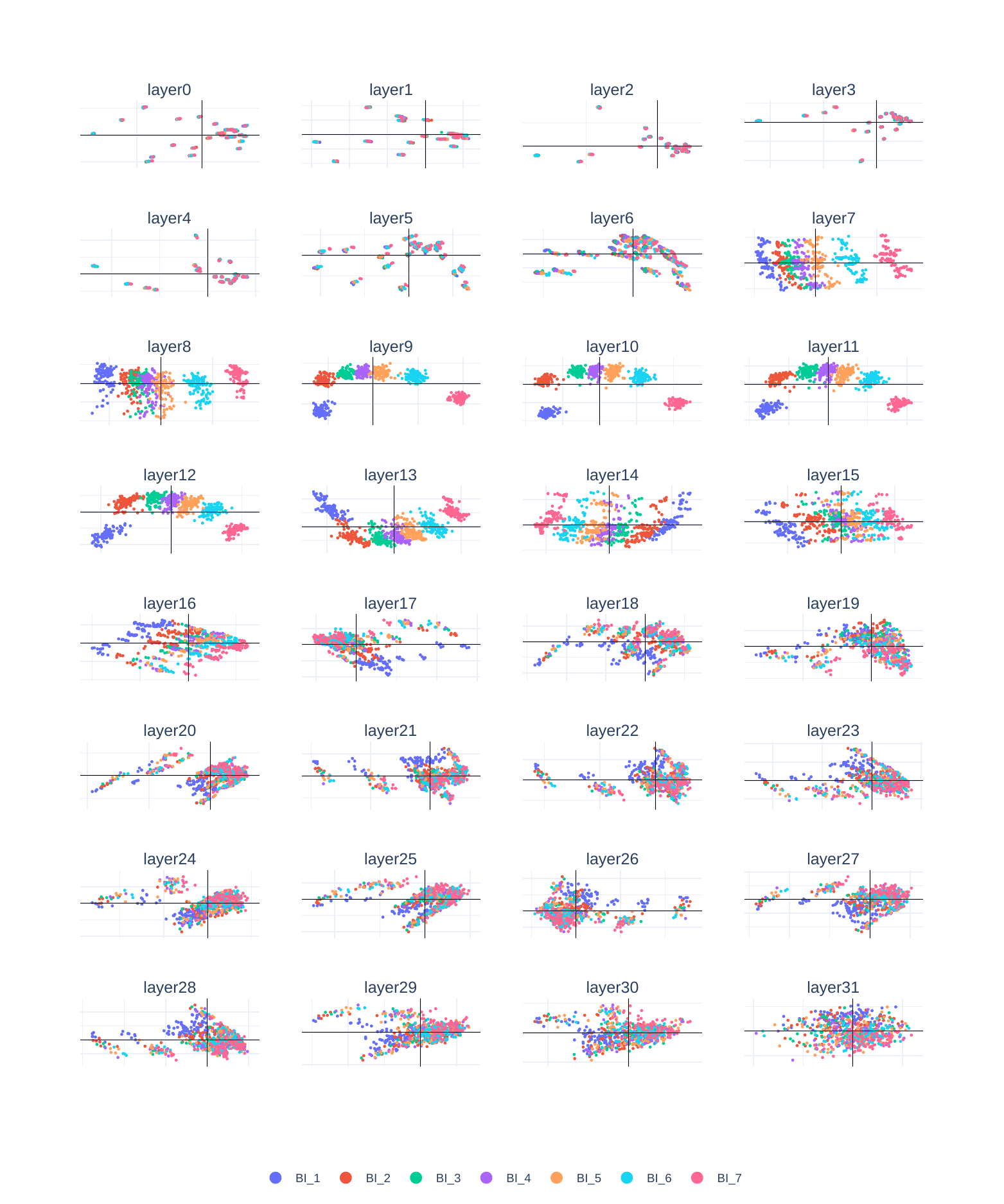}
  \caption{Float-7B}
\end{subfigure}
\caption{Layer-wise OI subspace visualization on Llama3-8B and Float-7B.}
\label{fig:layer_wise_pca_llama3_float}
\end{figure*}

\clearpage
\subsection{OI Subspace on other LM Families}
\label{sec:other_llm}
\begin{figure*}[!h]
\centering
\begin{subfigure}{.5\textwidth}
  \centering
  \includegraphics[width=\columnwidth]{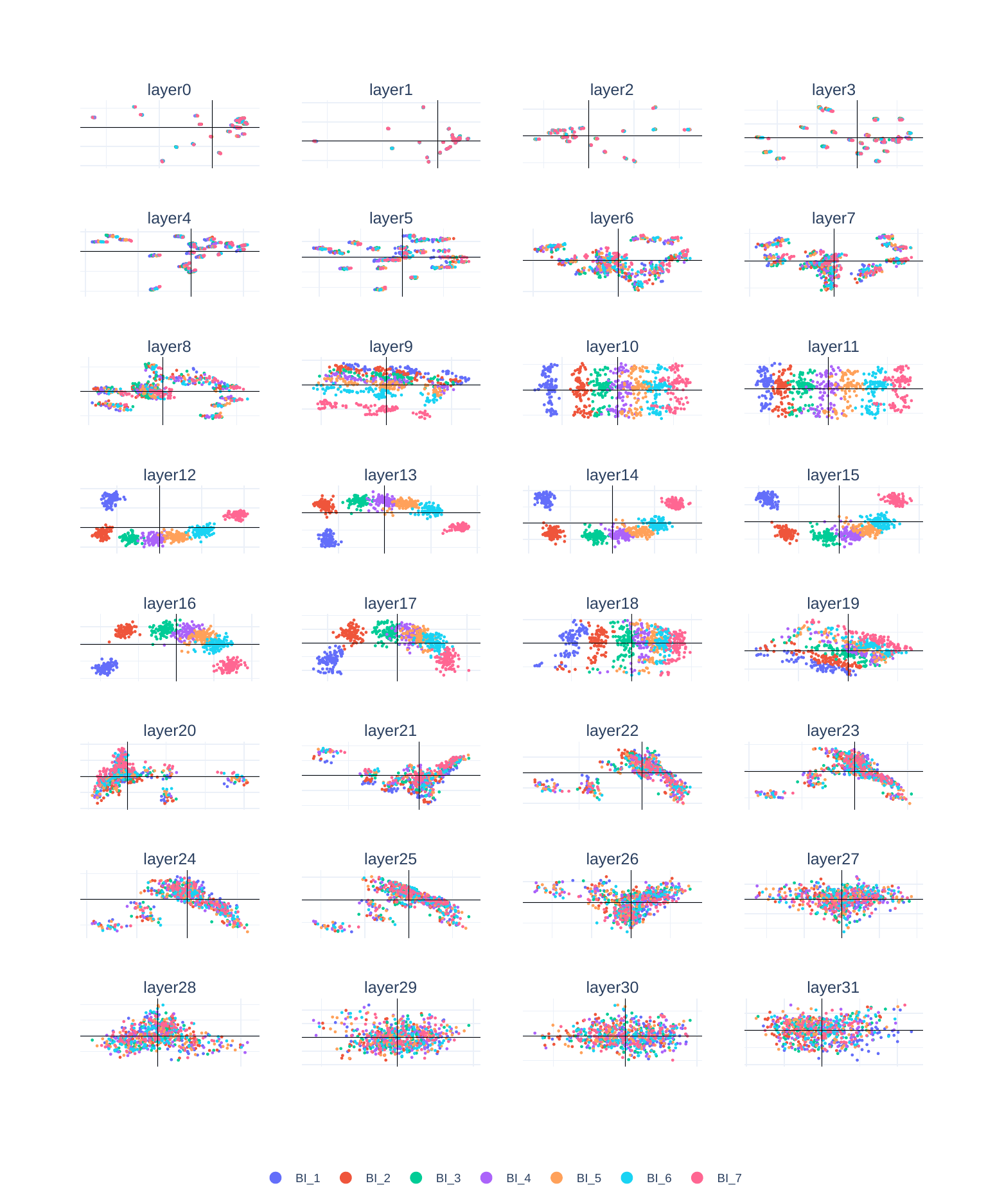}
  \caption{Qwen1.5-7B}
\end{subfigure}%
\begin{subfigure}{.5\textwidth}
  \centering
  \includegraphics[width=\columnwidth]{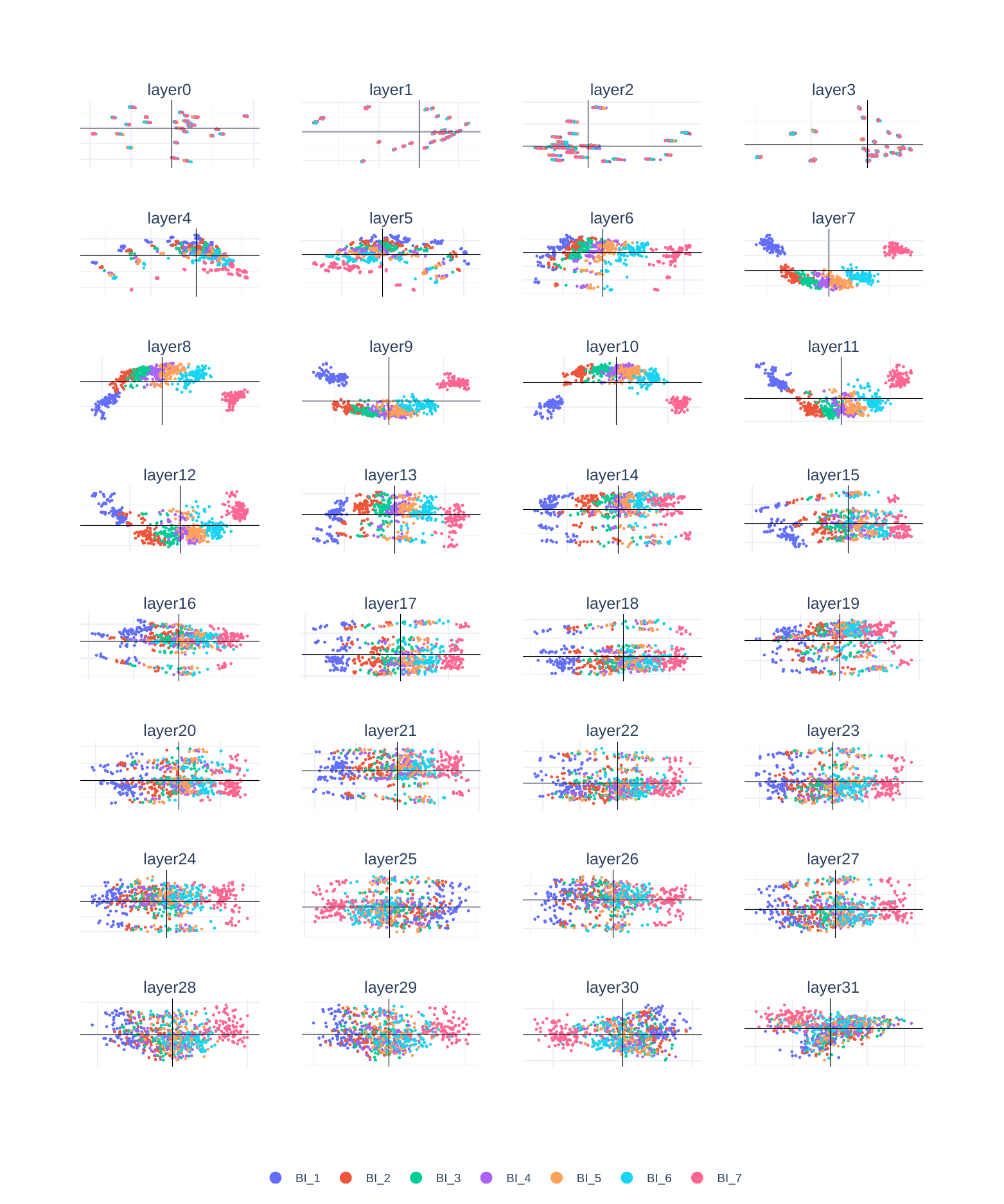}
  \caption{Pythia-6.9B}
\end{subfigure}
\caption{Layer-wise OI subspace visualization on Qwen1.5-7B and Pythia-6.9B.}
\label{fig:layer_wise_pca_qwen_pythia}
\end{figure*}

\begin{figure*}[!h]
\centering
\begin{subfigure}{.4\textwidth}
  \centering
  \includegraphics[width=\columnwidth]{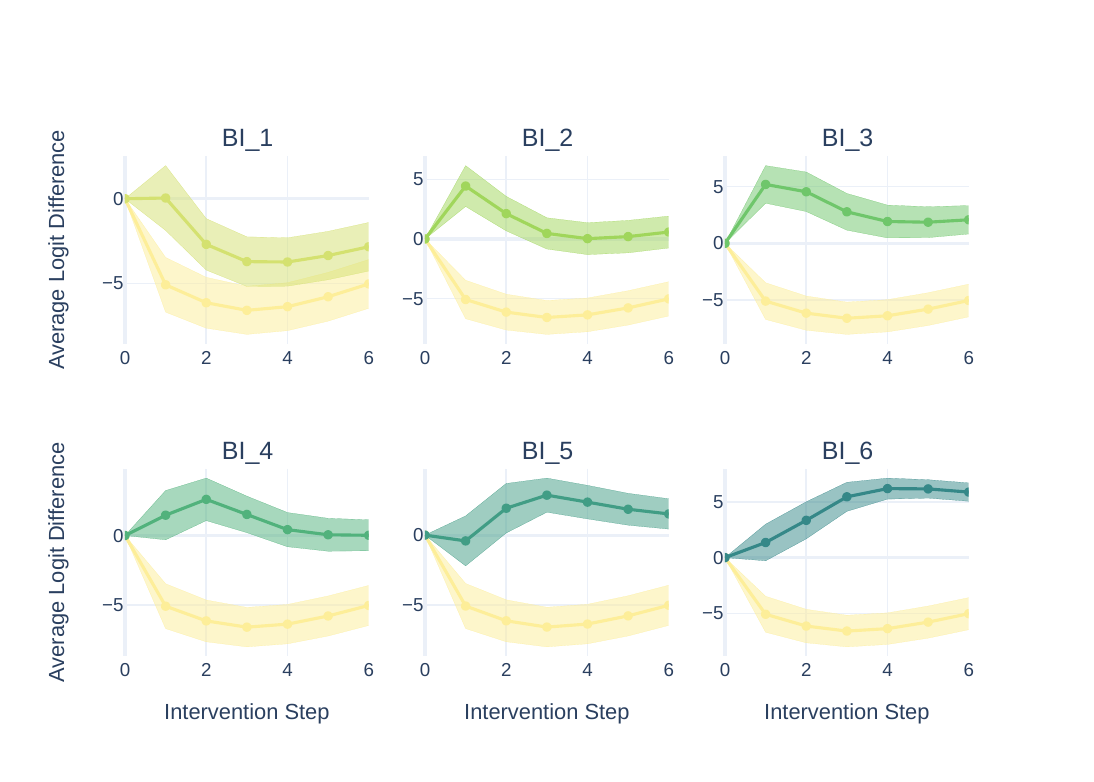}
  \caption{LD for Qwen1.5-7B.}
\end{subfigure}%
\begin{subfigure}{.4\textwidth}
  \centering
  \includegraphics[width=\columnwidth]{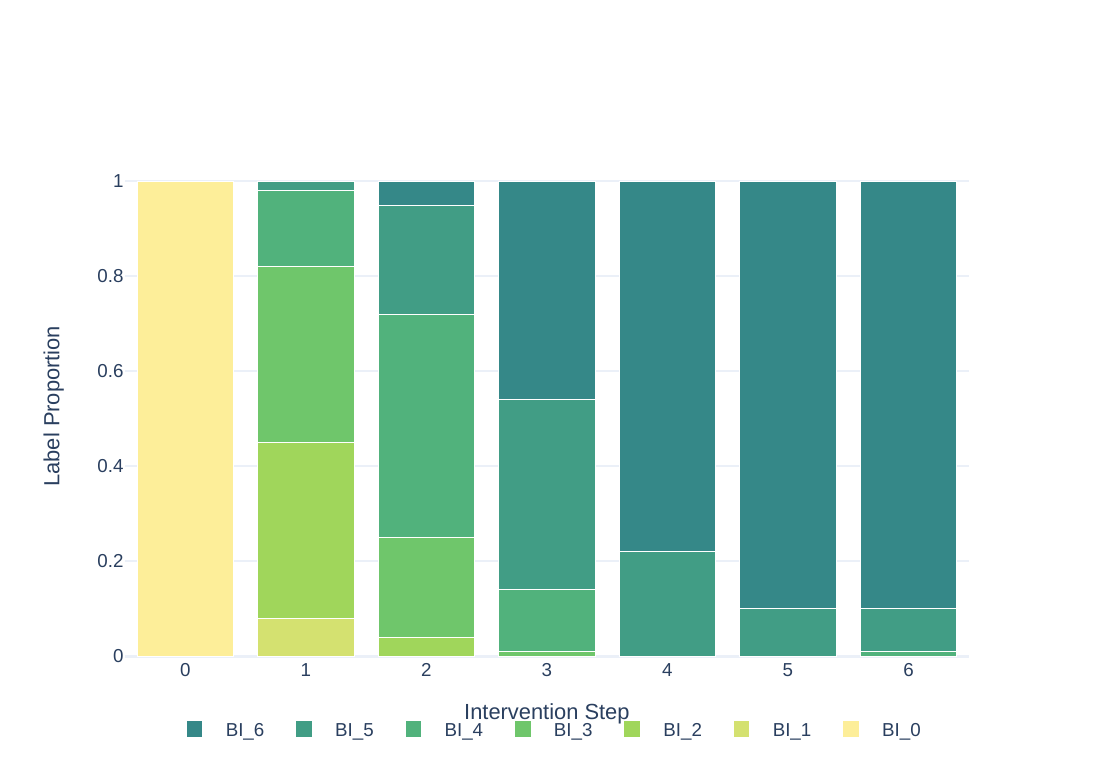}
  \caption{LF for Qwen1.5-7B.}
\end{subfigure}
\begin{subfigure}{.4\textwidth}
  \centering
  \includegraphics[width=\columnwidth]{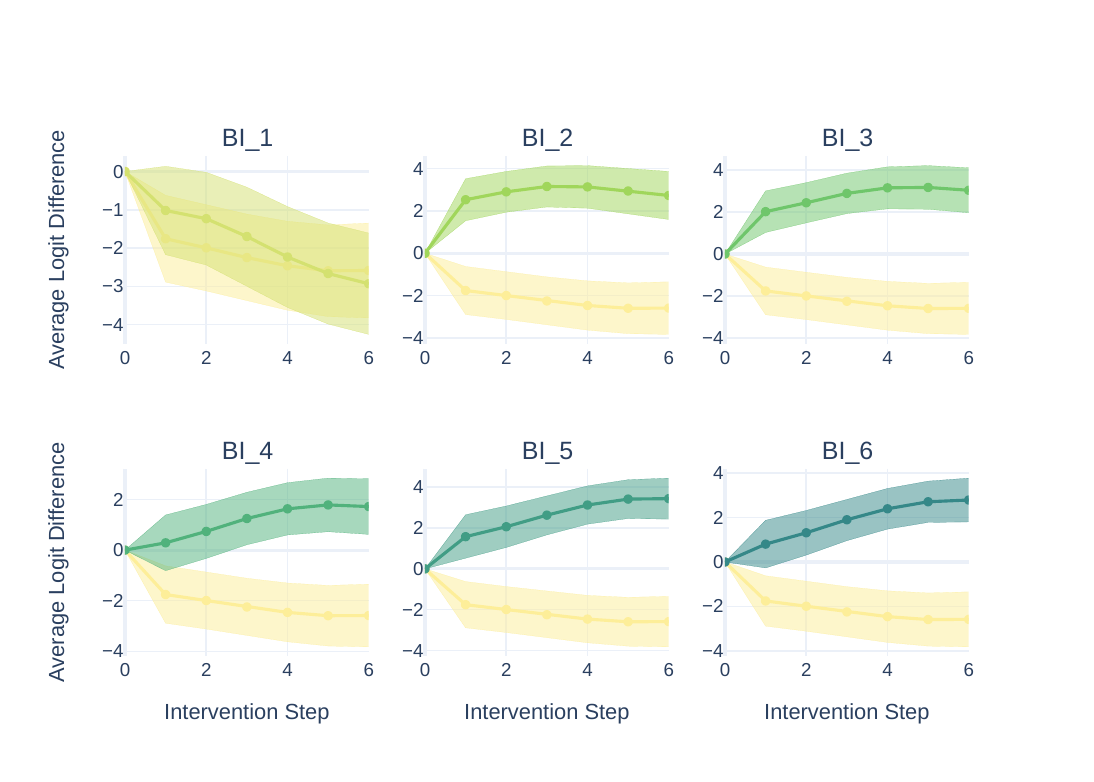}
  \caption{LD for Pythia-6.9B.}
\end{subfigure}%
\begin{subfigure}{.4\textwidth}
  \centering
  \includegraphics[width=\columnwidth]{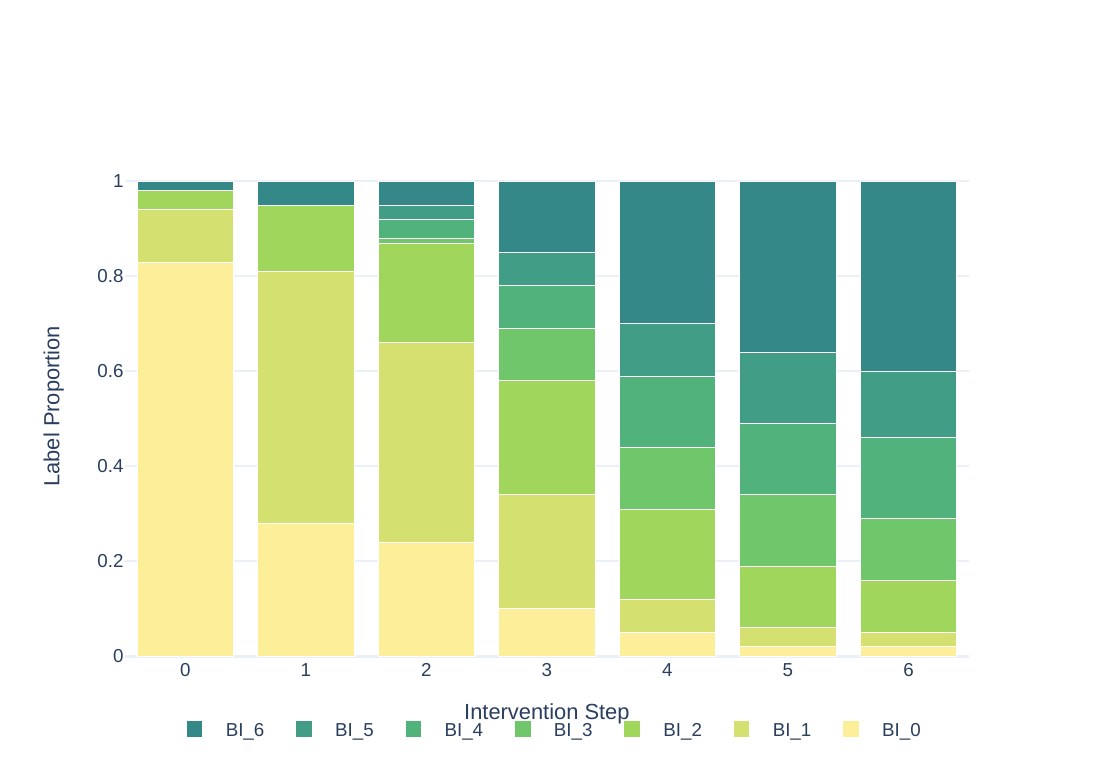}
  \caption{LF for Pythia-6.9B.}
\end{subfigure}
\caption{Logit Difference (LD) and Logit Flip (LF) for activation patching on the entity tracking dataset (i.e., r: is\_in).}
\label{fig:other_LM_results}
\end{figure*}

\clearpage
\subsection{Case Study on Llama2-7B}
\label{sec:case_study}
\begin{table*}[h]
\centering
\scalebox{0.65}{
\begin{tabular}{lccccccc}
\hline
 & &\multicolumn{6}{c}{\makecell[c]{\textbf{Answer for \# Step}}} \\\cmidrule(lr){3-8}
\textbf{Context} & \textbf{Query} & 1 & 2 & 3 & 4 & 5 & 6 \\\cmidrule(lr){1-8}
 \makecell[l]{The \textcolor{CBI1}{bug}is sold by person Esta,\\ the \textcolor{CBI2}{spawn} is sold by person Fritz,\\ the \textcolor{CBI3}{wine} is sold by person Inga,\\ the \textcolor{CBI4}{paste} is sold by person Ward,\\ the \textcolor{CBI5}{poison} is sold by person Albert,\\ the \textcolor{CBI6}{crow} is sold by person Davis,\\ the \textcolor{CBI7}{nest} is sold by person Val .} & 
 \makecell[l]{Person Esta\\ is selling the}& \textcolor{CBI2}{spawn} & \textcolor{CBI3}{wine} & \textcolor{CBI4}{paste} & \textcolor{CBI5}{poison} & \textcolor{CBI6}{crow} & \textcolor{CBI7}{nest} \\\cmidrule(lr){1-2}\cmidrule(lr){2-8}
\makecell[l]{The \textcolor{CBI1}{virus} is sold by person Anna,\\ the \textcolor{CBI2}{fur} is sold by person Earl,\\ the \textcolor{CBI3}{pill} is sold by person Flor,\\ the \textcolor{CBI4}{bean} is sold by person Roy,\\ the \textcolor{CBI5}{spawn} is sold by person Kam,\\ the \textcolor{CBI6}{farm} is sold by person Young,\\ the \textcolor{CBI7}{sheep} is sold by person Billy.} & 
 \makecell[l]{Person Anna\\ is selling the}& \textcolor{CBI2}{fur} & \textcolor{CBI3}{pill} & \textcolor{CBI5}{spawn} & \textcolor{CBI5}{spawn} & \textcolor{CBI6}{farm} & \textcolor{CBI7}{sheep} \\\cmidrule(lr){1-2}\cmidrule(lr){2-8}
\makecell[l]{The \textcolor{CBI1}{root} is sold by person Carl,\\ the \textcolor{CBI2}{mouse} is sold by person Marco,\\ the \textcolor{CBI3}{fruit} is sold by person Luke,\\ the \textcolor{CBI4}{bug} is sold by person Paul,\\ the \textcolor{CBI5}{grass} is sold by person Inga,\\ the \textcolor{CBI6}{pie} is sold by person Pok,\\ the \textcolor{CBI7}{cookie} is sold by person George.} & 
 \makecell[l]{Person Carl\\ is selling the}& \textcolor{CBI2}{mouse} & \textcolor{CBI3}{fruit} & \textcolor{CBI4}{bug} & \textcolor{CBI5}{grass} & \textcolor{CBI7}{cookie} & \textcolor{CBI7}{cookie}
\\\hline
\end{tabular}
}
\caption{Attributes inferred by Llama2-7B as a result of directed activation patching along OI-PC on the dataset of ``r: sell'', where color denotes the BI.}
\label{tab:ent_track_sell}
\end{table*}
\begin{table*}[h]
\centering
\scalebox{0.65}{
\begin{tabular}{lccccccc}
\hline
 & &\multicolumn{6}{c}{\makecell[c]{\textbf{Answer for \# Step}}} \\\cmidrule(lr){3-8}
\textbf{Context} & \textbf{Query} & 1 & 2 & 3 & 4 & 5 & 6 \\\cmidrule(lr){1-8}
 \makecell[l]{The \textcolor{CBI1}{carbon} is applied by person Wei,\\ the \textcolor{CBI2}{liquid} is applied by person Season,\\ the \textcolor{CBI3}{bath} is applied by person Robert,\\ the \textcolor{CBI4}{fog} is applied by person Daniel,\\ the \textcolor{CBI5}{heavy} is applied by person Roma,\\ the \textcolor{CBI6}{motor} is applied by person Ara,\\ the \textcolor{CBI7}{pool} is applied by person Jorge } & 
 \makecell[l]{Person Wei\\ applies the}& \textcolor{CBI2}{liquid} & \textcolor{CBI3}{bath} & \textcolor{CBI4}{fog} & \textcolor{CBI6}{motor} & \textcolor{CBI6}{motor} & \textcolor{CBI7}{pool} \\\cmidrule(lr){1-2}\cmidrule(lr){2-8}
\makecell[l]{The \textcolor{CBI1}{rain} is applied by person Kurt,\\ the \textcolor{CBI2}{gauge} is applied by person Jon,\\ the \textcolor{CBI3}{dust} is applied by person Newton,\\ the \textcolor{CBI4}{jet} is applied by person Dan,\\ the \textcolor{CBI5}{floor} is applied by person Alfred,\\ the \textcolor{CBI6}{low} is applied by person Mike,\\ the \textcolor{CBI7}{basket} is applied by person April} & 
 \makecell[l]{Person Kurt\\ applies the}& \textcolor{CBI2}{gauge} & \textcolor{CBI3}{dust} & \textcolor{CBI4}{jet} & \textcolor{CBI5}{floor} & \textcolor{CBI7}{basket} & \textcolor{CBI7}{basket} \\\cmidrule(lr){1-2}\cmidrule(lr){2-8}
\makecell[l]{The \textcolor{CBI1}{lamp} is applied by person Angel,\\ the \textcolor{CBI2}{bucket} is applied by person Carl,\\ the \textcolor{CBI3}{canvas} is applied by person Bert,\\ the \textcolor{CBI4}{cargo} is applied by person Otto,\\ the \textcolor{CBI5}{plain} is applied by person Johnny,\\ the \textcolor{CBI6}{floor} is applied by person John,\\ the \textcolor{CBI7}{heavy} is applied by person Era.} & 
 \makecell[l]{Person Angel\\ applies the}& \textcolor{CBI2}{bucket} & \textcolor{CBI3}{canvas} & \textcolor{CBI4}{cargo} & \textcolor{CBI5}{plain} & \textcolor{CBI6}{floor} & \textcolor{CBI7}{heavy}
\\\hline
\end{tabular}
}
\caption{Attributes inferred by Llama2-7B as a result of directed activation patching along OI-PC on the dataset of ``r: apply'', where color denotes the BI.}
\label{tab:ent_track_apply}
\end{table*}
\begin{table*}[h]
\centering
\scalebox{0.65}{
\begin{tabular}{lccccccc}
\hline
 & &\multicolumn{6}{c}{\makecell[c]{\textbf{Answer for \# Step}}} \\\cmidrule(lr){3-8}
\textbf{Context} & \textbf{Query} & 1 & 2 & 3 & 4 & 5 & 6 \\\cmidrule(lr){1-8}
 \makecell[l]{The \textcolor{CBI1}{lip} is moved by person Mack,\\ the \textcolor{CBI2}{tract} is moved by person Sommer,\\ the \textcolor{CBI3}{pen} is moved by person Son,\\ the \textcolor{CBI4}{tip} is moved by person August,\\ the \textcolor{CBI5}{bat} is moved by person Monte,\\ the \textcolor{CBI6}{socket} is moved by person Marco,\\ the \textcolor{CBI7}{hook} is moved by person Paul.} & 
 \makecell[l]{Person Mack\\ moved the}& \textcolor{CBI2}{tract} & \textcolor{CBI3}{pen} & \textcolor{CBI4}{tip} & \textcolor{CBI5}{bat} & \textcolor{CBI7}{hook} & \textcolor{CBI7}{hook} \\\cmidrule(lr){1-2}\cmidrule(lr){2-8}
\makecell[l]{The \textcolor{CBI1}{mask} is moved by person Jules,\\ the \textcolor{CBI2}{timer} is moved by person Ward,\\ the \textcolor{CBI3}{bullet} is moved by person Ana,\\ the \textcolor{CBI4}{eye} is moved by person Val,\\ the \textcolor{CBI5}{button} is moved by person Andy,\\ the \textcolor{CBI6}{lock} is moved by person Arnold,\\ the \textcolor{CBI7}{colon} is moved by person Betty.} & 
 \makecell[l]{Person Jules\\ moved the}& \textcolor{CBI2}{timer} & \textcolor{CBI3}{bullet} & \textcolor{CBI5}{button} & \textcolor{CBI6}{lock} & \textcolor{CBI6}{lock} & \textcolor{CBI7}{colon} \\\cmidrule(lr){1-2}\cmidrule(lr){2-8}
\makecell[l]{The \textcolor{CBI1}{mask} is moved by person Cole,\\ the \textcolor{CBI2}{neck} is moved by person Donald,\\ the \textcolor{CBI3}{pad} is moved by person Beth,\\ the \textcolor{CBI4}{cone} is moved by person Jorge,\\ the \textcolor{CBI5}{tail} is moved by person Lou,\\ the \textcolor{CBI6}{thread} is moved by person Alfred,\\ the \textcolor{CBI7}{toe} is moved by person Edward.} & 
 \makecell[l]{Person Cole\\ moved the}& \textcolor{CBI2}{neck} & \textcolor{CBI3}{pad} & \textcolor{CBI4}{cone} & \textcolor{CBI5}{tail} & \textcolor{CBI7}{toe} & \textcolor{CBI7}{toe}
\\\hline
\end{tabular}
}
\caption{Attributes inferred by Llama2-7B as a result of directed activation patching along OI-PC on the dataset of ``r: move'', where color denotes the BI.}
\label{tab:ent_track_move}
\end{table*}
\begin{table*}[h]
\centering
\scalebox{0.65}{
\begin{tabular}{lccccccc}
\hline
 & &\multicolumn{6}{c}{\makecell[c]{\textbf{Answer for \# Step}}} \\\cmidrule(lr){3-8}
\textbf{Context} & \textbf{Query} & 1 & 2 & 3 & 4 & 5 & 6 \\\cmidrule(lr){1-8}
 \makecell[l]{The \textcolor{CBI1}{creature} is brought by person Tam,\\ the \textcolor{CBI2}{guitar} is brought by person Frank,\\ the \textcolor{CBI3}{dress} is brought by person Stuart,\\ the \textcolor{CBI4}{block} is brought by person Victor,\\ the \textcolor{CBI5}{brain} is brought by person David,\\ the \textcolor{CBI6}{coffee} is brought by person Mack,\\ the \textcolor{CBI7}{radio} is brought by person Roger.} & 
 \makecell[l]{Person Tam\\ brings the}& \textcolor{CBI2}{guitar} & \textcolor{CBI3}{dress} & \textcolor{CBI4}{block} & \textcolor{CBI5}{brain} & \textcolor{CBI6}{coffee} & \textcolor{CBI7}{radio} \\\cmidrule(lr){1-2}\cmidrule(lr){2-8}
\makecell[l]{The \textcolor{CBI1}{boat} is brought by person Luke,\\ the \textcolor{CBI2}{pipe} is brought by person Clara,\\ the \textcolor{CBI3}{pot} is brought by person Han,\\ the \textcolor{CBI4}{bill} is brought by person Chi,\\ the \textcolor{CBI5}{milk} is brought by person Scott,\\ the \textcolor{CBI6}{card} is brought by person Henry,\\ the \textcolor{CBI7}{brick} is brought by person Morris} & 
 \makecell[l]{Person Luke\\ brings the}& \textcolor{CBI2}{pipe} & \textcolor{CBI3}{pot} & \textcolor{CBI4}{bill} & \textcolor{CBI6}{card} & \textcolor{CBI7}{brick} & \textcolor{CBI7}{brick} \\\cmidrule(lr){1-2}\cmidrule(lr){2-8}
\makecell[l]{The \textcolor{CBI1}{fan} is brought by person Van,\\ the \textcolor{CBI2}{note} is brought by person Clara,\\ the \textcolor{CBI3}{block} is brought by person Alex,\\ the \textcolor{CBI4}{newspaper} is brought by person Peg,\\ the \textcolor{CBI5}{crown} is brought by person Jan,\\ the \textcolor{CBI6}{car} is brought by person Pok,\\ the \textcolor{CBI7}{magnet} is brought by person Golden.} & 
 \makecell[l]{Person Van\\ brings the}& \textcolor{CBI2}{note} & \textcolor{CBI3}{block} & \textcolor{CBI5}{crown} & \textcolor{CBI6}{car} & \textcolor{CBI7}{magnet} & \textcolor{CBI7}{magnet}
\\\hline
\end{tabular}
}
\caption{Attributes inferred by Llama2-7B as a result of directed activation patching along OI-PC on the dataset of ``r: bring'', where color denotes the BI.}
\label{tab:ent_track_bring}
\end{table*}
\begin{table*}[h]
\centering
\scalebox{0.65}{
\begin{tabular}{lccccccc}
\hline
 & &\multicolumn{6}{c}{\makecell[c]{\textbf{Answer for \# Step}}} \\\cmidrule(lr){3-8}
\textbf{Context} & \textbf{Query} & 1 & 2 & 3 & 4 & 5 & 6 \\\cmidrule(lr){1-8}
 \makecell[l]{The \textcolor{CBI1}{load} is pushed by person Mike,\\ the \textcolor{CBI2}{atom} is pushed by person Mira,\\ the \textcolor{CBI3}{tin} is pushed by person Juli,\\ the \textcolor{CBI4}{stud} is pushed by person Sam,\\ the \textcolor{CBI5}{sedan} is pushed by person Pia,\\ the \textcolor{CBI6}{bath} is pushed by person Leo,\\ the \textcolor{CBI7}{growth} is pushed by person Pat.} & 
 \makecell[l]{Person Mike\\ pushes the}& \textcolor{CBI2}{atom} & \textcolor{CBI3}{tin} & \textcolor{CBI4}{stud} & \textcolor{CBI6}{bath} & \textcolor{CBI7}{growth} & \textcolor{CBI7}{growth} \\\cmidrule(lr){1-2}\cmidrule(lr){2-8}
\makecell[l]{The \textcolor{CBI1}{mud} is pushed by person Thomas,\\ the \textcolor{CBI2}{heavy} is pushed by person Ralph,\\ the \textcolor{CBI3}{tile} is pushed by person Pierre,\\ the \textcolor{CBI4}{import} is pushed by person Perry,\\ the \textcolor{CBI5}{arm} is pushed by person Robert,\\ the \textcolor{CBI6}{lung} is pushed by person Kurt,\\ the \textcolor{CBI7}{cabin} is pushed by person Ernest.} & 
 \makecell[l]{Person Thomas\\ pushes the}& \textcolor{CBI2}{heavy} & \textcolor{CBI3}{tile} & \textcolor{CBI4}{import} & \textcolor{CBI5}{arm} & \textcolor{CBI7}{cabin} & \textcolor{CBI7}{cabin} \\\cmidrule(lr){1-2}\cmidrule(lr){2-8}
\makecell[l]{The \textcolor{CBI1}{bed} is pushed by person Fran,\\ the \textcolor{CBI2}{lever} is pushed by person Lan,\\ the \textcolor{CBI3}{cord} is pushed by person Paris,\\ the \textcolor{CBI4}{vent} is pushed by person Gene,\\ the \textcolor{CBI5}{thumb} is pushed by person Marie,\\ the \textcolor{CBI6}{mouth} is pushed by person Asia,\\ the \textcolor{CBI7}{ear} is pushed by person Lang.} & 
 \makecell[l]{Person Fran\\ pushes the}& \textcolor{CBI2}{lever} & \textcolor{CBI3}{cord} & \textcolor{CBI4}{vent} & \textcolor{CBI5}{thumb} & \textcolor{CBI5}{thumb} & \textcolor{CBI7}{ear}
\\\hline
\end{tabular}
}
\caption{Attributes inferred by Llama2-7B as a result of directed activation patching along OI-PC on the dataset of ``r: push'', where color denotes the BI.}
\label{tab:ent_track_push}
\end{table*}

\clearpage
\subsection{Activation Steering on Llama2-7B}
\label{sec:as_all}
\begin{figure*}[h]
    \centering % <-- added
\begin{subfigure}{0.33\textwidth}
  \includegraphics[width=\linewidth]{graph/steer_logit_diff_llama_0.pdf}
  \caption{r: is\_in}
  \label{fig:stl1}
\end{subfigure}\hfil % <-- added
\begin{subfigure}{0.33\textwidth}
  \includegraphics[width=\linewidth]{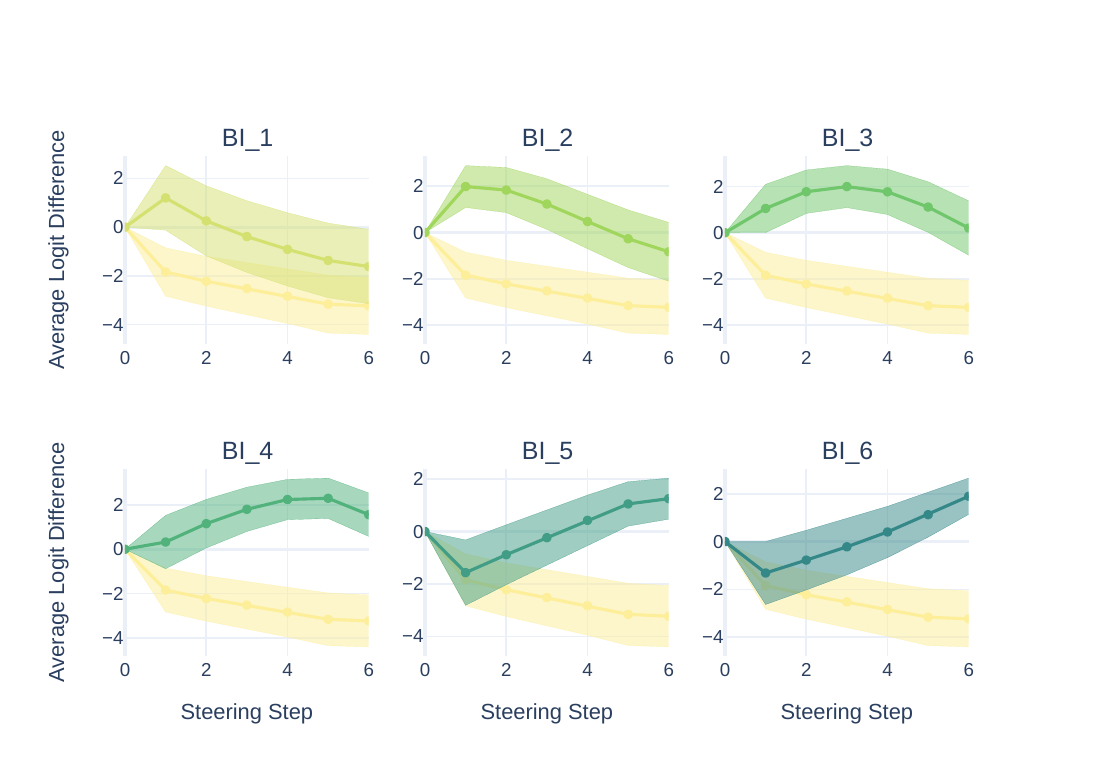}
  \caption{r: sell}
  \label{fig:stl2}
\end{subfigure}\hfil % <-- added
\begin{subfigure}{0.33\textwidth}
  \includegraphics[width=\linewidth]{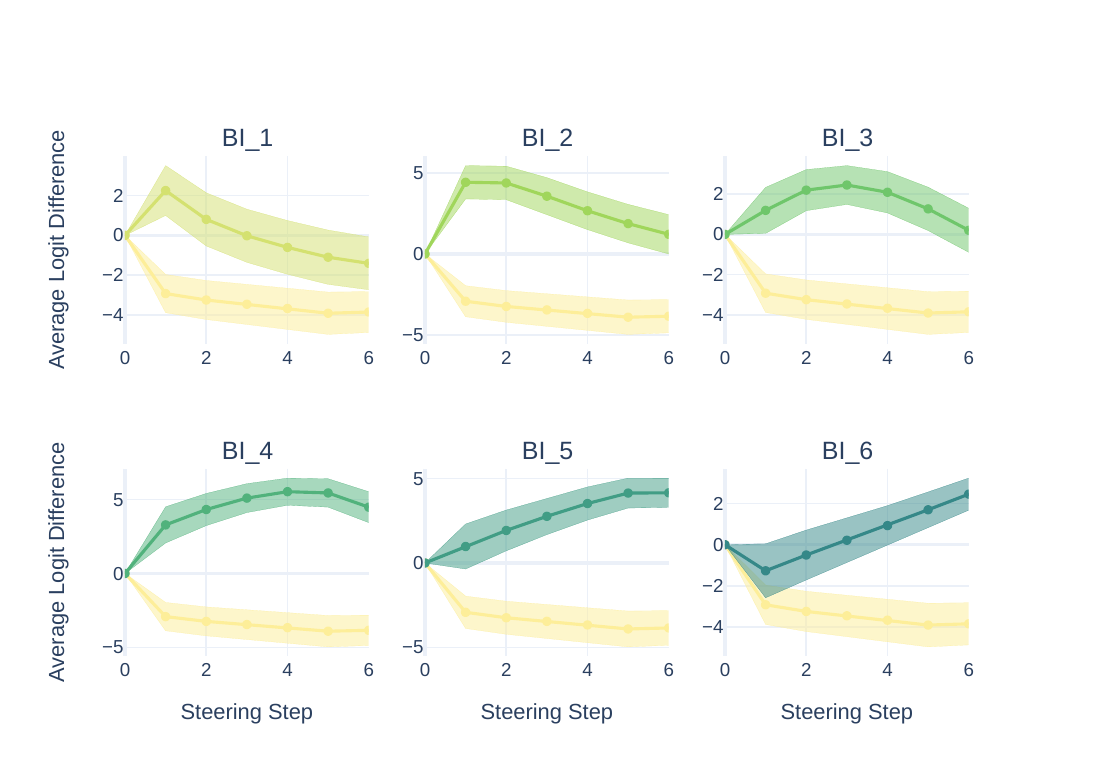}
  \caption{r: apply}
  \label{fig:stl3}
\end{subfigure}

\medskip
\begin{subfigure}{0.33\textwidth}
  \includegraphics[width=\linewidth]{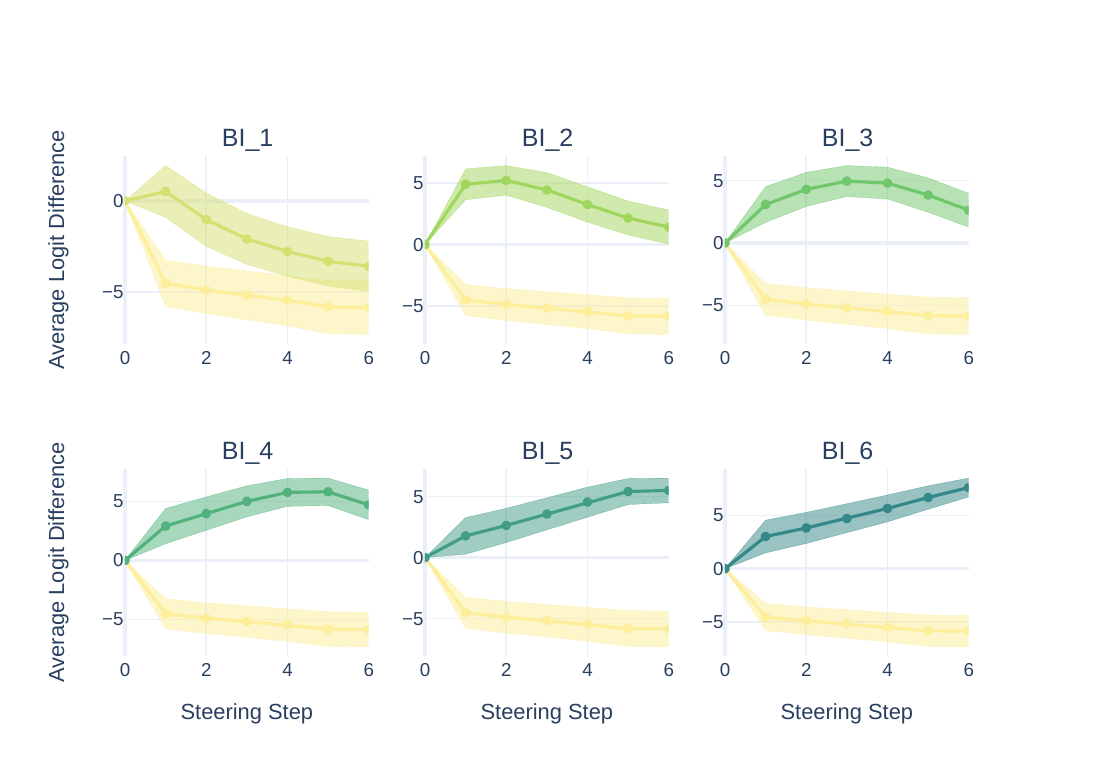}
  \caption{r: move}
  \label{fig:stl4}
\end{subfigure}\hfil % <-- added
\begin{subfigure}{0.33\textwidth}
  \includegraphics[width=\linewidth]{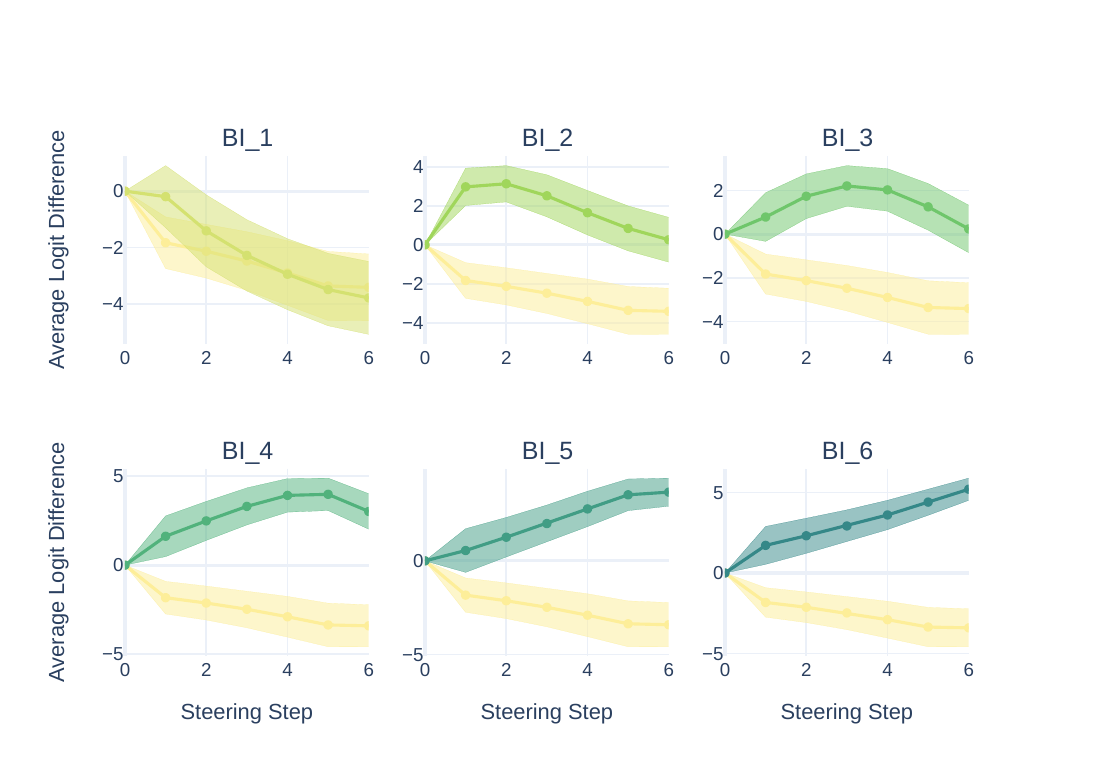}
  \caption{r:bring}
  \label{fig:stl5}
\end{subfigure}\hfil % <-- added
\begin{subfigure}{0.33\textwidth}
  \includegraphics[width=\linewidth]{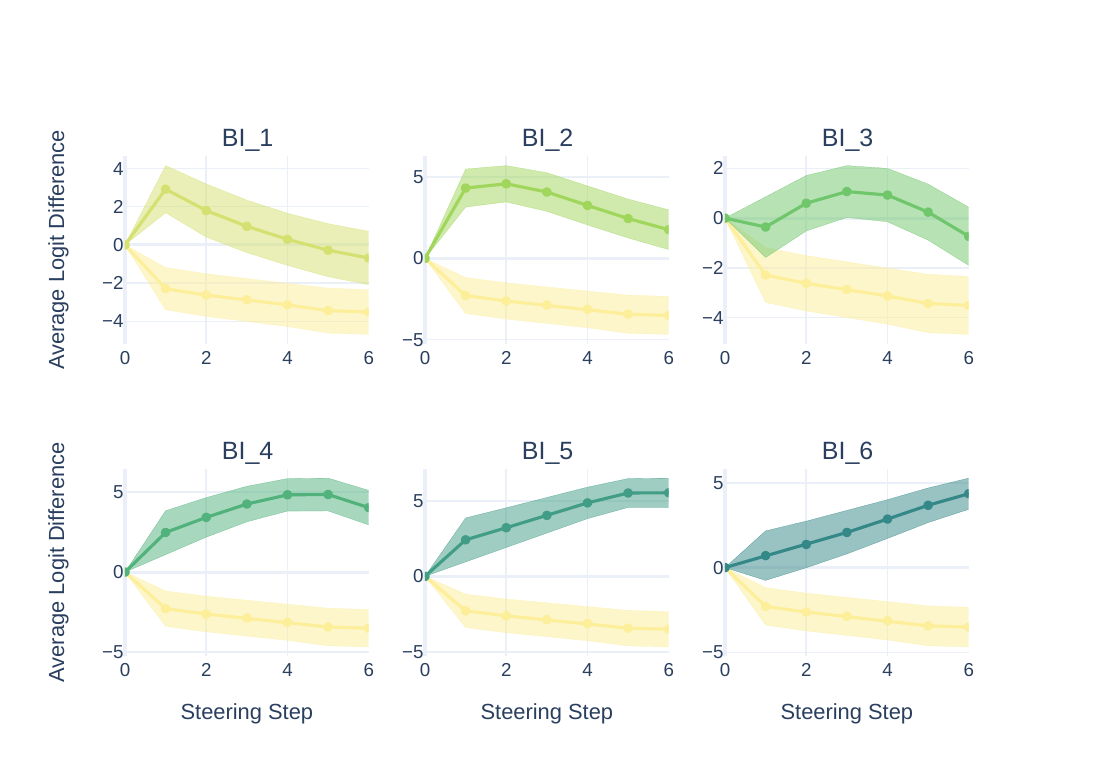}
  \caption{r: push}
  \label{fig:stl6}
\end{subfigure}
\caption{Logit Difference (LD) for OI subspace based activation steering across datasets on Llama2-7B, where x axis represents the intervention of $\mathbf{s}_{0 \rightarrow bi}$ on the activation of $e_0$. Here, $l = 8$ and $\alpha = 1.25$.}
\label{fig:stld_a}
\end{figure*}

\begin{figure*}[!h]
    \centering % <-- added
\begin{subfigure}{0.33\textwidth}
  \includegraphics[width=\linewidth]{graph/steer_flip_llama_0.pdf}
  \caption{r: is\_in}
  \label{fig:sf1}
\end{subfigure}\hfil % <-- added
\begin{subfigure}{0.33\textwidth}
  \includegraphics[width=\linewidth]{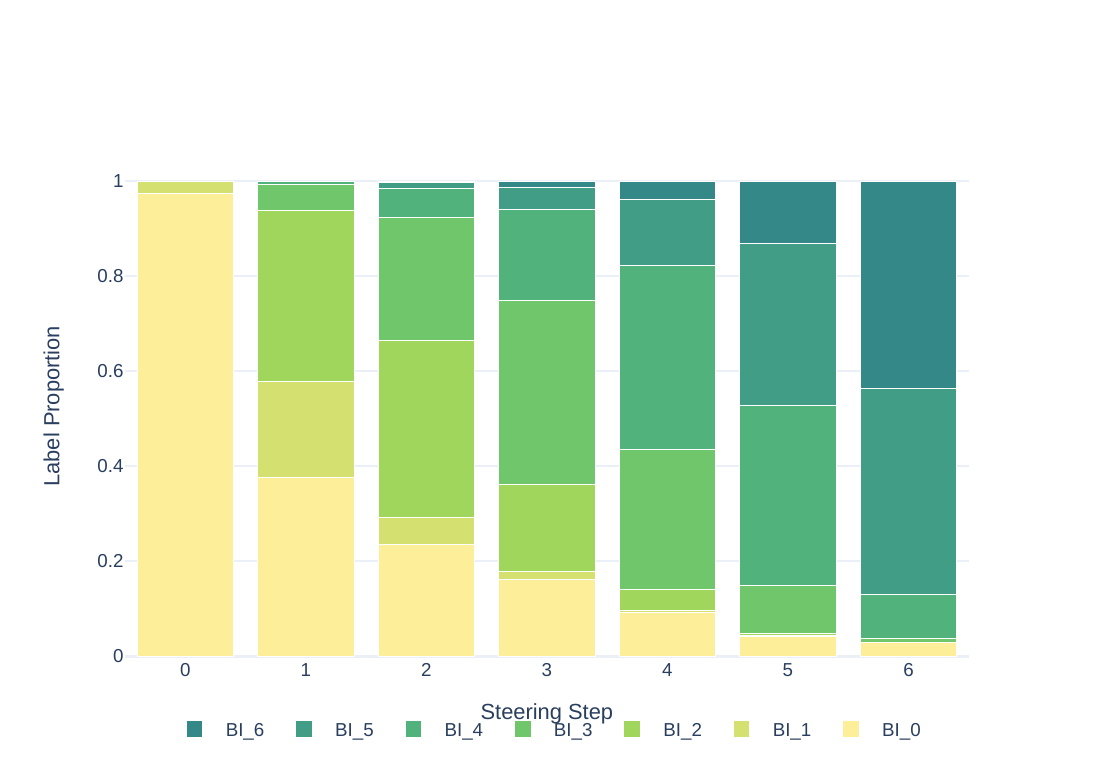}
  \caption{r: sell}
  \label{fig:sf2}
\end{subfigure}\hfil % <-- added
\begin{subfigure}{0.33\textwidth}
  \includegraphics[width=\linewidth]{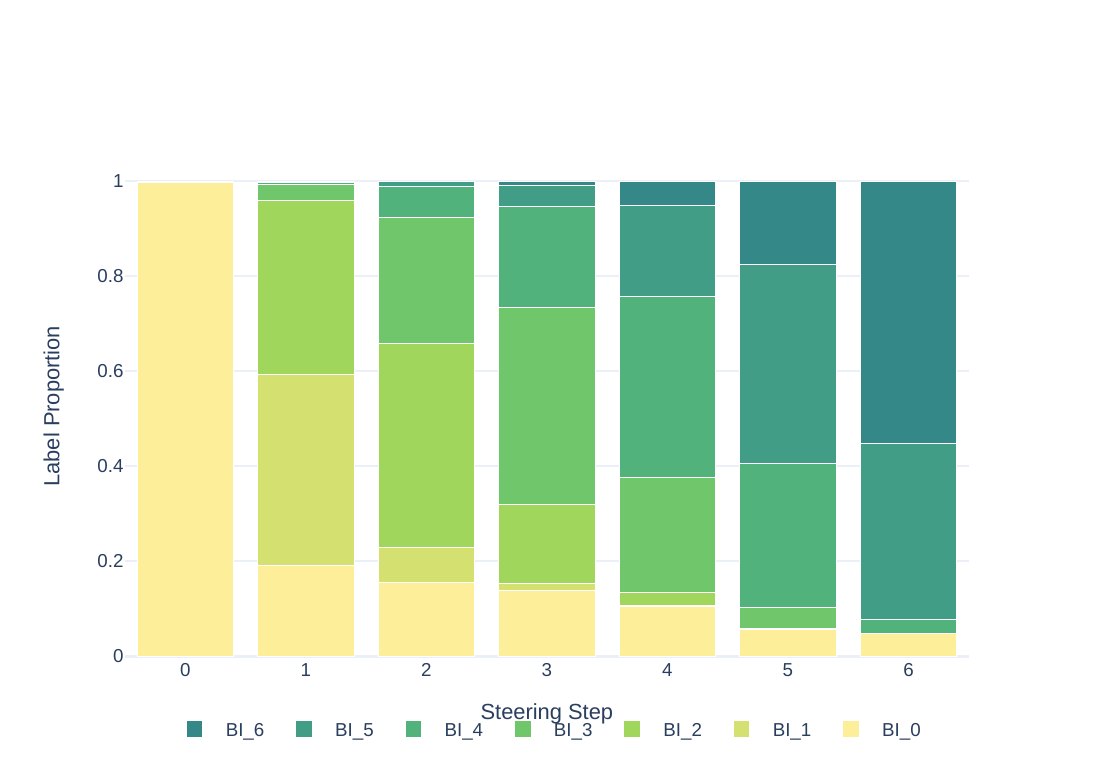}
  \caption{r: apply}
  \label{fig:sf3}
\end{subfigure}

\medskip
\begin{subfigure}{0.33\textwidth}
  \includegraphics[width=\linewidth]{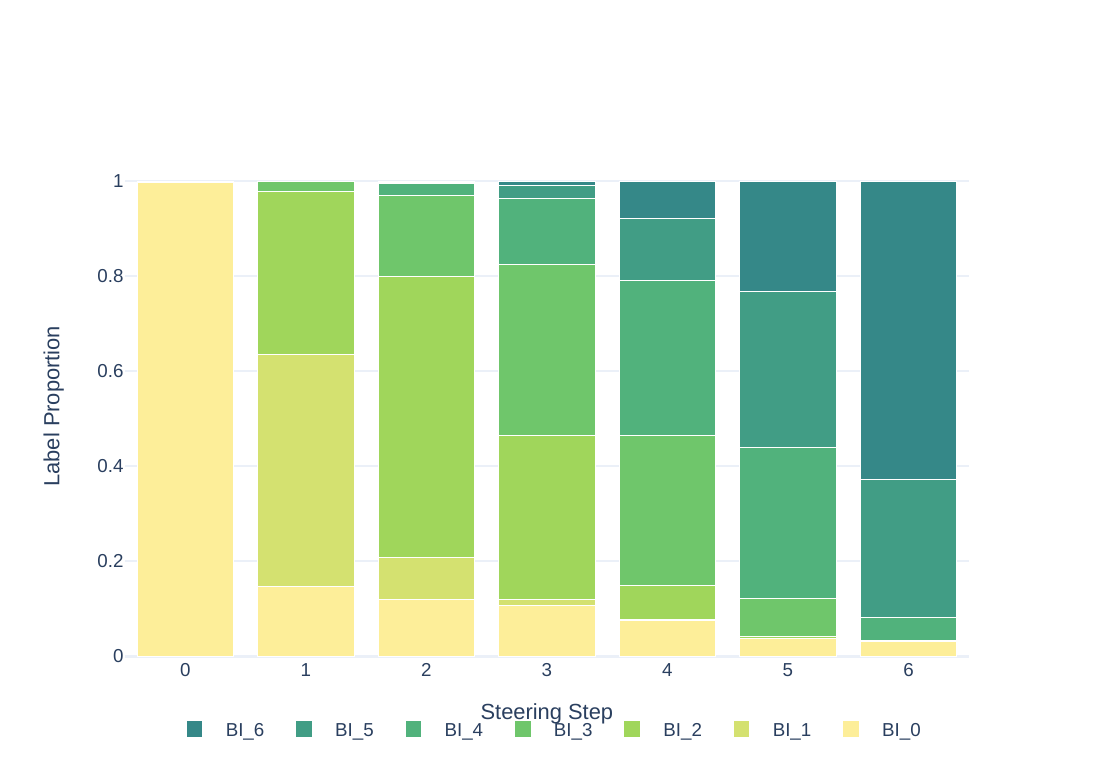}
  \caption{r: move}
  \label{fig:sf4}
\end{subfigure}\hfil % <-- added
\begin{subfigure}{0.33\textwidth}
  \includegraphics[width=\linewidth]{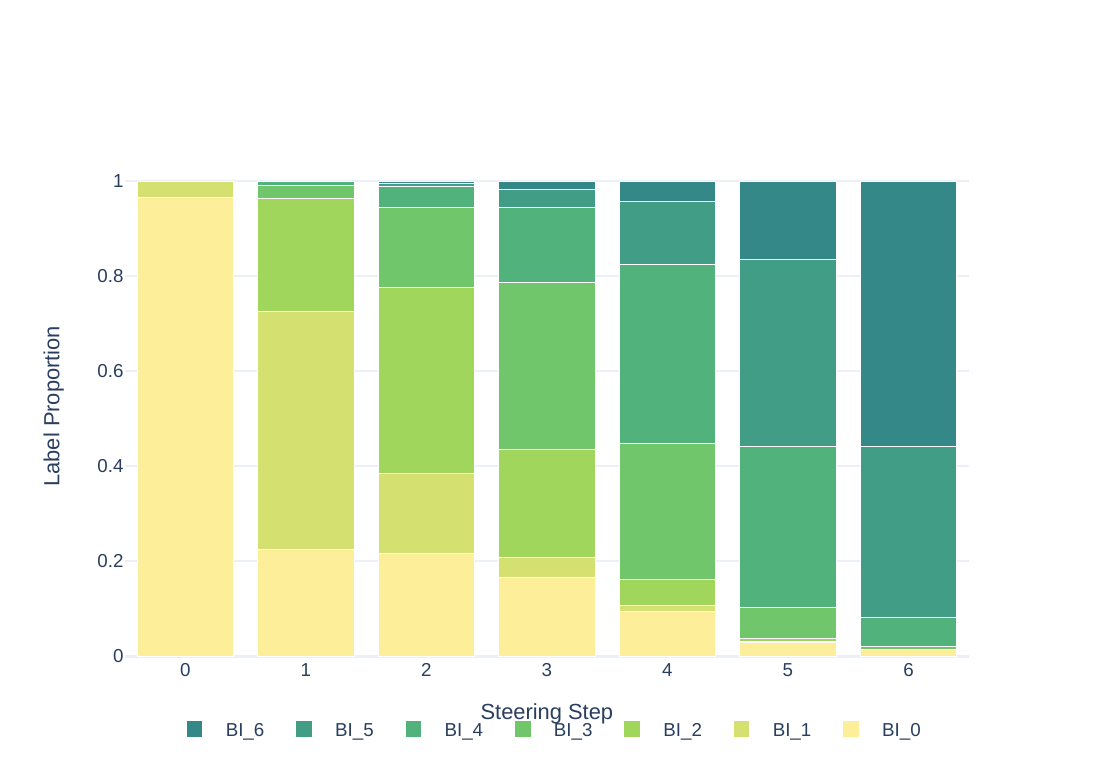}
  \caption{r: bring}
  \label{fig:sf5}
\end{subfigure}\hfil % <-- added
\begin{subfigure}{0.33\textwidth}
  \includegraphics[width=\linewidth]{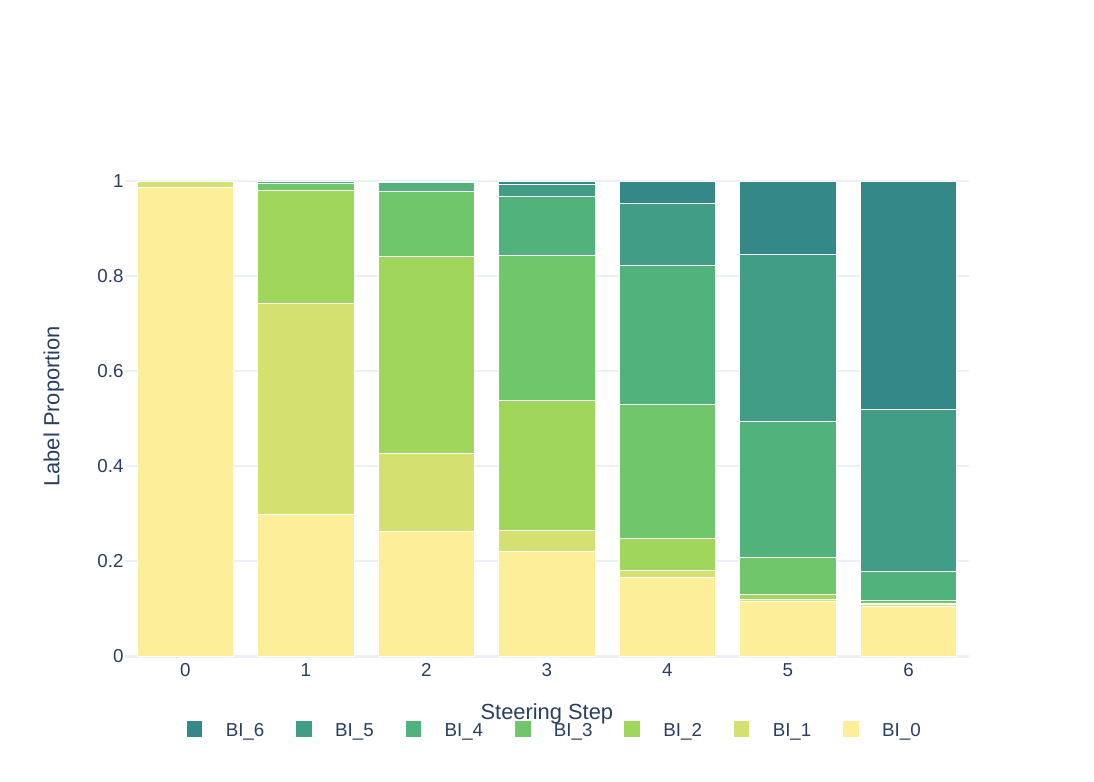}
  \caption{r: push}
  \label{fig:sf6}
\end{subfigure}
\caption{Logit flip for OI subspace based activation steering across datasets on Llama2-7B, where x axis represents the intervention of $\mathbf{s}_{0 \rightarrow bi}$ on the activation of $e_0$.}
\label{fig:stlf_a}
\end{figure*}

\clearpage
\subsection{Results on Fine-Tuned LM}
\label{sec:ap_float}

\begin{figure*}[h]
    \centering % <-- added
\begin{subfigure}{0.33\textwidth}
  \includegraphics[width=\linewidth]{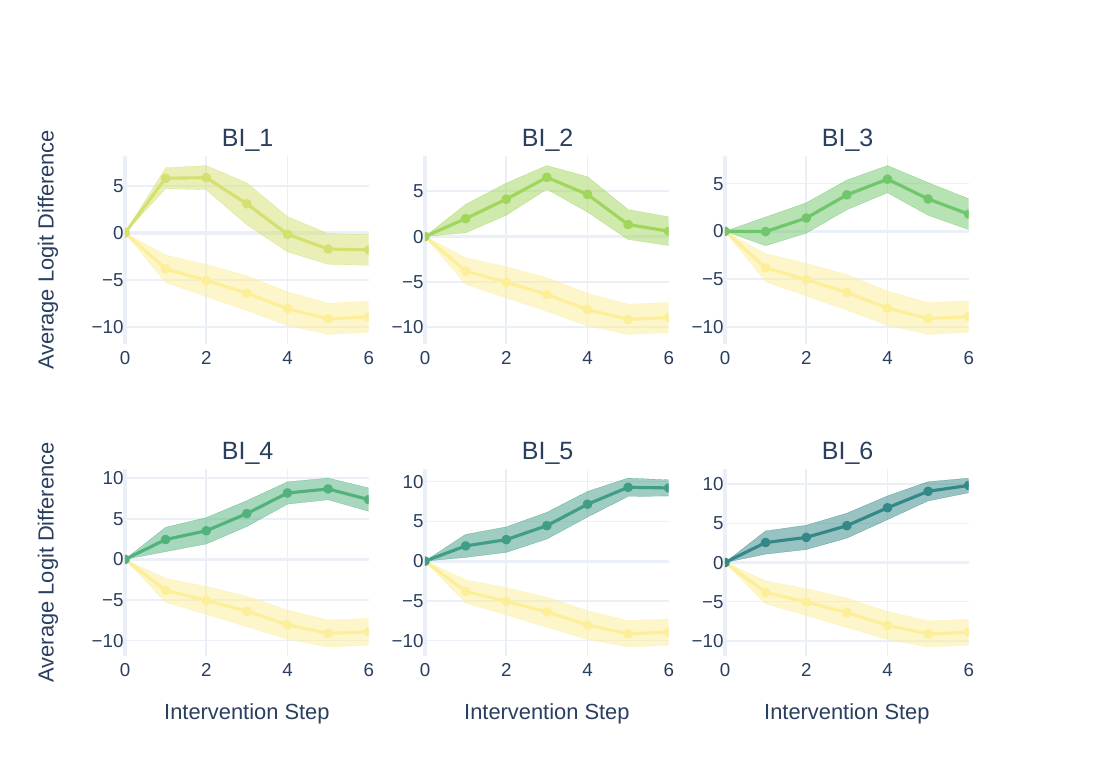}
  \caption{r: is\_in}
  \label{fig:fll1}
\end{subfigure}\hfil % <-- added
\begin{subfigure}{0.33\textwidth}
  \includegraphics[width=\linewidth]{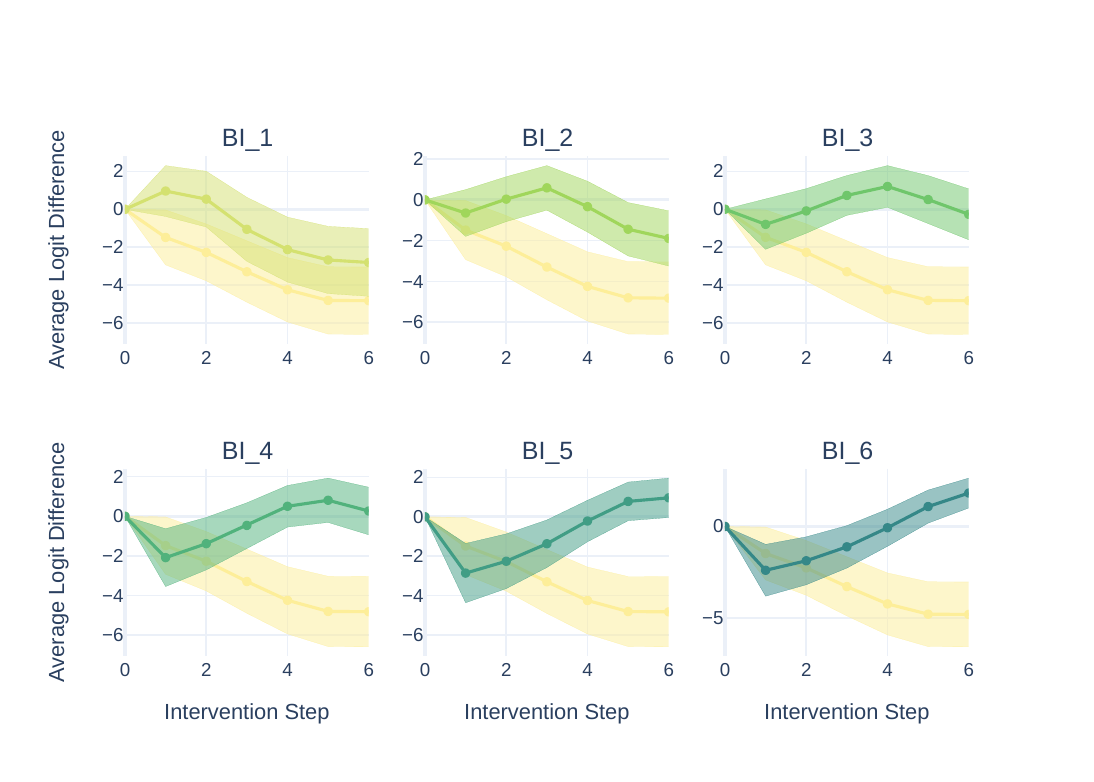}
  \caption{r: sell}
  \label{fig:fll2}
\end{subfigure}\hfil % <-- added
\begin{subfigure}{0.33\textwidth}
  \includegraphics[width=\linewidth]{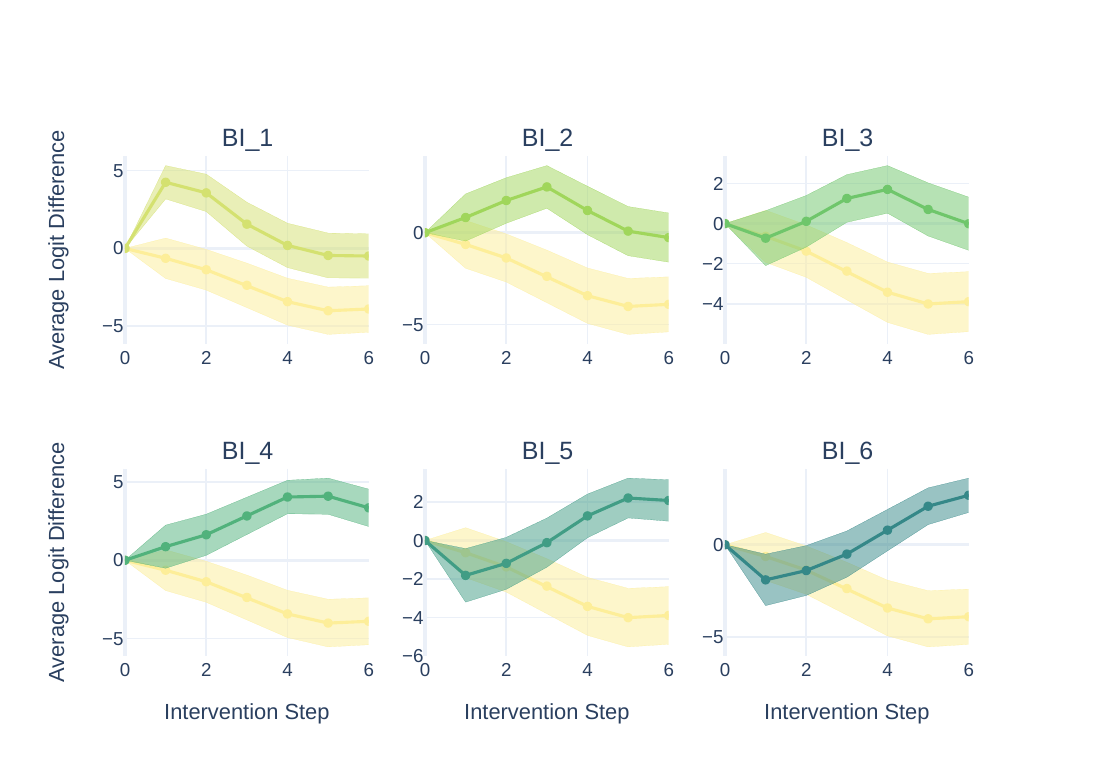}
  \caption{r: apply}
  \label{fig:fll3}
\end{subfigure}

\medskip
\begin{subfigure}{0.33\textwidth}
  \includegraphics[width=\linewidth]{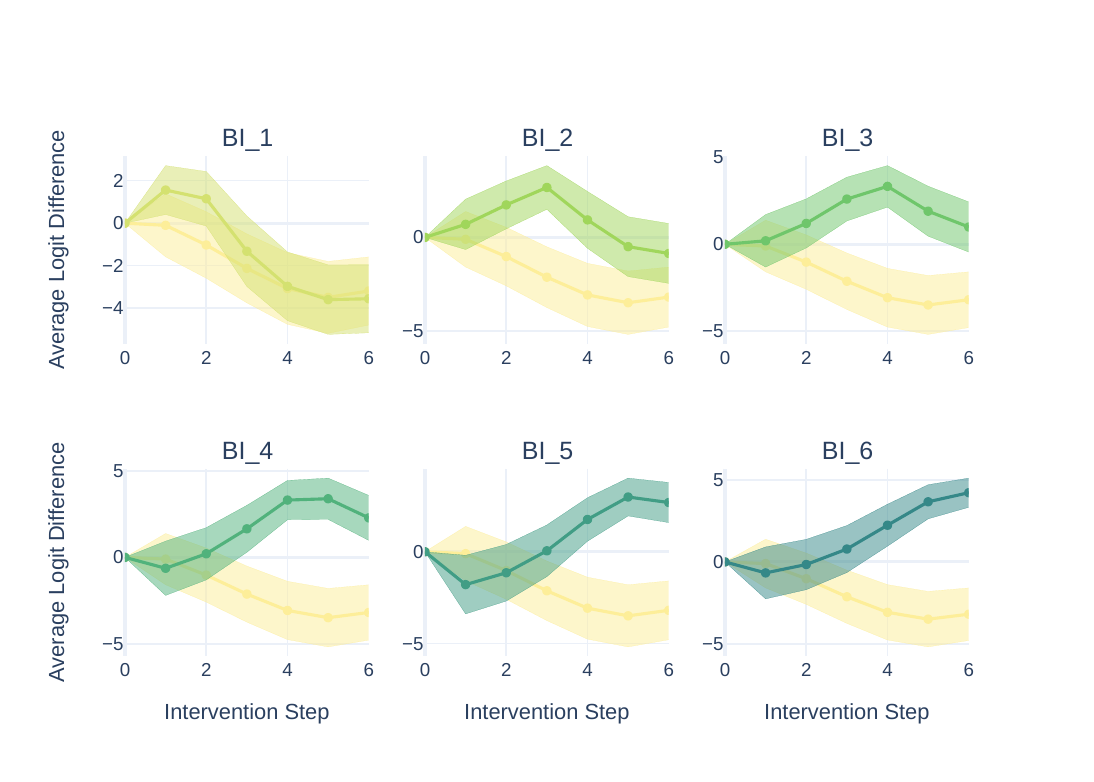}
  \caption{r: move}
  \label{fig:fll4}
\end{subfigure}\hfil % <-- added
\begin{subfigure}{0.33\textwidth}
  \includegraphics[width=\linewidth]{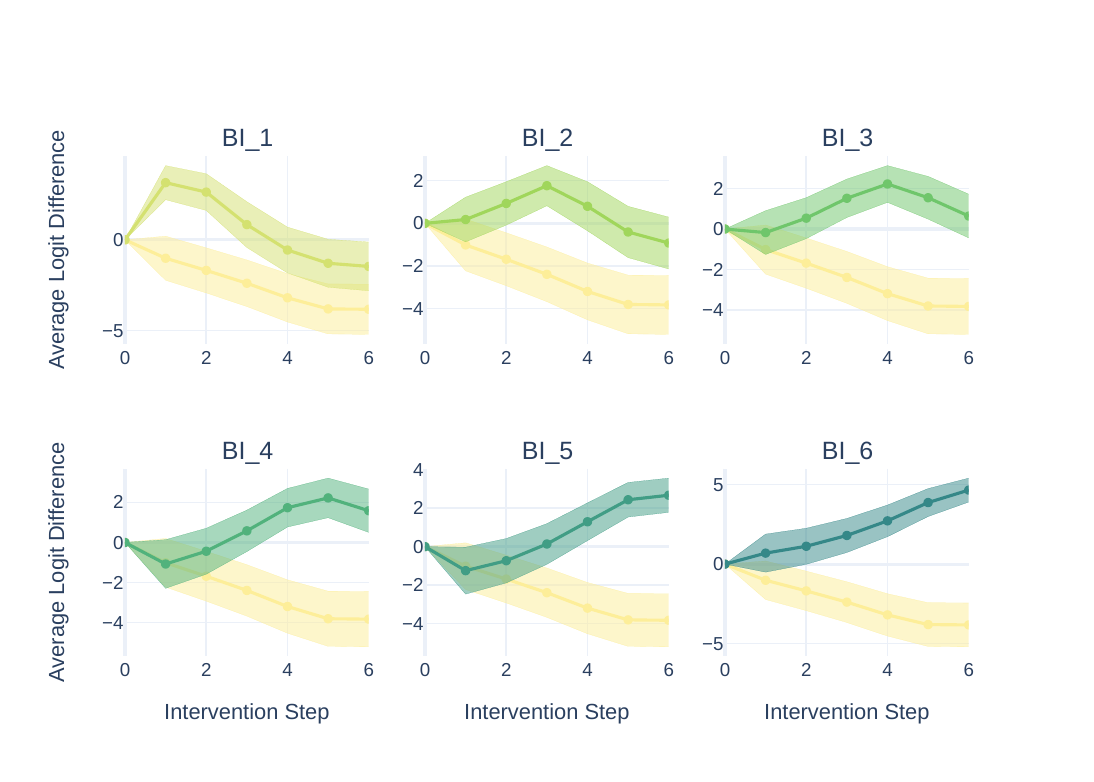}
  \caption{r:bring}
  \label{fig:fll5}
\end{subfigure}\hfil % <-- added
\begin{subfigure}{0.33\textwidth}
  \includegraphics[width=\linewidth]{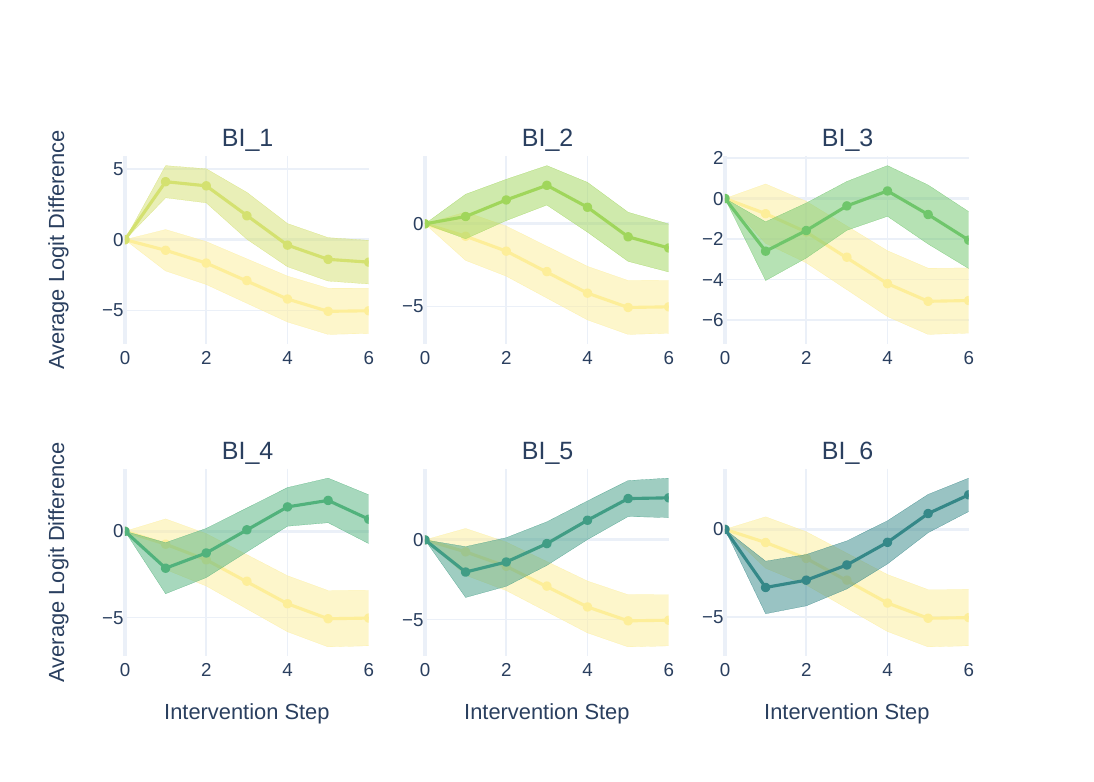}
  \caption{r: push}
  \label{fig:fll6}
\end{subfigure}
\caption{Logit Difference (LD) for OI-PC based intervention across datasets on Float-7B, where x axis denotes the number of intervention steps on $e_0$, y axis does the LD, BI\_i represents each target attribute and the light yellow bottom line indicates the LD of original attribute (i.e., $a_0$). Here, $l = 10$, $v = 2.55$,  and $\alpha = 5.0$.}
\label{fig:flld_a}
\end{figure*}

\begin{figure*}[h]
    \centering % <-- added
\begin{subfigure}{0.33\textwidth}
  \includegraphics[width=\linewidth]{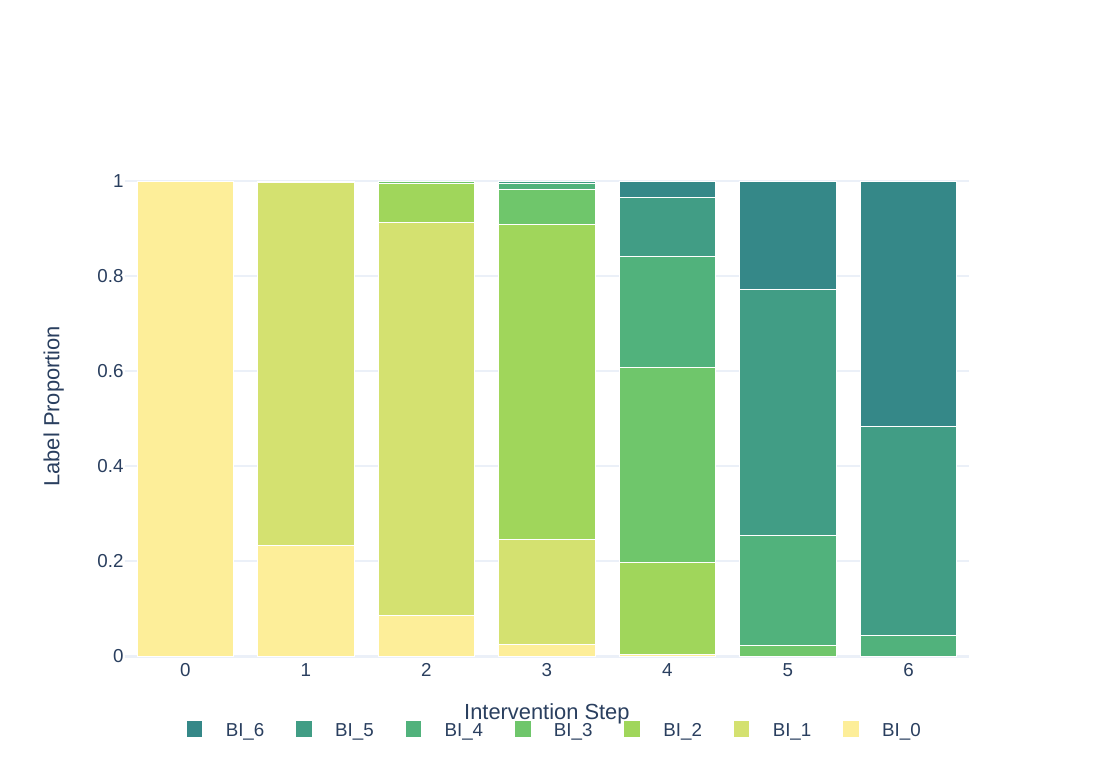}
  \caption{r: is\_in}
  \label{fig:flf1}
\end{subfigure}\hfil % <-- added
\begin{subfigure}{0.33\textwidth}
  \includegraphics[width=\linewidth]{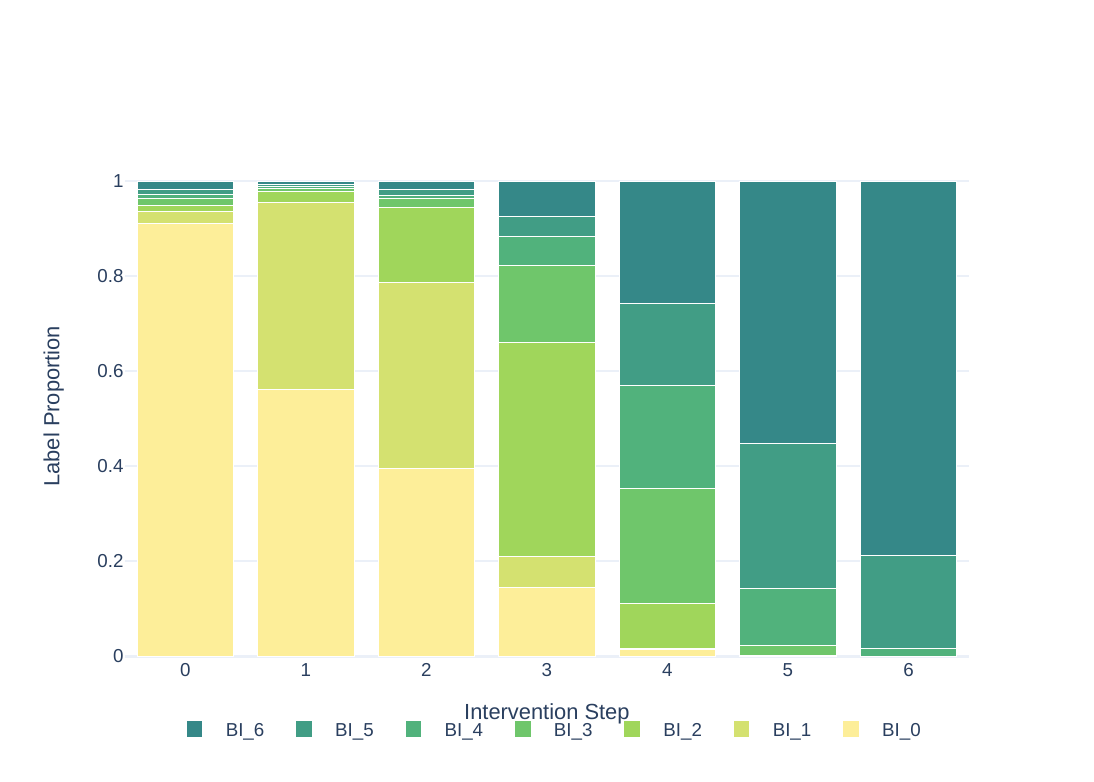}
  \caption{r: sell}
  \label{fig:flf2}
\end{subfigure}\hfil % <-- added
\begin{subfigure}{0.33\textwidth}
  \includegraphics[width=\linewidth]{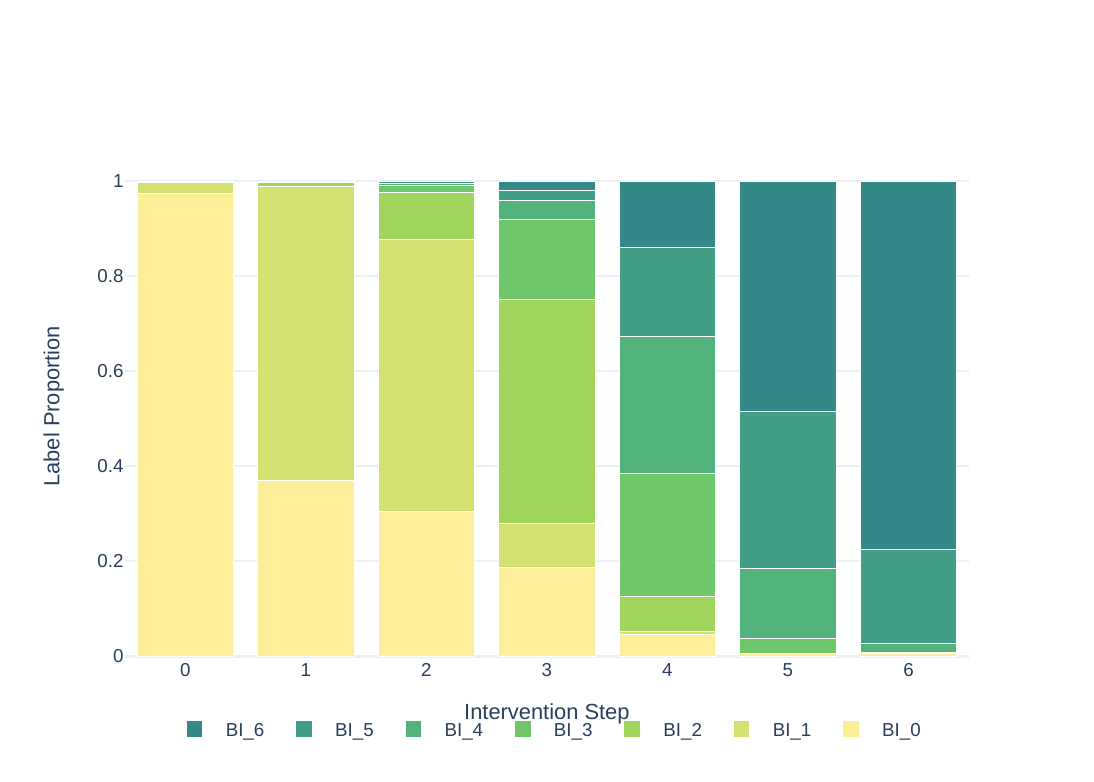}
  \caption{r: apply}
  \label{fig:flf3}
\end{subfigure}

\medskip
\begin{subfigure}{0.33\textwidth}
  \includegraphics[width=\linewidth]{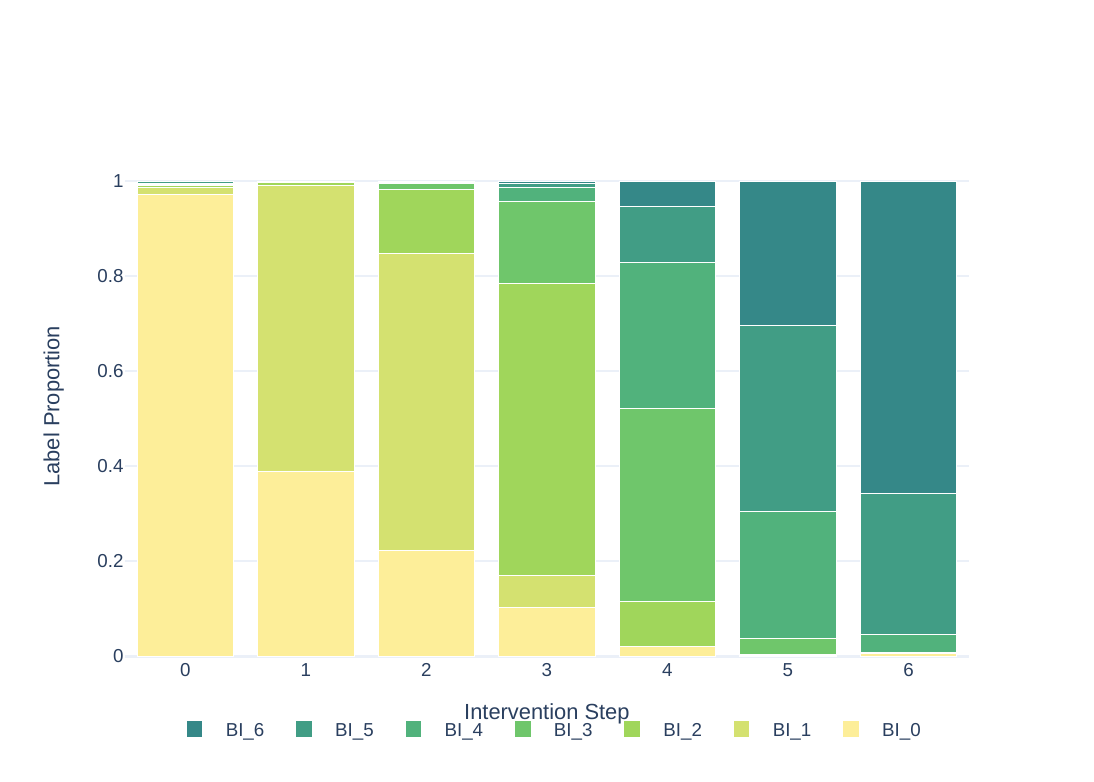}
  \caption{r: move}
  \label{fig:flf4}
\end{subfigure}\hfil % <-- added
\begin{subfigure}{0.33\textwidth}
  \includegraphics[width=\linewidth]{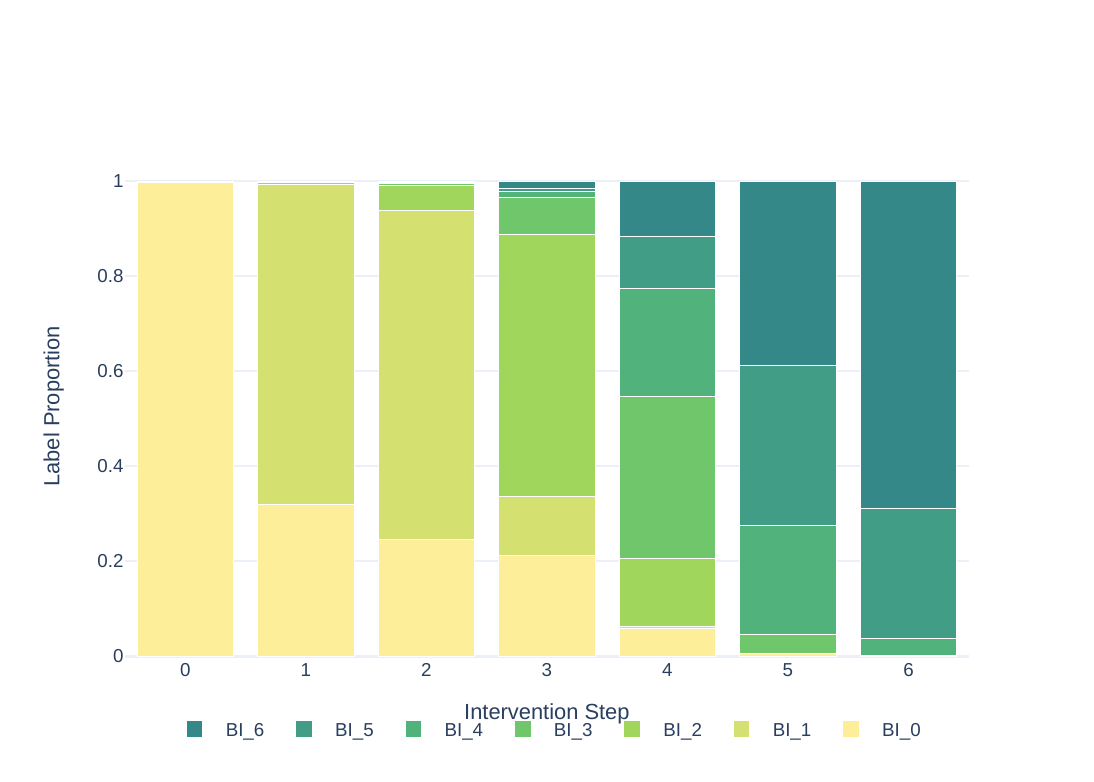}
  \caption{r: bring}
  \label{fig:flf5}
\end{subfigure}\hfil % <-- added
\begin{subfigure}{0.33\textwidth}
  \includegraphics[width=\linewidth]{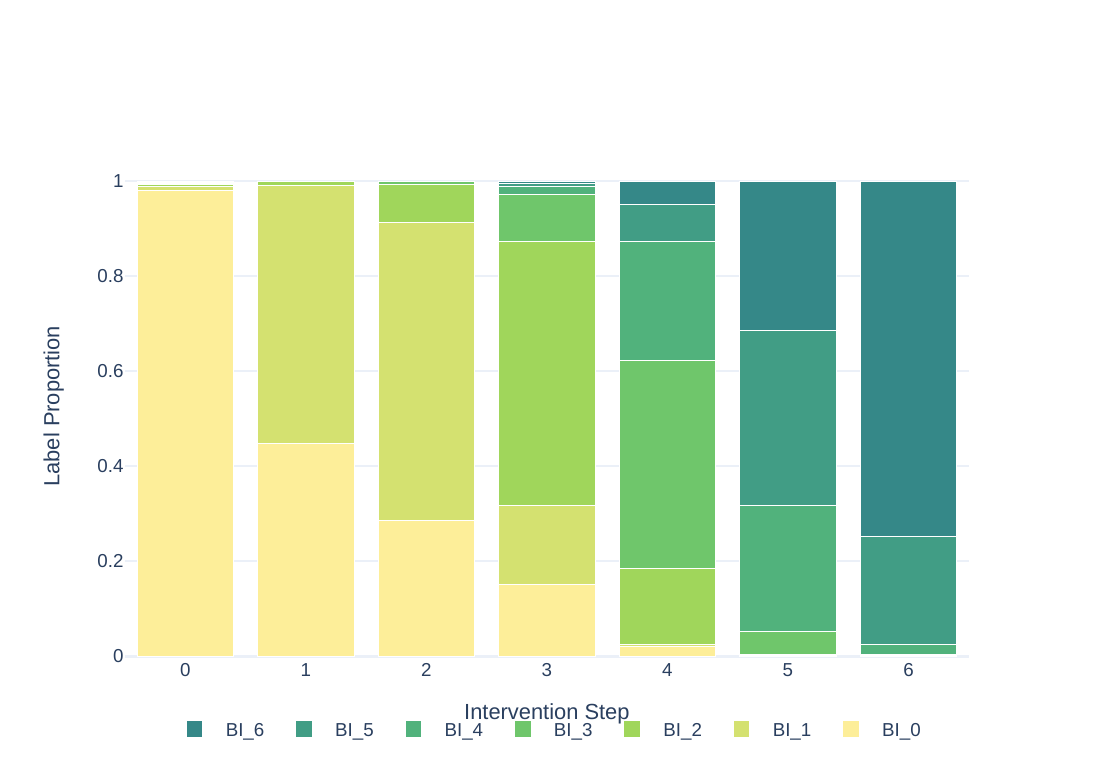}
  \caption{r: push}
  \label{fig:flf6}
\end{subfigure}
\caption{Logit flip for OI-PC based intervention across datasets on Float-7B, where x axis denotes the number of intervention steps on $e_0$, y axis does the proportion of each inferred attribute in model output.}
\label{fig:flf_a}
\end{figure*}

\citet{kramar2024atp,kim2024code} claim that the code fine-tuned LM, such as Float-7B~\cite{prakash2024fine} outperforms the pretrained LM on the entity tracking task~\cite{kim2023entity,prakash2024fine}. Since the code fine-tuned LM performs well on the entity tracking task that requires the OI subspace based computation, we hypothesize that OI subspace also exists in the code fine-tuned LM and the intervention along OI-PC will causally affect the model output. To prove the hypothesis, we conduct the intervention on Float-7B and show results in Figure~\ref{fig:flld_a} and Figure~\ref{fig:flf_a}. We found that the OI subspace based intervention on Float-7B achieves the similar results as on Llama2-7B, indicating that the OI subspace not only exists in the pretrained LM but also in the fine-tuned one. In addition, adding the same step value (i.e., $v$) on Float-7B will achieve higher LD value than Llama2-7B, indicating that the code fine-tuned LM is more sensitive to the OI subspace based intervention. For instance, the maximum LD of $a_4$ in the former is around $10$, and it is $2$ times larger than the one in the latter, which is around $5$. This might partially explains why the code fine-tuned LM performs better than the original one, because code fine-tuning might enhance the function of OI subspace so that it is more sensitive on the intervention and more effective on the in-context entity tracking task.

\clearpage
\subsection{Activation Patching on Llama3-8B}
\label{sec:llama3}
\begin{figure*}[h]
    \centering % <-- added
\begin{subfigure}{0.33\textwidth}
  \includegraphics[width=\linewidth]{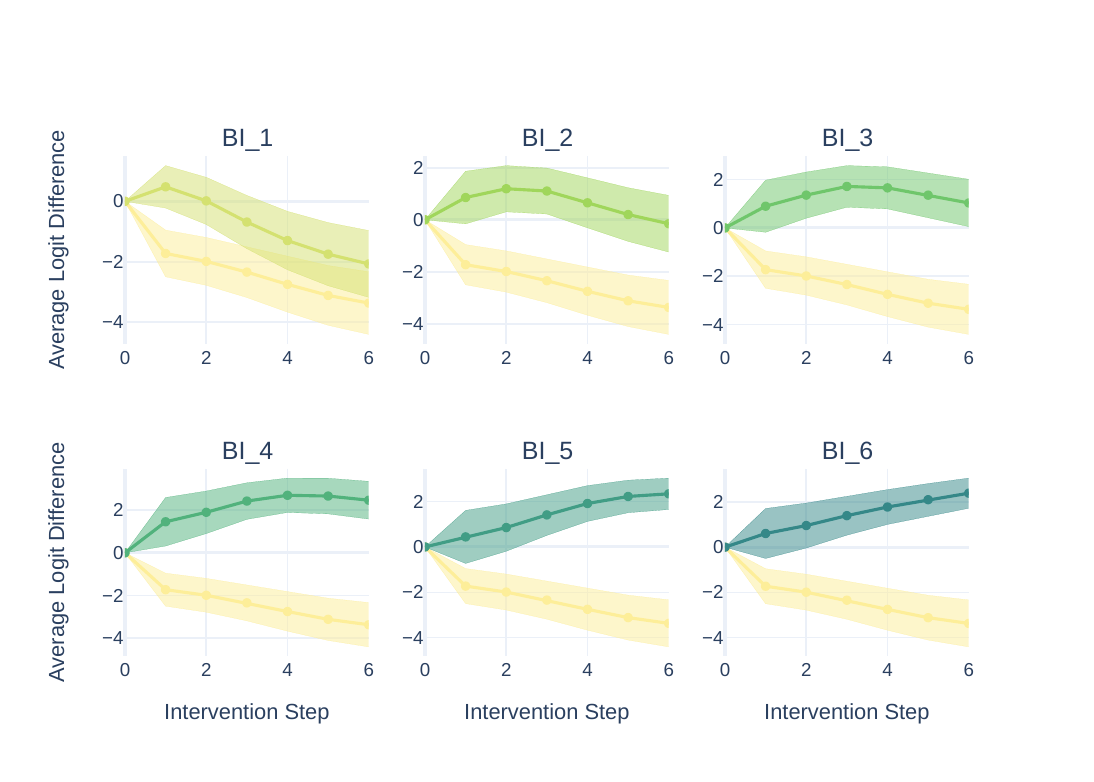}
  \caption{r: is\_in}
  \label{fig:la3l1}
\end{subfigure}\hfil % <-- added
\begin{subfigure}{0.33\textwidth}
  \includegraphics[width=\linewidth]{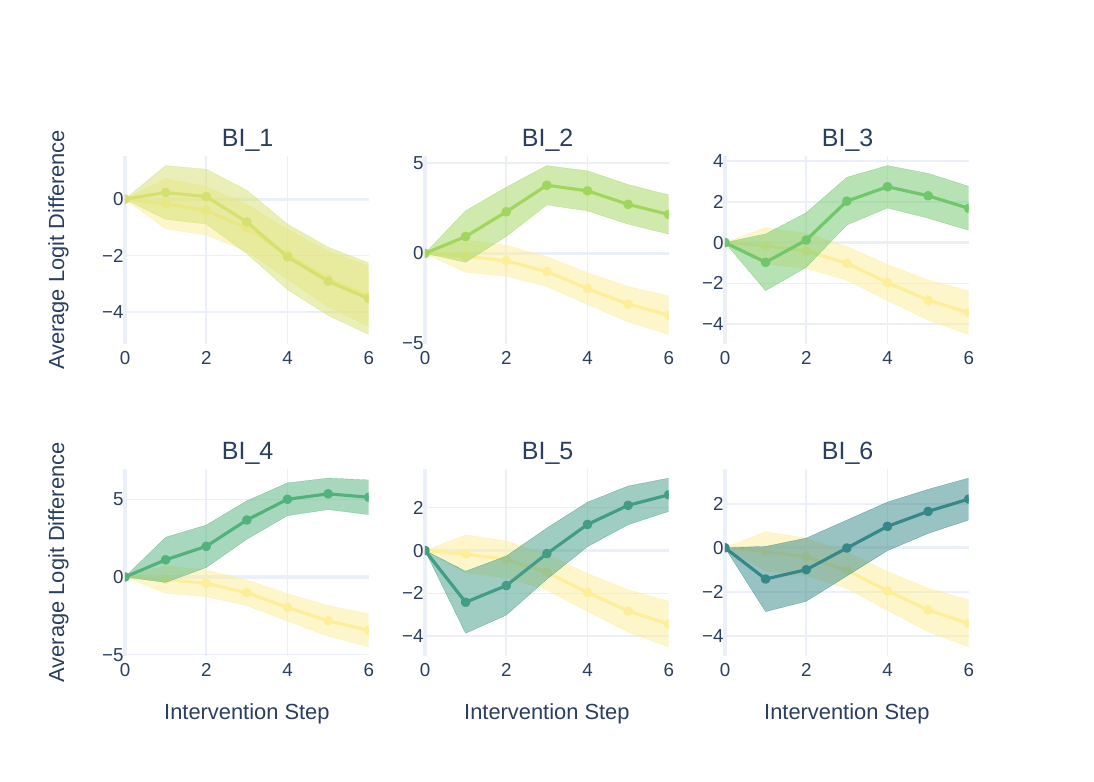}
  \caption{r: sell}
  \label{fig:la3l2}
\end{subfigure}\hfil % <-- added
\begin{subfigure}{0.33\textwidth}
  \includegraphics[width=\linewidth]{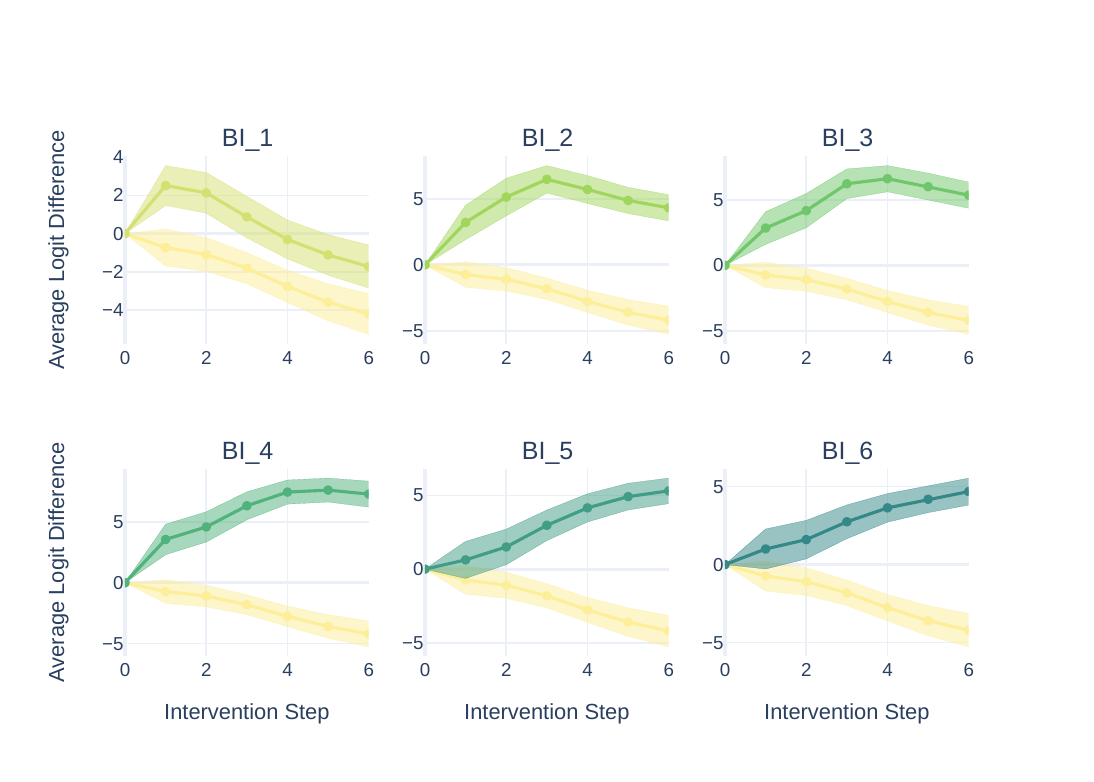}
  \caption{r: apply}
  \label{fig:la3l3}
\end{subfigure}

\medskip
\begin{subfigure}{0.33\textwidth}
  \includegraphics[width=\linewidth]{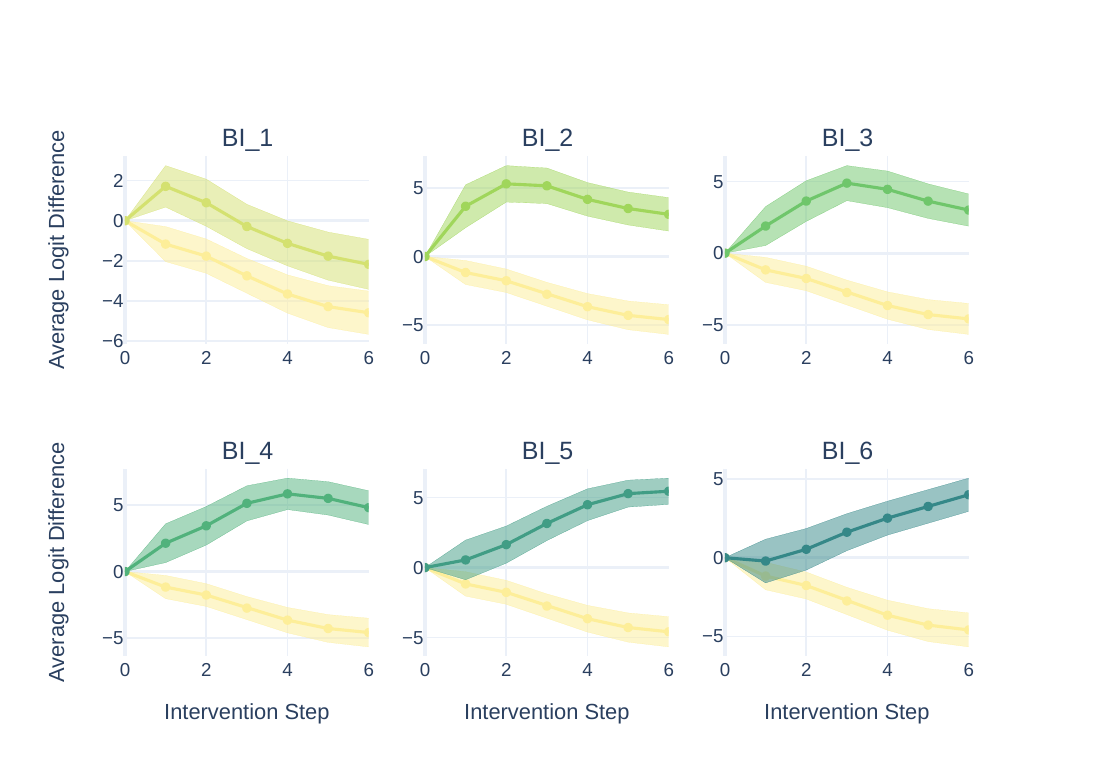}
  \caption{r: move}
  \label{fig:la3l4}
\end{subfigure}\hfil % <-- added
\begin{subfigure}{0.33\textwidth}
  \includegraphics[width=\linewidth]{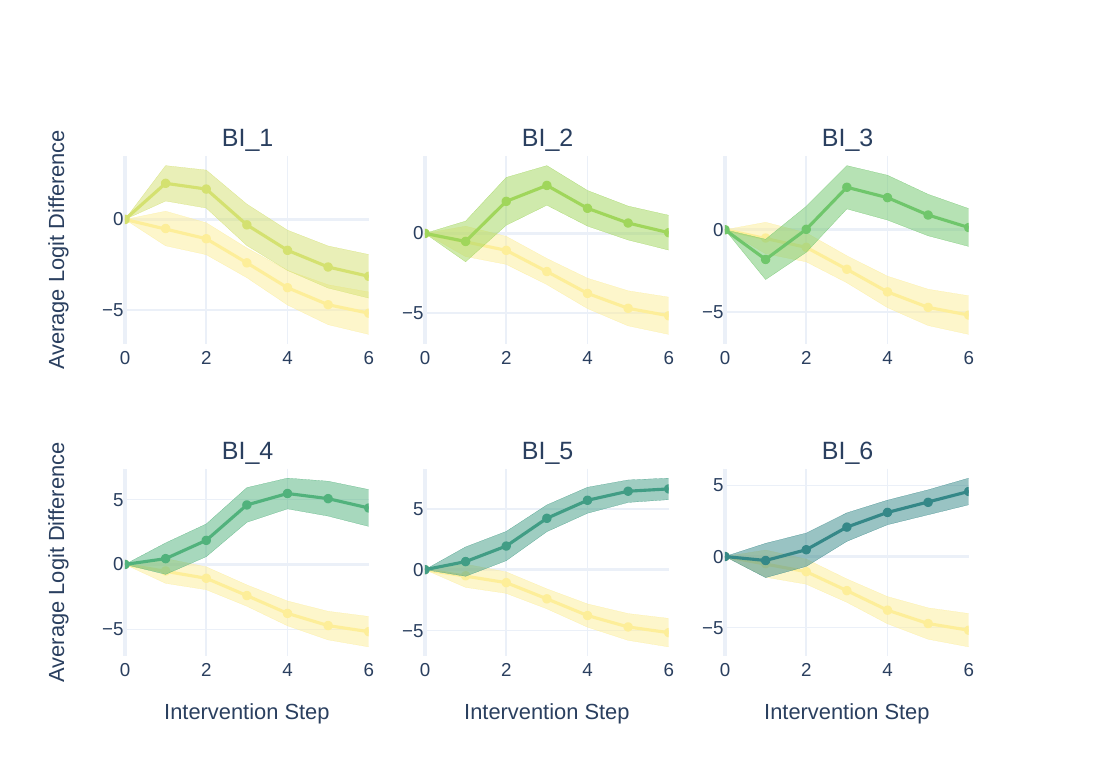}
  \caption{r:bring}
  \label{fig:la3l5}
\end{subfigure}\hfil % <-- added
\begin{subfigure}{0.33\textwidth}
  \includegraphics[width=\linewidth]{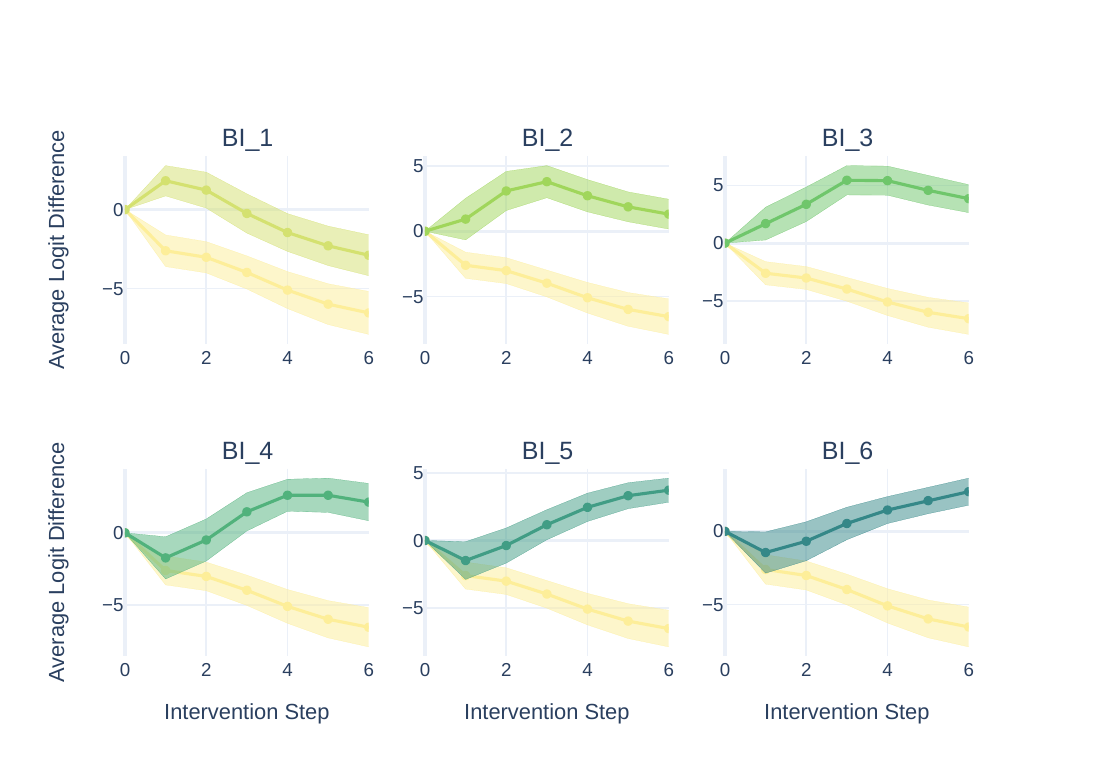}
  \caption{r: push}
  \label{fig:la3l6}
\end{subfigure}
\caption{Logit Difference (LD) for OI-PC based intervention across datasets on Llama3-8B, where x axis denotes the number of intervention steps on $e_0$, y axis does the LD, BI\_i represents each target attribute and the blue line indicates the LD of original attribute (i.e., $a_0$). Here, $l = 10$, $v = 0.65$,  and $\alpha = 2.0$.}
\label{fig:la3ld_a}
\end{figure*}

\begin{figure*}[!h]
    \centering % <-- added
\begin{subfigure}{0.33\textwidth}
  \includegraphics[width=\linewidth]{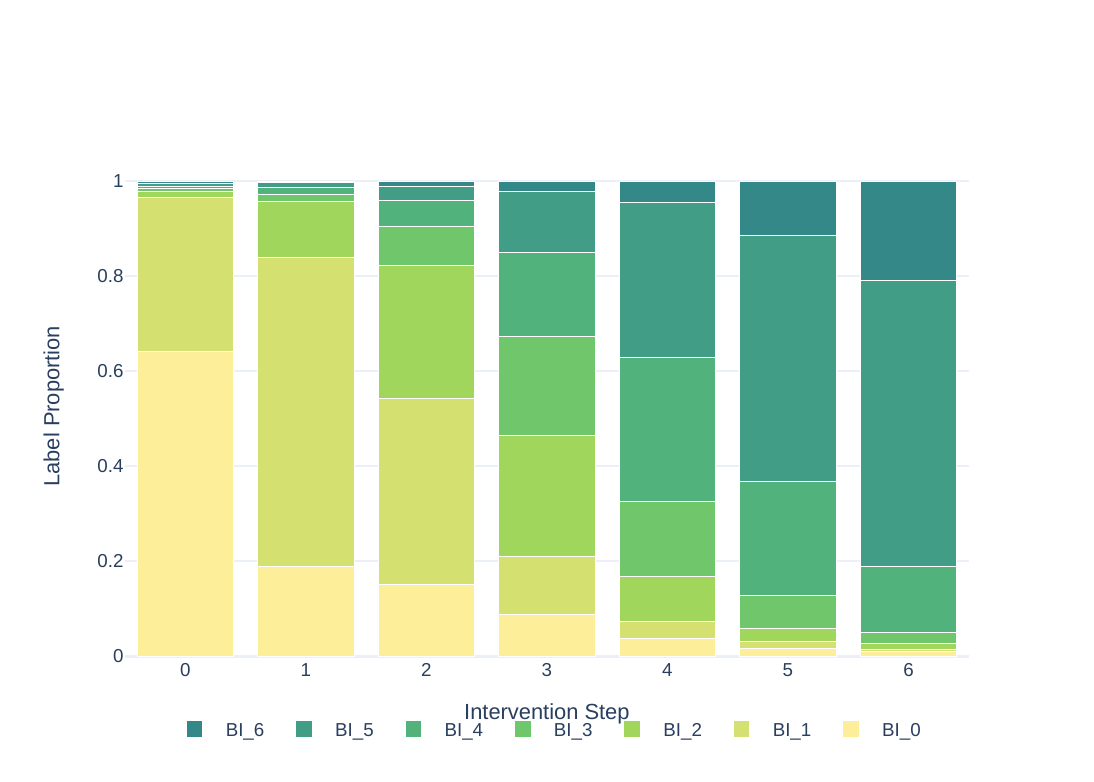}
  \caption{r: is\_in}
  \label{fig:la3f1}
\end{subfigure}\hfil % <-- added
\begin{subfigure}{0.33\textwidth}
  \includegraphics[width=\linewidth]{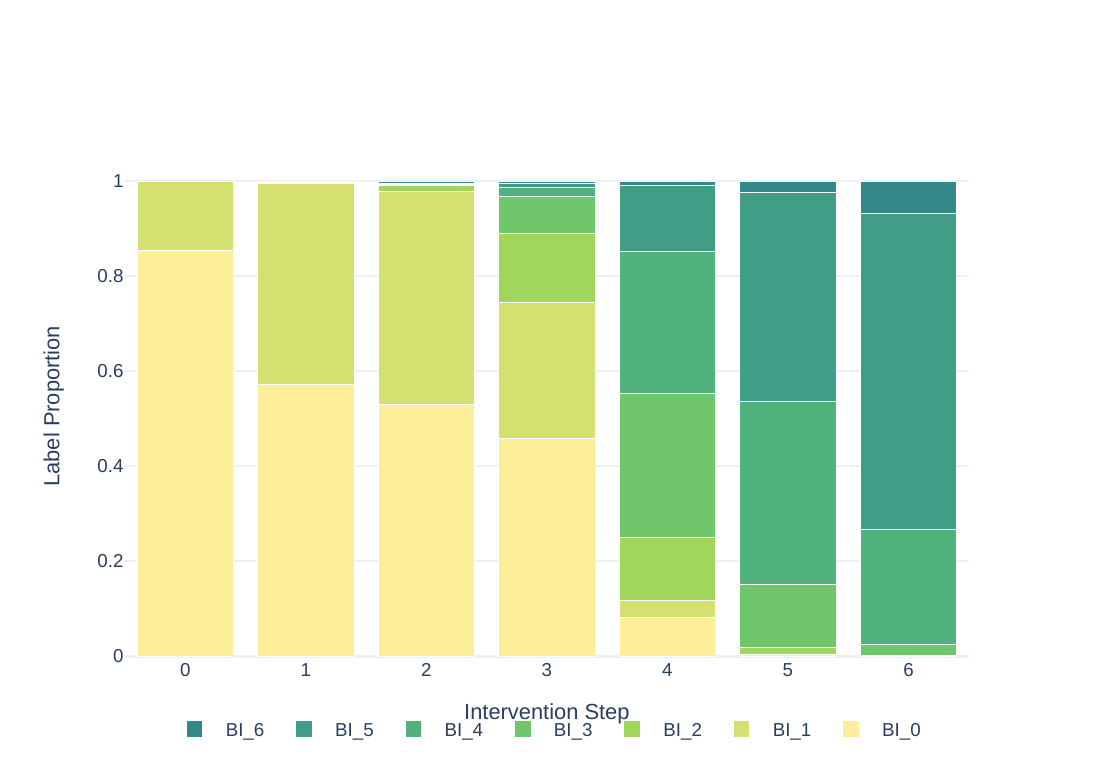}
  \caption{r: sell}
  \label{fig:la3f2}
\end{subfigure}\hfil % <-- added
\begin{subfigure}{0.33\textwidth}
  \includegraphics[width=\linewidth]{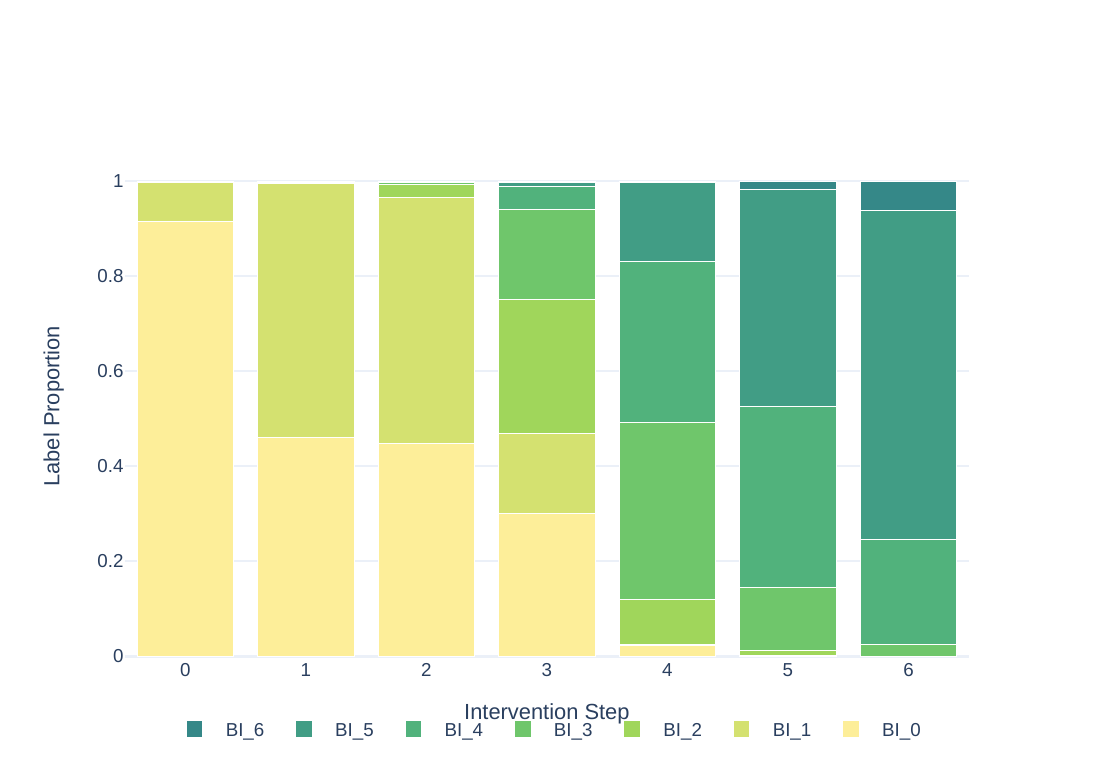}
  \caption{r: apply}
  \label{fig:la3f3}
\end{subfigure}

\medskip
\begin{subfigure}{0.33\textwidth}
  \includegraphics[width=\linewidth]{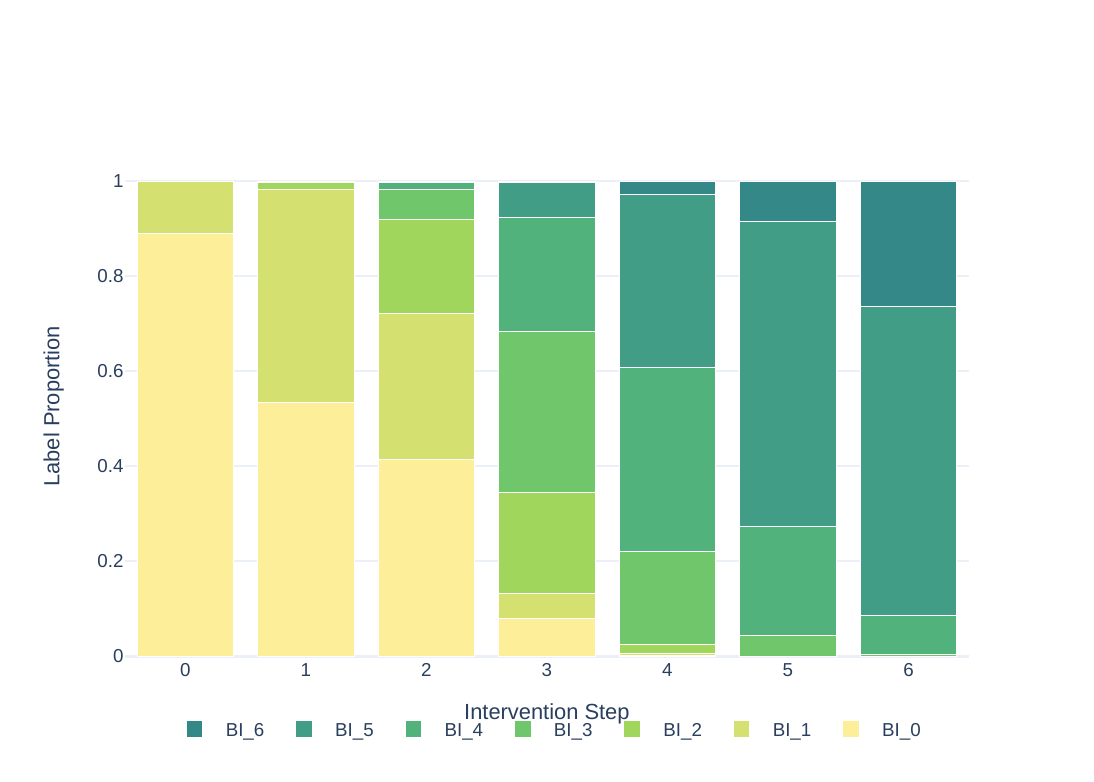}
  \caption{r: move}
  \label{fig:la3f4}
\end{subfigure}\hfil % <-- added
\begin{subfigure}{0.33\textwidth}
  \includegraphics[width=\linewidth]{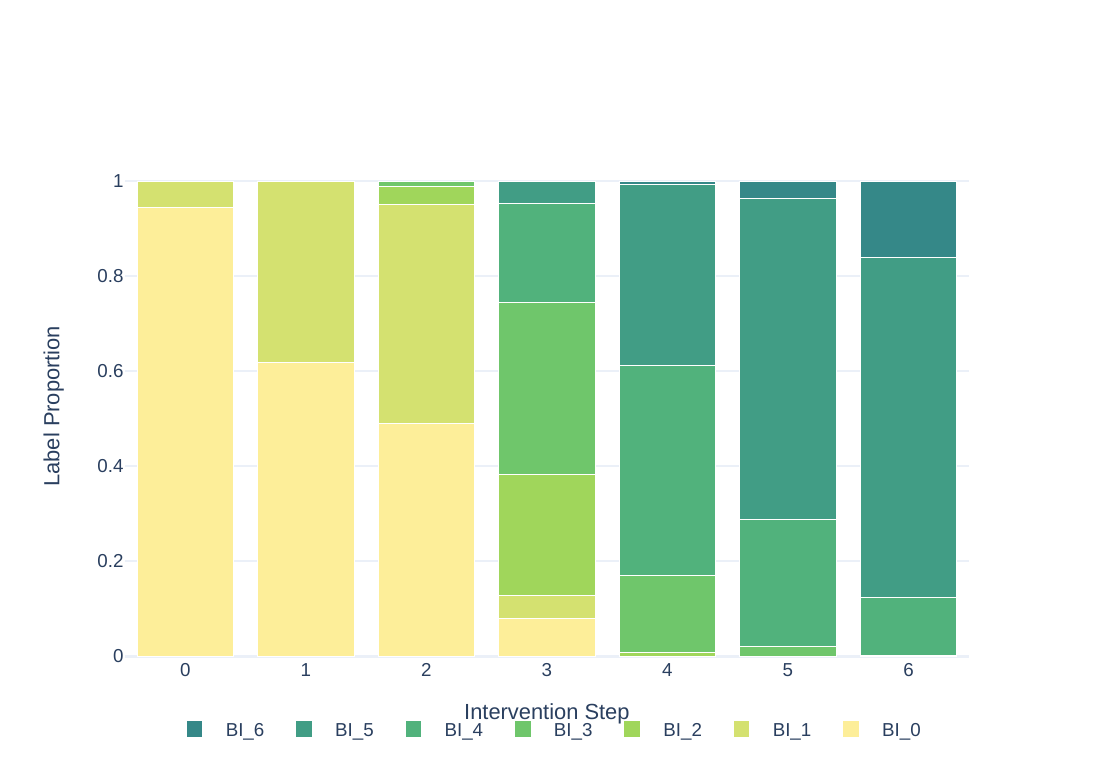}
  \caption{r: bring}
  \label{fig:la3f5}
\end{subfigure}\hfil % <-- added
\begin{subfigure}{0.33\textwidth}
  \includegraphics[width=\linewidth]{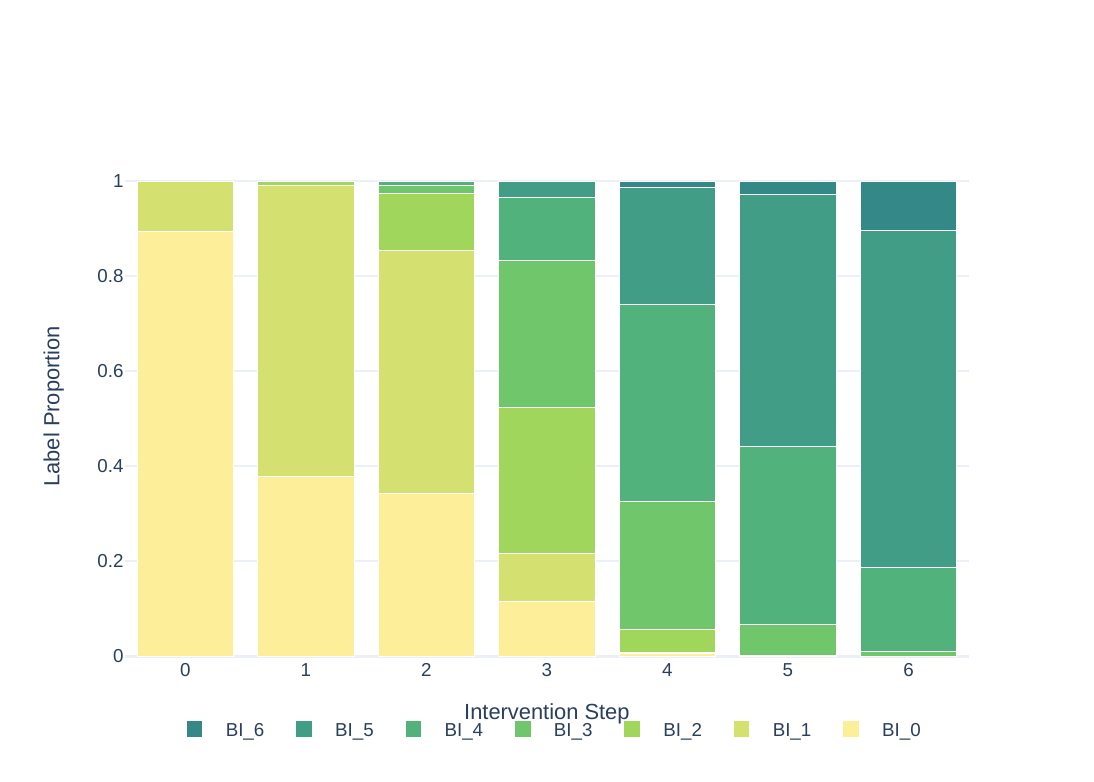}
  \caption{r: push}
  \label{fig:la3f6}
\end{subfigure}
\caption{Logit flip for OI-PC based intervention across datasets on Llama3-8B, where x axis denotes the number of intervention steps on $e_0$, y axis does the proportion of each inferred attribute in model output.}
\label{fig:la3f_a}
\end{figure*}

\clearpage
\subsection{New Dataset with Interjections}
\label{sec:ij_all}

The dataset is created by inserting a sequence of interjections after the first attribute entity pair, as illustrated in Table~\ref{tab:ctx_ij}. Since there is no BI information in the interjection (e.g., $j^{p3}$), adding it only changes the PI of its following entities and attributes. We set the number of interjections as that the PI of last interjection token is larger than the last PI of its original input (e.g., $pi > p5$ in Table~\ref{tab:ctx_ij}). 

\begin{figure*}[h]
    \centering % <-- added
\begin{subfigure}{0.33\textwidth}
  \includegraphics[width=\linewidth]{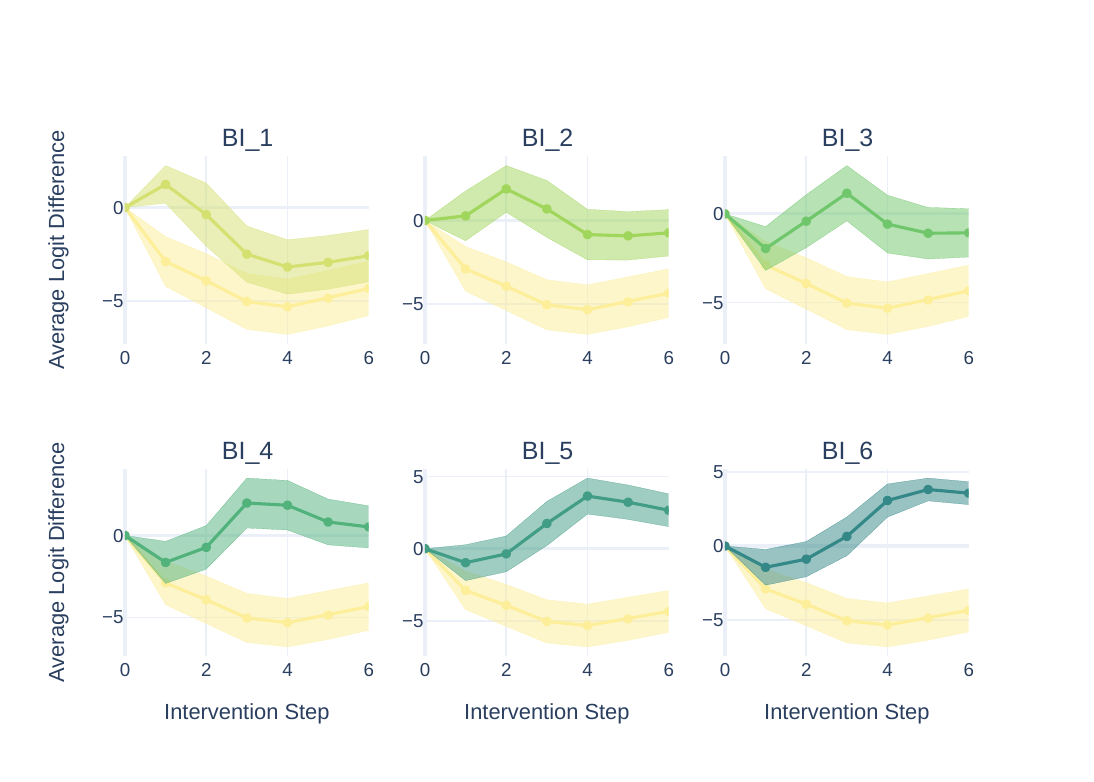}
  \caption{r: is\_in}
  \label{fig:lal1ij}
\end{subfigure}\hfil % <-- added
\begin{subfigure}{0.33\textwidth}
  \includegraphics[width=\linewidth]{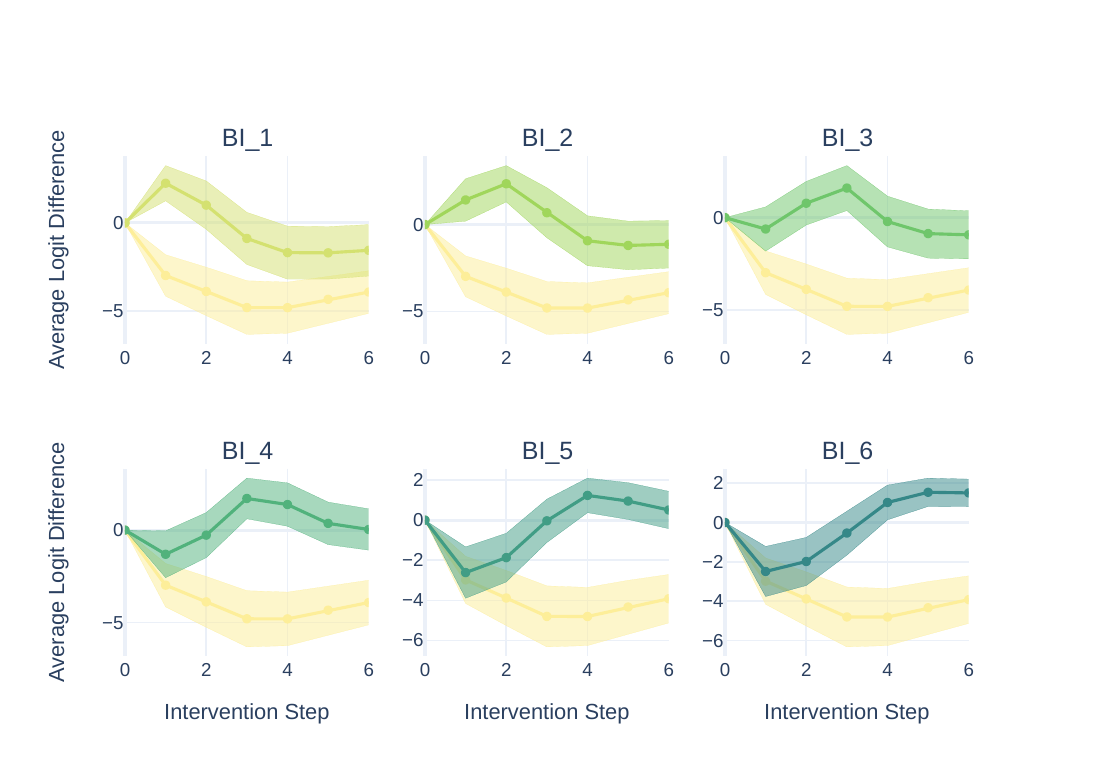}
  \caption{r: sell}
  \label{fig:lal2ij}
\end{subfigure}\hfil % <-- added
\begin{subfigure}{0.33\textwidth}
  \includegraphics[width=\linewidth]{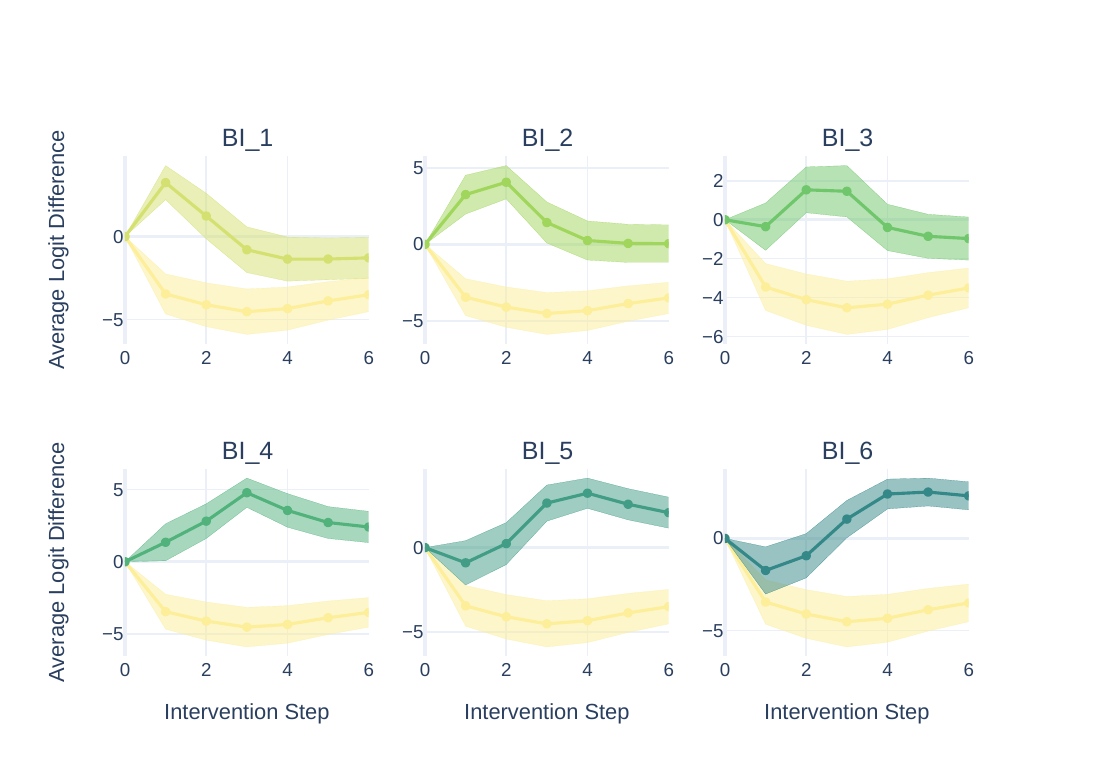}
  \caption{r: apply}
  \label{fig:lal3ij}
\end{subfigure}

\medskip
\begin{subfigure}{0.33\textwidth}
  \includegraphics[width=\linewidth]{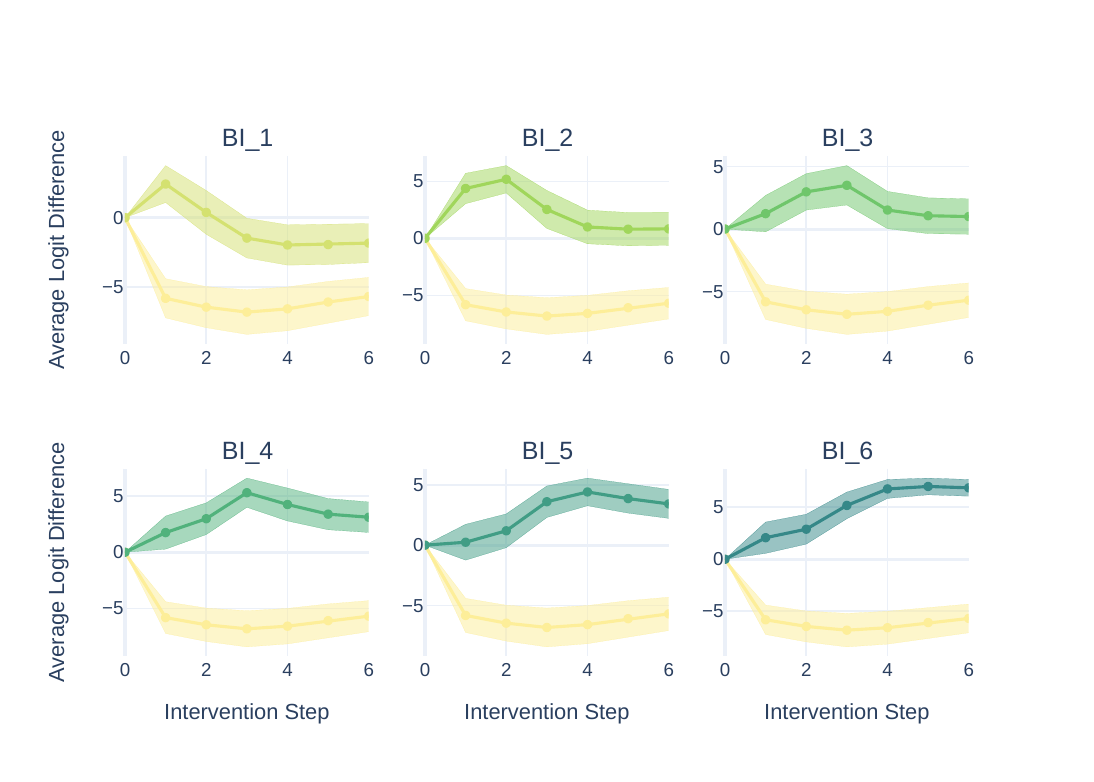}
  \caption{r: move}
  \label{fig:lal4ij}
\end{subfigure}\hfil % <-- added
\begin{subfigure}{0.33\textwidth}
  \includegraphics[width=\linewidth]{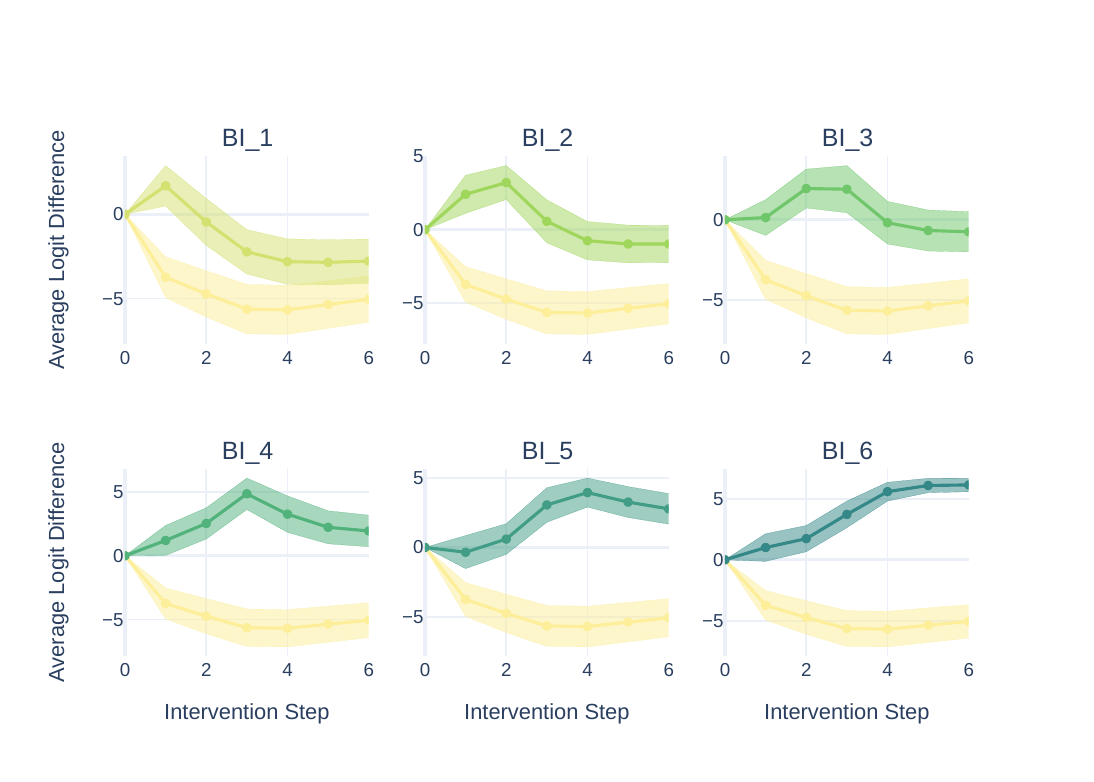}
  \caption{r:bring}
  \label{fig:lal5ij}
\end{subfigure}\hfil % <-- added
\begin{subfigure}{0.33\textwidth}
  \includegraphics[width=\linewidth]{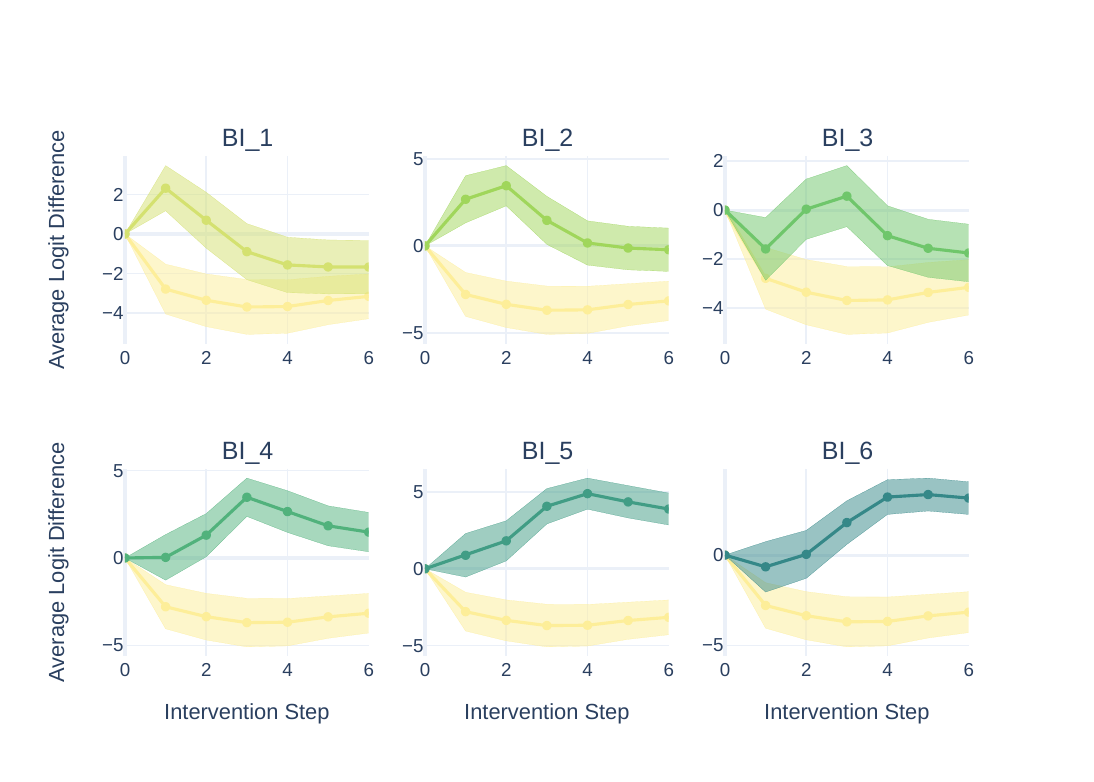}
  \caption{r: push}
  \label{fig:lal6ij}
\end{subfigure}
\caption{Logit Difference for activation patching on the dataset with interjections. Here, $l = 8$, $v = 2.5$,  and $\alpha = 3.0$.}
\label{fig:lald_aij}
\end{figure*}

\begin{figure*}[h]
    \centering % <-- added
\begin{subfigure}{0.33\textwidth}
  \includegraphics[width=\linewidth]{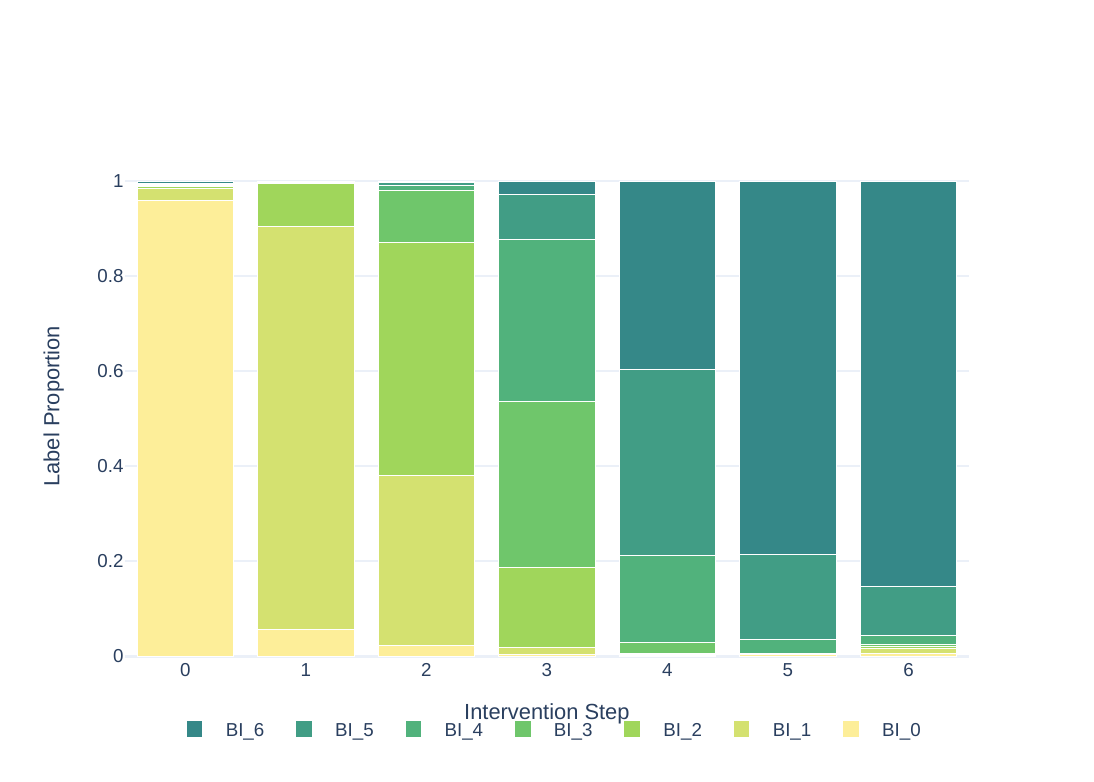}
  \caption{r: is\_in}
  \label{fig:laf1ij}
\end{subfigure}\hfil % <-- added
\begin{subfigure}{0.33\textwidth}
  \includegraphics[width=\linewidth]{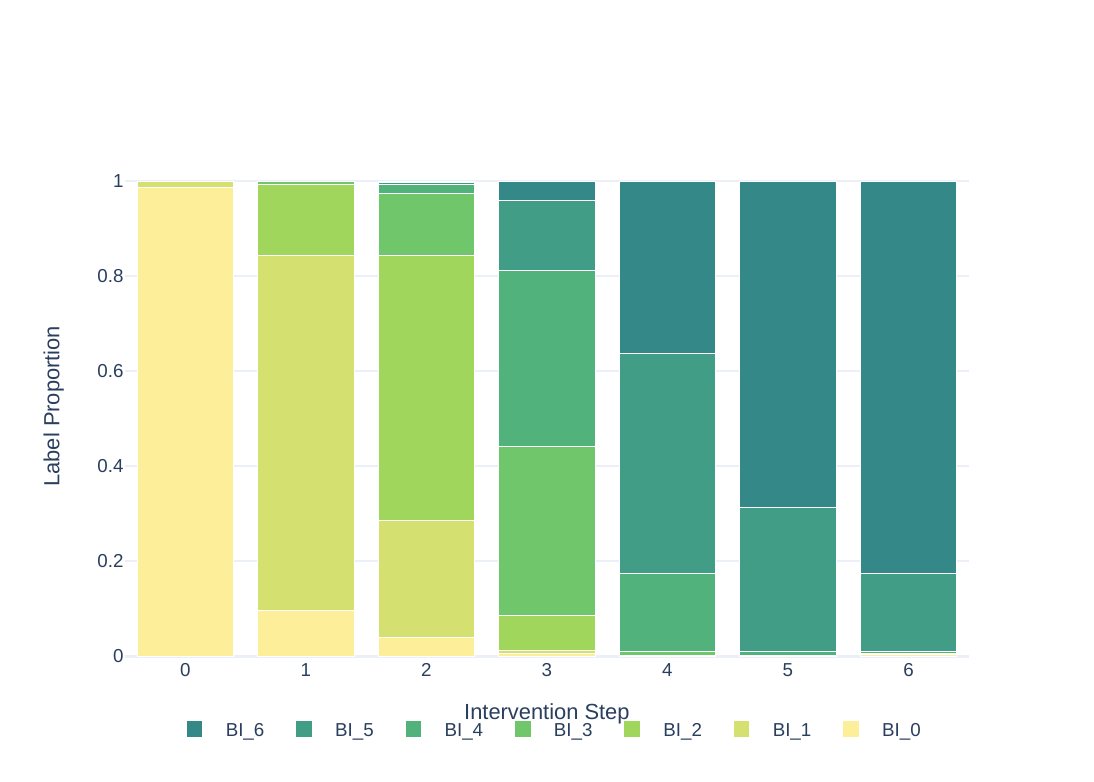}
  \caption{r: sell}
  \label{fig:laf2ij}
\end{subfigure}\hfil % <-- added
\begin{subfigure}{0.33\textwidth}
  \includegraphics[width=\linewidth]{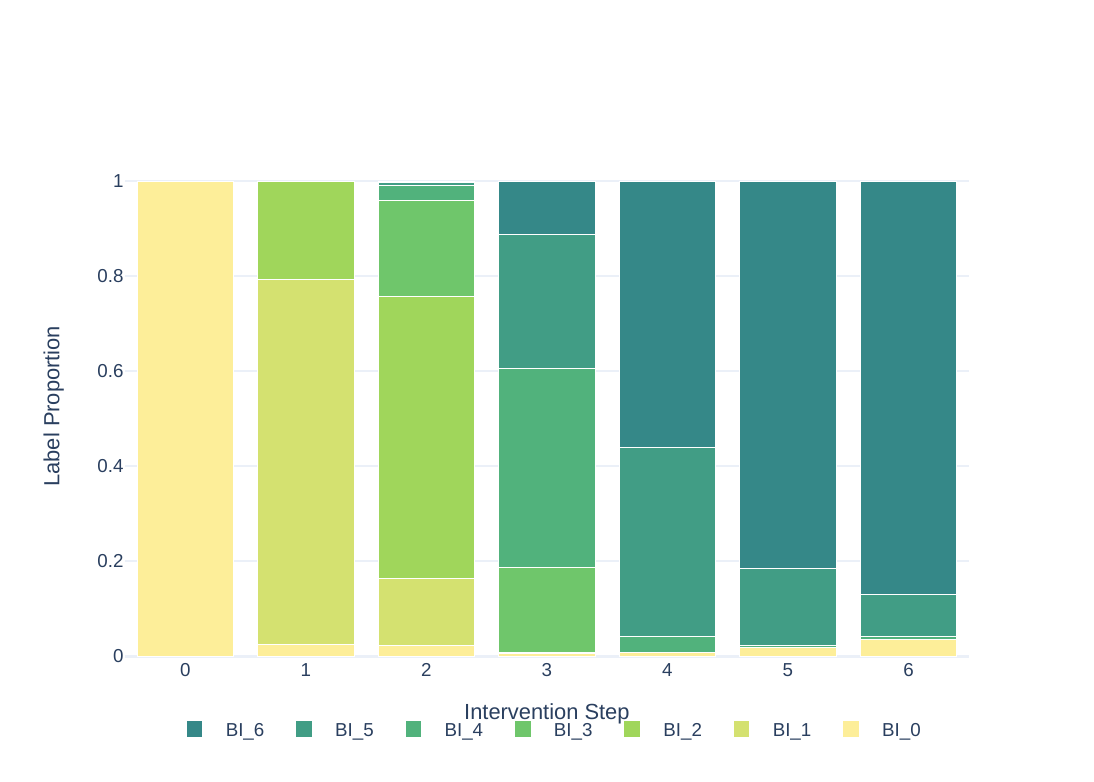}
  \caption{r: apply}
  \label{fig:laf3ij}
\end{subfigure}

\medskip
\begin{subfigure}{0.33\textwidth}
  \includegraphics[width=\linewidth]{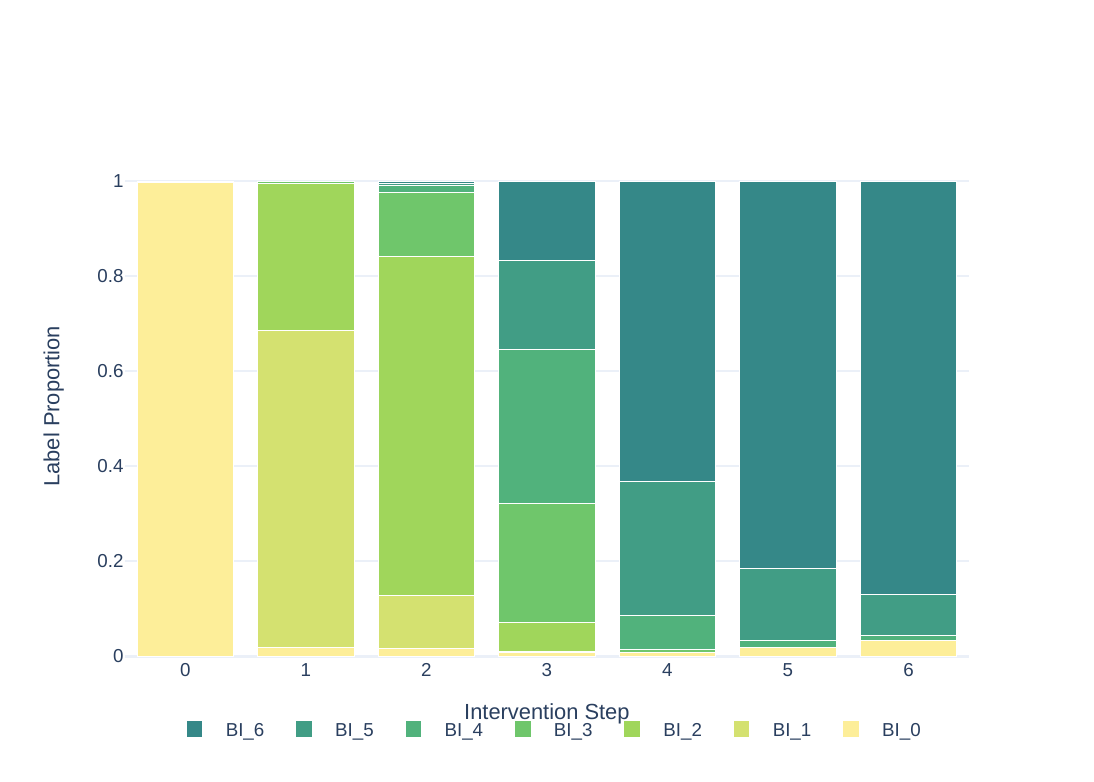}
  \caption{r: move}
  \label{fig:laf4ij}
\end{subfigure}\hfil % <-- added
\begin{subfigure}{0.33\textwidth}
  \includegraphics[width=\linewidth]{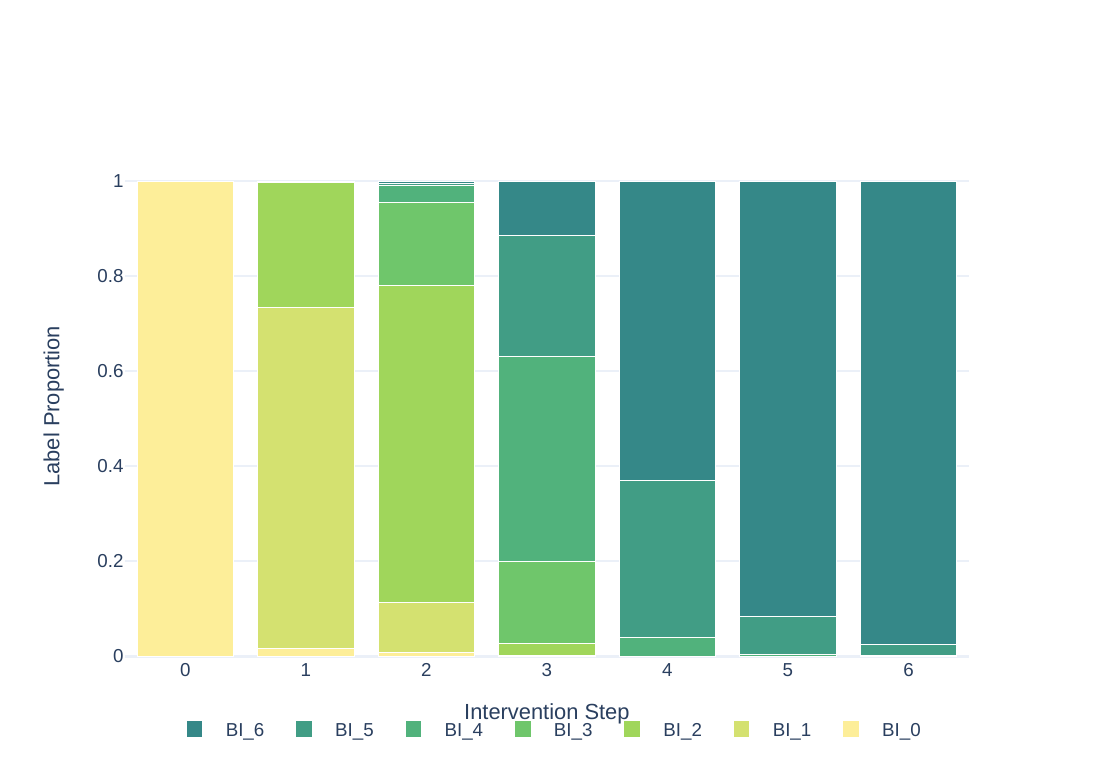}
  \caption{r: bring}
  \label{fig:laf5ij}
\end{subfigure}\hfil % <-- added
\begin{subfigure}{0.33\textwidth}
  \includegraphics[width=\linewidth]{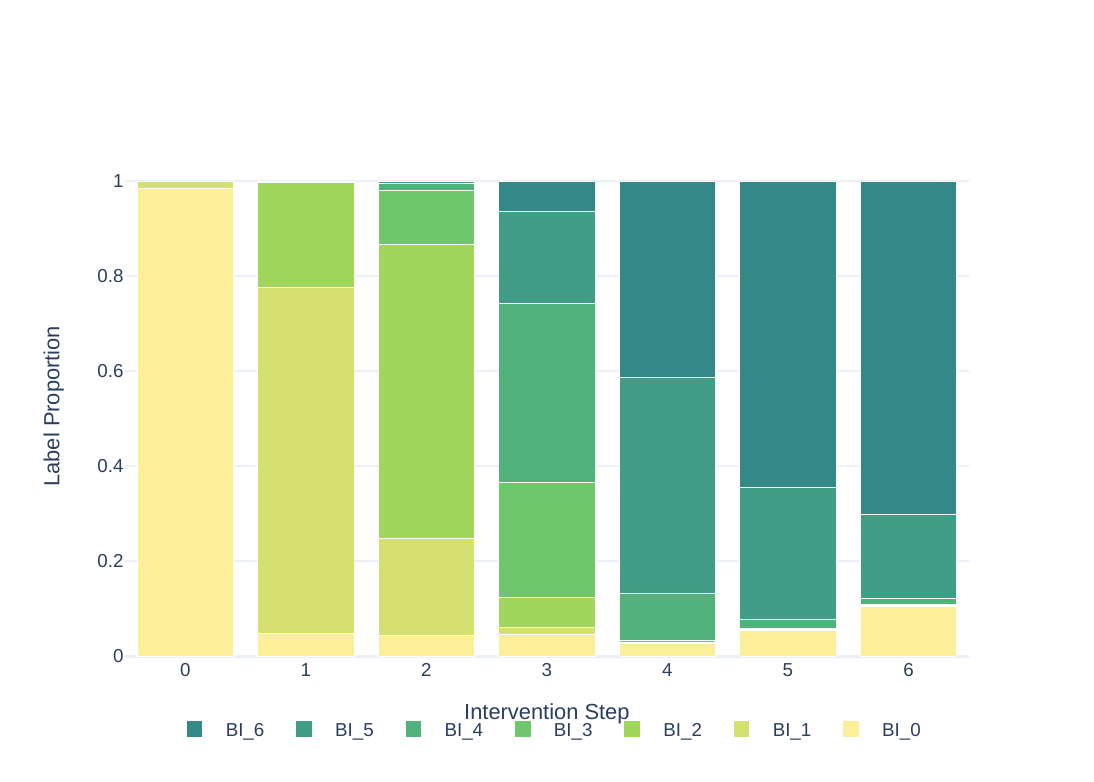}
  \caption{r: push}
  \label{fig:laf6ij}
\end{subfigure}
\caption{Logit Flip for activation patching on the dataset with interjections.}
\label{fig:laf_aij}
\end{figure*}

Based on this dataset, we conduct the same intervention on its OI subspace, as mentioned in Section \secref{sec:ap_subspace}. The counter argument is that the subspace only captures Position Information (PI), and the intervening step only changes the PI information. Specifically, adding one unit of $v$ on $\overrightarrow{e_0^{p1}}$ might convert the PI of target entity from $p1$ to $p3$, and $p3$ is the PI of $e_1$, and its attribute is $a_1$, as shown in Table~\ref{tab:ctx_ij}, and thus the LM swaps the answer from $a_0$ to $a_1$. If it is true, then the same intervention will not change the answer on the new dataset, because following the counter argument, after adding one unit of $v$ on $\overrightarrow{e_0^{p1}}$, the PI of target entity becomes $p3$, and $p3$ is the PI of $j^{p3}$ (i.e., an interjection token). The LM thus would not select $a_1$ as its answer. However, the results on Figure~\ref{fig:lald_aij} and Figure~\ref{fig:laf_aij} show that the subspace intervention on the new dataset achieves similar results as the original one, as shown in Figure~\ref{fig:ld} and Figure~\ref{fig:lf}, proving the counter argument wrong and indicating the independence between OI subspace and PI.

\begin{table}[!h]
\centering
\scalebox{0.9}{
\begin{tabular}{l}
\hline
\textbf{Input (original)} \\\cmidrule(l){1-1}
\makecell[l]{$a_0^{p0} \, r \, e_0^{p1}, \, a_1^{p2} \, r \, e_1^{p3}, \, a_2^{p4} \, r \, e_2^{p5}. \quad e_0^{p1} \ r^{-1} \ ?$} \\\cmidrule(l){1-1}
\textbf{Input (with interjection)} \\\cmidrule(l){1-1}
\makecell[l]{$a_0^{p0} \, r \, e_0^{p1}, \, j^{p2} \,  j \, j^{p3} \, ... j^{pi} \, a_1^{pi+1} \, r \, e_1^{pi+1} ... \quad e_0^{p1} \ r^{-1} \ ?$} \\\cmidrule(l){1-1}
\hline
\end{tabular}
}
\caption{Simplified expression of original input and the one modified with a sequence of interjections, where $j^{p3}$ denotes an interjection $j$, such as ``ah'', with position of $p3$, which is also the position of $e_1^{p3}$ in the original input.}
\label{tab:ctx_ij}
\end{table}

\end{document}